\documentclass[floatfix,amsmath,amssymb,superscriptaddress,longbibliography,twocolumn,showpacs]{revtex4-2}
\usepackage{amssymb}
\usepackage{amsmath}
\usepackage{bbold}
\usepackage{bm}

\usepackage{braket}
\usepackage[title]{appendix}
\usepackage{xcolor}
\usepackage{soul}
\usepackage{graphicx}
\usepackage{textcomp}
\usepackage{enumerate}
\usepackage{multirow}
\usepackage{hyperref}
\usepackage{lipsum, babel}
\graphicspath{{../figura1/}{../figura2/}}

\newcommand*\diff{\mathop{}\!\mathrm{d}}

\newcommand{\rvline}{\hspace*{-\arraycolsep}\vline\hspace*{-\arraycolsep}}
\usepackage{soul}

\begin{document}

\title{Local Kernel Renormalization as a mechanism for feature learning in overparametrized Convolutional Neural Networks}
\author{R. Aiudi}
\affiliation{Dipartimento di Scienze Matematiche, Fisiche e Informatiche,
Universit\`a degli Studi di Parma, Parco Area delle Scienze, 7/A 43124 Parma, Italy}
\affiliation{INFN, Gruppo Collegato di Parma, Parco Area delle Scienze 7/A, 43124 Parma, Italy}
\author{R. Pacelli}
\affiliation{Dipartimento di Scienza Applicata e Tecnologia, Politecnico di Torino, 10129 Torino, Italy}
\affiliation{Artificial Intelligence Lab, Bocconi University, 20136 Milano, Italy}
\date{\today}
\author{A. Vezzani}
\affiliation{Istituto dei Materiali per l'Elettronica ed il Magnetismo (IMEM-CNR), Parco Area delle Scienze, 37/A-43124 Parma, Italy}
\affiliation{Dipartimento di Scienze Matematiche, Fisiche e Informatiche,
Universit\`a degli Studi di Parma, Parco Area delle Scienze, 7/A 43124 Parma, Italy}
\affiliation{INFN, Gruppo Collegato di Parma, Parco Area delle Scienze 7/A, 43124 Parma, Italy}
\author{R. Burioni}
\affiliation{Dipartimento di Scienze Matematiche, Fisiche e Informatiche,
Universit\`a degli Studi di Parma, Parco Area delle Scienze, 7/A 43124 Parma, Italy}
\affiliation{INFN, Gruppo Collegato di Parma, Parco Area delle Scienze 7/A, 43124 Parma, Italy}
\author{P. Rotondo}
\affiliation{Dipartimento di Scienze Matematiche, Fisiche e Informatiche,
Universit\`a degli Studi di Parma, Parco Area delle Scienze, 7/A 43124 Parma, Italy}
\affiliation{INFN, Gruppo Collegato di Parma, Parco Area delle Scienze 7/A, 43124 Parma, Italy}

\begin{abstract}

Feature learning, or the ability of deep neural networks to automatically learn relevant features from raw data, underlies their exceptional capability to solve complex tasks. 
However, feature learning seems to be realized in different ways in fully-connected (FC) or convolutional architectures (CNNs). Empirical evidence shows that FC neural networks in the infinite-width limit eventually outperform their finite-width counterparts. Since the kernel that describes infinite-width networks does not evolve during training, whatever form of feature learning occurs in deep FC architectures is not very helpful in improving generalization. 
On the other hand, state-of-the-art architectures with convolutional layers achieve optimal performances in the finite-width regime, suggesting that an effective form of feature learning emerges in this case. 
In this work, we present a simple theoretical framework that provides a rationale for these differences, in one hidden layer networks. First, we show that the generalization performance of a finite-width FC network can be obtained by an infinite-width network, with a suitable choice of the Gaussian priors (technically, this result holds in the asymptotic limit where the size of the training set $P$ and the size of the hidden layer $N_1$ are taken to infinity keeping their ratio $\alpha_1 = P/N_1$ fixed).
Second, we derive a finite-width effective action for an architecture with one convolutional hidden layer and compare it with the result available for FC networks. Remarkably, we identify a completely different form of kernel renormalization: whereas the kernel of the FC architecture is just globally renormalized by a single scalar parameter, the CNN kernel undergoes a local renormalization, meaning that the network can select the local components that
will contribute to the final prediction in a data-dependent way. This finding highlights a simple mechanism for feature learning that can take place in overparametrized shallow CNNs, but not in shallow FC architectures or in locally connected neural networks without weight sharing.
\end{abstract}

\maketitle

\section{Introduction} 

Deep learning achieves state-of-the-art performances on a variety of pattern recognition tasks, ranging from computer vision to natural language processing \cite{GoodfellowBook}. A minimal theory of deep learning should be able (at least) to predict practically relevant scores, such as the training and test accuracy, from knowledge of the training data. Another even more fundamental aspect that such a theory should quantitatively address is the so-called \emph{feature learning regime} \cite{6472238, yu2013feature}, i.e. the capability of deep networks to automatically detect useful representations from raw data.

Concerning the first aspect, fundamental theoretical progress has been achieved in the so-called \emph{infinite-width limit} \cite{Neal,NIPS1996_ae5e3ce4,g.2018gaussian,LeeGaussian,garriga-alonso2018deep,novak2019bayesian, JacotNTK,ChizatLazy,lee2019wide}, which is informally defined as the regime where the size of each hidden layer $N_{\ell}$ ($\ell =1,\dots, L$, $L$ being the (finite) depth of the network) is much larger than the size of the training set $P$. Here, one shows that the stochastic process that describes information flow in the deep neural network is a familiar Gaussian process, which is completely determined by a non-linear kernel.

A fundamental consequence of this finding is that learning in the infinite-width limit is equivalent to kernel learning \cite{cortes1995support,PhysRevLett.82.2975, pmlr-v119-bordelon20a, canatar2021} with a static kernel, completely fixed by the statistics of the weights at initialization, that does not evolve during training. This last observation suggests that feature learning is essentially absent in infinite-width networks \cite{Vyas2022}. Notably, whereas original work on the infinite-width limit considered deep architectures with FC hidden layers only, the generalization to architectures with convolutional layers is straightforward \cite{novak2019bayesian}, if one replaces neurons in the FC hidden layers with the channels/convolutional filters of the CNN.

\begin{figure*}
    \centering
    \includegraphics[width = 1.\textwidth]{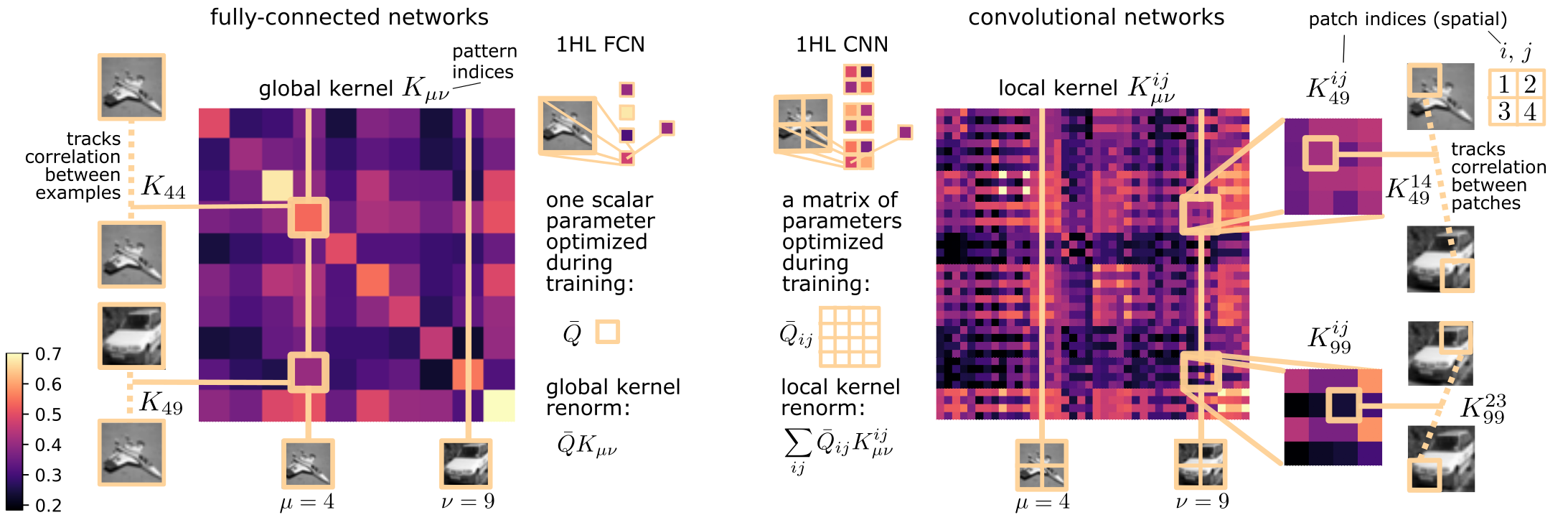}
    \caption{\textbf{Global and local kernels in FC and convolutional one hidden layer networks}. Our calculations in the proportional limit $P,N_1 \to \infty$ at fixed $\alpha_1=P/N_1$ reveal a striking difference between FCNs and CNNs. In the first case, the NNGP kernel $K_{\mu\nu}$ that enters into the Bayesian effective action is a global one, which is only capable of tracking global correlations between different training patterns $(\mu,\nu)$ (left). A FCN network in this regime will use the information contained in the dataset to fine-tune a single parameter, $\bar Q$, that will globally rescale the kernel matrix $K_{\mu \nu}$. On the other hand, in the CNN case, we identify a local kernel with two additional (spatial) indices. If we consider $2d$ images, the effect of the local components is to track local correlations between different patches $i$ and $j$ of (possibly) different pairs of images $(\mu,\nu)$ (right). The spatial information contained in these correlations allows the CNN to optimize a matrix of parameters $\bar Q_{ij}$, which will renormalize the kernel matrix $K^{ij}_{\mu \nu}$ in a non-trivial way. In the figure, we display the four-index local kernel $K_{\mu\nu}^{ij}$ as a matrix using the multi-index notation $K_{(\mu,i),(\nu,j)}$ and choosing an ordering for the multi-index $(\mu,i)$. Global and local kernels are computed for a subset of $P = 10$ gray-scaled CIFAR10 images down-sized to $N_0 = 784$. The local kernels are empirically evaluated for $2d$ convolutions with linear size of the filter mask $M = 14$ and stride $S = 14$ (non-overlapping filters).}
    \label{fig:kernel_local_vs_global}
\end{figure*}

Recently, the authors of Ref. \cite{NEURIPS2020_ad086f59} conducted a large-scale empirical study comparing finite-width FC networks, CNNs with finite number of channels and their infinite-width limit kernel counterparts, where they explore a plenitude of possible settings to improve generalization. 
%The authors explored a variety of possible settings to improve the generalization performance combining centering, large learning rate, L2 regularization, ensembling, under-fitting by early-stopping and ZCA whitening of the inputs. 
This analysis prompts a few striking empirical observations: (i) infinite-width kernels systematically outperform their finite-width counterpart in the case of FC deep neural networks; (ii) CNNs with finite number of channels often outperform their corresponding infinite-width kernel performance. 
%Such evidence is further corroborated by a difference in the behavior of the test accuracy as a function of the size of the HLs: whereas for FC networks the accuracy monotonically approaches the kernel performance from below, in CNNs one often observes a non-monotonic behavior with a region, at finite-width, where the network is capable of outperforming the CNN GP. 

Several similar observations have been reported in the literature: the authors of Ref. \cite{atanasov2022onset} very recently pointed out that infinite-width deep FC neural networks eventually outperform their finite-width counterpart as the size of the training set grows. In one of the seminal papers dealing with the infinite-width limit \cite{LeeGaussian}, the authors observe that increasing the hidden layers size leads to optimal test accuracy on deep FC architectures trained on MNIST and on CIFAR10, two of the benchmark datasets for computer vision learning problems. In Ref. \cite{novak2019bayesian} this observation is extended to \emph{locally connected networks} without weight sharing (LCNs). The interplay between the lazy training regime and the mean field limit \cite{doi:10.1073/pnas.1806579115} has been the subject of a thorough investigation in \cite{GEIGER20211,Geiger_2020}. 

These observations suggest that CNNs leverage a better feature-learning mechanism at finite width than FCNs and LCNs, and prompt at least two conceptual questions: (i) why is it ultimately convenient to employ large-width architectures when only FC layers are available? (ii) Why is this not the case when convolutional layers are employed? And how does a CNN operatively exploit the finite-width regime for efficient feature learning? Preliminary theoretical work in the direction of understanding the feature learning regime of deep nets was carried out in Refs. \cite{seroussi2023natcomm, NEURIPS2021_b24d2101, NEURIPS2021_cf9dc5e4, PhysRevE.105.064118, zavatone-veth2021exact, PDLT-2022, hanin2023random, antognini2019finite, yaida2020nonGauss, aitchison2020bigger, yang2023theory, favaro2023quantitative}. The authors of Refs.  \cite{cagnetta2023what, favero2021locality} analytically investigated the advantages of employing convolutional neural tangent kernels in the infinite-width limit and understood why infinite-width FCNs perform worse in the mean field regime than in the lazy-training one \cite{petrini2022learning}. 

In this work, we try to rationalize the aforementioned empirical observations through the lens of \emph{kernel renormalization}. This represents a physical consequence of a recently-derived effective action for Bayesian learning in finite-width deep neural networks \cite{ariosto2022statistical} (formally speaking, the finite-width regime is defined as the thermodynamic limit where the size of the FC hidden layers and of the training set are taken to infinity, $N_\ell, P \to \infty$ and their ratio $\alpha_\ell = P/N_\ell$ is kept finite). First, we show that the theoretical framework of Ref. \cite{ariosto2022statistical} provides a simple, yet very instructive answer to question (i) for one hidden layer networks, by showing that the performance of a finite-width shallow network can be obtained with the corresponding infinite-width kernel and a suitable choice of the Gaussian priors over the weights of each layer. Second, we derive an effective action for a simple architecture with one convolutional hidden layer, and we compare the result with the one available for FCNs. We find a striking difference in the way the kernel of the two architectures renormalize at finite width: whereas the FC kernel is just globally renormalized by a scalar parameter, the CNN kernel undergoes a local renormalization, meaning that many more free parameters are allowed to be fine-tuned during training (see also Fig. \ref{fig:kernel_local_vs_global}). We employ this finding to provide preliminary insight to question (ii): we highlight a simple mechanism for feature learning that can take place in finite-width shallow CNNs, but neither in shallow FC architectures nor in LCNs without weight sharing.

\section*{Results}
 
\section{Finite-width one hidden layer FC NNs cannot outperform infinite-width kernels in the overparametrized regime} 
\label{sec:fc_cannot_outperform}

 We consider a supervised regression problem with training set $\mathcal T_P = \{ x^\mu, y^\mu\}_{\mu=1}^P$, where each $x^\mu \in \mathbb R^{N_0}$ and the corresponding labels $y^\mu \in \mathbb R$. We also restrict our analysis to quadratic loss function $\mathcal L$. For a generic one hidden layer network that implements the function $f_{\mathrm{1HL}}(x)$, we define the loss function as:
 \begin{equation}
    \mathcal L = \frac{1}{2} \sum_{\mu} \left[ y^\mu - f_{\mathrm{1HL}}(x^\mu) \right]^2 + \frac{T \,\lambda_0}{2 } \Vert W \Vert^2 + \frac{T \, \lambda_1}{2} \Vert v \Vert^2 
    \label{loss}
 \end{equation}
 where $W$ and $v$ are respectively the first and last layer weights, and $T$ is the temperature. The parameters $\lambda_0, \lambda_1$ can be equivalently thought as Gaussian priors over the weights of each layer, or as $L^2$ regularization terms rescaled by the temperature $T$. A neural network with one FC hidden layer is formally defined starting from the pre-activations at the first layer:
\begin{equation}
h_{i_1} (x)= \frac{1}{\sqrt {N_{0}}} \sum_{i_0=1}^{N_{0}} W_{i_1 i_0} x_{i_0} + b_{i_1}\,,
\end{equation}
%
%where $W$ and $b$ are respectively the weights and the biases. 
where $b$ are the first-layer biases.
Since we are interested in regression problems, we add one last readout layer, and we define the function implemented by the one hidden layer FCN as:
\begin{equation}
f_{\textrm{FCN}}(x) = \frac{1}{\sqrt{N_1}} \sum_{i_1=1}^{N_1} v_{i_1} \sigma \!\left[ h_{i_1} (x)\right]\,,
\label{FCN_function}
\end{equation}
%where $ v$ is the vector of weights of the last layer and 
where $\sigma$ is a non-linear activation function. An effective action arises in this setting when one considers the canonical partition function at temperature $T=1/\beta$ associated to the train loss function $\mathcal L$ and the thermodynamic limit $ N_1, P \to \infty$ at fixed ratios $\alpha_1 = P/N_1$. The calculation amounts to reduce the integral over the weights of the network to a form suitable to saddle-point integration over two order parameters $\{Q, \bar Q\}$, $Z = \int d Q d \bar Q e^{-N_1/2 S_{\textrm{FCN}} (\{Q, \bar Q\})}$. In the case of odd-activation function, the action is given by \cite{ariosto2022statistical}:

\begin{align}
S_{\textrm{FCN}}&=  -Q \bar Q + \log(1+Q) \nonumber \\
&+\frac{\alpha_1}{P}\text{Tr}\log \beta \left(  \frac{\mathbb{1}}{\beta} +K^{(\mathrm{R})}(\bar Q)\right) \notag\\
&+\frac{\alpha_1}{P} y^T \left( \frac{\mathbb{1}}{\beta} + K^{(\mathrm{R})}(\bar Q)\right)^{-1} y\,,
\label{DNNeffectiveaction}
\end{align}
where $y$ is the vector of labels $y = (y^1,\dots,y^P)$. The renormalized kernel $K^{(\mathrm{R})}(\bar Q)$ is a $P \times P$ matrix that processes pairs of input training data $( x^\mu,  x^\nu)$ and it is given by: 
\begin{align}
K^{(\mathrm{R})}(\bar Q) = \frac{\bar Q}{\lambda_1} K (C)\,, \, \, \, \quad C_{\mu \nu} = \frac{1}{\lambda_0} \frac{ x^\mu \cdot  x^\nu}{N_0}\,.
\label{K_LQ}
\end{align}
The non-linear operator $K$ takes as input any (symmetric) $P \times P$ matrix $M$ and computes a new matrix in the following way:
\begin{align}
\label{eq:K_munu}
K_{\mu\nu} (M) &= \int d^2 t\, \mathcal N_{t} \left(0,  \tilde M_{\mu\nu}\right) \sigma (t_1) \sigma(t_2)\,, \\
\tilde M_{\mu\nu} &= \begin{pmatrix}
M_{\mu\mu} & M_{\mu\nu} \\
M_{\mu\nu} & M_{\nu\nu}
\end{pmatrix} \,.
\end{align}
Here and in the following we are denoting normalized Gaussians as $\mathcal N_x \left( 0, \Sigma \right) \equiv \textrm{exp} \left( -x^\top \Sigma^{-1} x/2\right)/\sqrt{\mathrm{det}2\pi\Sigma }$. 
%The parameters $\lambda_0, \lambda_1$ can be equivalently thought as Gaussian priors over the weights of each layer or as $L^2$ regularization terms in the loss rescaled by the temperature $T$. 
%\MP{la loss non viene mai scritta esplicitamente, forse vale la pena metterla per far capire meglio chi sono i prior, oppure è scontato per chi è della comunità?}.

It is worth noticing that the minimization of the effective action in Eq. \eqref{DNNeffectiveaction} is straightforward if $\alpha_1 \to 0$, since one easily finds that $Q^* = 0$, $\bar Q = 1$ and recovers the well-known infinite-width limit Neural Network Gaussian Process (NNGP) kernel. At finite $\alpha_1$, one finds data-dependent solutions for the order parameters that produce a renormalization of the infinite-width kernel, as expressed by Eq. \eqref{K_LQ}.

The effective action at finite width presented in Eq. \eqref{DNNeffectiveaction} has been firstly derived for deep linear networks in Ref. \cite{SompolinskyLinear}. The non-asymptotic evaluation of the partition function in the linear case is given in Ref. \cite{doi:10.1073/pnas.2301345120} in terms of Meyer-G functions. An effective action for globally gated deep linear networks has also been derived \cite{li2022globally}. In Ref. \cite{ariosto2022statistical} we obtained the effective action for one FC hidden layer networks with generic activation function leveraging on a Gaussian equivalence \cite{mei2019, goldt2020gaussian, Gerace_2021,loureiro2021learning} informally justified via a  generalized central limit theorem due to Breuer and Major \cite{BM}.  A result for $L$ layers can be somewhat bootstrapped from the observation that the stochastic process that describes information flow in shallow non-linear and deep linear networks in the proportional limit is related to a Student's $t$-process \cite{Shah2014} (see also \cite{ariosto2022statistical} for a more detailed discussion). Recent work considers the proportional setting in a teacher-student scenario, averaging over the data distribution using the replica method \cite{cui2023optimal, schroder2023deterministic} and a first preliminary rigorous result in an almost-proportional limit \cite{camilli2023fundamental} appeared very recently, leveraging the interpolation method \cite{guerra2002thermodynamic, AGLIARI2020254}.

We now highlight a straightforward consequence of the theoretical framework presented above. First, we notice that the renormalized kernel $K^{(\mathrm{R})}$ enters into the predictor's statistics for a new unseen test element exactly in the same way the NNGP kernel does in the infinite-width limit, and thus it completely determines the generalization performance at finite width, once the saddle-point equations for the parameter $\bar Q$ are solved \cite{ariosto2022statistical}. Second, we observe that the scalar parameter $\bar Q$ always appears in combination with the corresponding Gaussian prior $\lambda_1$ as $\bar Q/\lambda_1$, meaning that, once evaluated on the saddle-point, $\bar Q$ is just a data-dependent scalar renormalization of the Gaussian prior. This implies that, once the size of the training set and the activation function $\sigma$ are fixed, it is always possible to re-obtain the performance of any finite-width network just by carefully fine-tuning the Gaussian prior $\lambda_1$ in the corresponding infinite-width kernel. Therefore, the generalization performance of any finite-width one hidden layer FC network is bounded by the one of a suitable infinite-width kernel with optimal choice of the Gaussian prior.

The empirical observation that infinite-width one hidden layer FCNs seem to systematically outperform their finite-width counterpart --which now finds a possible explanation in this framework-- points to the somewhat disappointing consequence that feature learning is not particularly effective in networks with FC layers alone at finite width. %We notice, \emph{en passant}, that this is also suggested from the way the infinite-width kernel renormalizes in the deep learning effective action: in the particular case of one-hidden layer architectures, the finite-width kernel $K_1^{(\mathrm{R})}$ is just a global renormalization by a data-dependent scalar factor of the kernel $K_1$ that describes the infinite-width limit. 

We stress here that the aforementioned observation is not ruling out at all other possible forms of feature learning in FC architectures: %(i) the deep learning effective action for $L > 1$, at the moment, is just an approximation and we have no convincing reasons to believe it is exact at finite width. More complex forms of kernel renormalization may be possible with more than one hidden layer; 
(i) Renormalization of the infinite-width kernel is not the only source of feature learning possible. Higher order kernels, irrelevant in the infinite-width limit, may play a role in feature learning, especially as long as one considers the case where the size of the dataset $P$ roughly scales as the number of parameters $\sim L \times N^2$ ($N_\ell = N$ $\forall \ell$) of the FC deep neural network. The recent work \cite{doi:10.1073/pnas.2201854119} is a notable example of how a FC network can learn a convolutional structure in special settings; (ii) Our result holds for Bayesian learning, i.e. when the weights of the networks are sampled from the canonical Gibbs ensemble. This is only obtained if training is governed by a Markov chain Monte Carlo. Practical learning algorithms with state-of-the-art optimizers may behave differently and possibly activate alternative forms of feature learning; (iii) we are limiting our analysis to the standard parametrization setting where the last layer is normalized as $1/\sqrt{N_1}$, and we cannot describe the mechanism for feature learning that may occur in the mean field regime \cite{doi:10.1073/pnas.1806579115} where the last layer is normalized as $1/N_1$ (we note that the framework of \cite{seroussi2023natcomm} seems more suitable to deal with this case). Nonetheless, it must be also said that the empirical evidence provided in Refs. \cite{atanasov2022onset, LeeGaussian, novak2019bayesian, NEURIPS2020_ad086f59, Geiger_2020} is quite against the fact that these forms of feature learning can play a significant role in improving the generalization performance of FC networks. 

We now move to the second question raised in the Introduction of this manuscript. How --contrarily to what occurs in deep nets with FC layers only-- do deep architectures with convolutional layers exploit the finite-width regime for efficient feature learning? In view of the previous discussion, one possibility is that CNNs at finite width could be able, in some way, to break the global (trivial) renormalization of the infinite-width kernel that occurs in FC deep networks. In order to test this hypothesis, in the next section we investigate a simple model of CNN with one convolutional hidden layer.

\section{Finite-width effective action for a shallow convolutional network} 

The two fundamental ingredients of a convolutional layer are \emph{local connectivity} and \emph{weight sharing}: whereas a given neuron in a FC hidden layer receives input from all the neurons in the previous layer, neurons in a CNN are arranged in a $d$-dimensional array that reflects the corresponding $d$-dimensional arrangement of the input data (e.g. $d=2$ for images). Each neuron in a given layer here interacts only with a local neighborhood of neurons in the previous layer, in a translational invariant way implemented via a shared (usually small) $d$-dimensional mask of learnable weights. These operations define a single convolutional channel that takes as input a $d$-dimensional array and outputs another $d$-dimensional array, whose dimensions are determined by technical details (such as the stride, padding and dimension of the filter mask). Usually, many of these channels are piled up in a convolutional layer to form, overall, a $(d+1)$-dimensional array. 

Let us now define a simple architecture with one hidden convolutional layer. For simplicity, we will restrict our analysis to one-dimensional convolutions, but the model can be easily generalized to $d$-dimensional convolutions. Pre-activations in the hidden layer are given by:
\begin{equation}
h_{i}^a (x)= \frac{1}{\sqrt{M}}\sum_{m = -\lfloor M/2\rfloor}^{\lfloor M/2\rfloor} W_m^{a} x_{S i +m}\,.
\end{equation}
Here $M$ is the dimension of the channel mask, $S$ is the stride, the index $i =1,\dots, \lfloor N_0/S\rfloor$ runs over the input coordinates, the index $a = 1,\dots, N_c$ runs over the channels and the index $m$ moves through the spatial mask of the convolutional filter. For simplicity, we define periodic boundary conditions (PBCs) over the input coordinates, and we consider odd values of $M$.  As for the one hidden layer FC network, we add one readout layer. In conclusion, the function implemented by the CNN is given by:
\begin{equation}
f_{\textrm{CNN}} (x) = \frac{1}{\sqrt{N_c \lfloor N_0/S \rfloor }} \sum_{i=1}^{\lfloor N_0/S \rfloor} \sum_{a=1}^{N_c} v_i^a \sigma \left[h_i^a (x)\right]\,,
\label{1hl_CNN}
\end{equation}
where $\sigma$ is a odd activation function and the $v$'s are the learnable weights of the readout layer. Our goal is to derive an effective action in the same setting of Refs. \cite{ariosto2022statistical}, in the thermodynamic limit where $(N_c, P) \to \infty$ and their ratio $\alpha_c= P/N_c$ is finite.

The partition function for the learning problem with the aforementioned simple CNN can be approximated as an integral over an ensemble of $\lfloor N_0/S \rfloor \times \lfloor N_0/S \rfloor$ matrices $Q$ and $\bar Q$: $Z_{\textrm {CNN}} = \int \mathcal D Q \mathcal D \bar Q  e^{-N_c/2 S_{\textrm{CNN}}(Q,\bar Q)}$ (see Appendices for more details on the derivation), where the effective action is given by:
\begin{align}
S_{\textrm{CNN}}({Q},{\bar{Q}}) \equiv & - \textrm{Tr}\, Q\bar{Q} +\textrm{Tr} \log  (\mathbb{1}+{Q}) + \nonumber \\
& + \frac{\alpha_c}{P} \textrm{Tr} \log \beta \Big(\frac{\mathbb{1}}{\beta} +  {K}_{\textrm{CNN}}^{(\mathrm{R})}\Big) + \nonumber\\
&  + \frac{\alpha_c}{P} {y}^\top  \Big(\frac{\mathbb{1}}{\beta} +  {K}_{\textrm{CNN}}^{(\mathrm{R})} \Big)^{-1}   {y} \,.
\label{CNNaction}
\end{align}
The trace in the first two terms is over $\lfloor N_0/S \rfloor \times \lfloor N_0/S \rfloor$ operators, whereas the renormalized kernel $K_{\textrm{CNN}}^{(\mathrm{R})}$ is a $P \times P$ matrix and therefore the trace in the third term and the scalar products with the output labels lie in a $P$-dimensional vectorial space. The matrix elements of the renormalized kernel are given by:
\begin{equation}
\left[{K}_{\textrm{CNN}}^{(\mathrm{R})} ({\bar Q})\right]_{\mu\nu} = \frac{1}{\lambda_1\lfloor N_0/S \rfloor}\sum_{ij =1} ^{\lfloor N_0/S \rfloor} \bar Q_{ij} K_{\mu\nu}^{ij}\,,
\label{KR_CNN}
\end{equation}
where the kernel $K_{\mu\nu}^{ij}$ is given in terms of the elements of the following local covariance matrix 
\begin{equation}
C_{\mu\nu}^{ij} = \frac{1}{\lambda_0 M} \sum_{m = -\lfloor M/2\rfloor}^{\lfloor M/2\rfloor} x^\mu_{S i+m} x^\nu_{S j+m},
\label{local_covariance_matrix}
\end{equation}
with $\lambda_0$ being the hidden layer Gaussian prior, via the same functional relations of the FC case:
\begin{align}
K_{\mu\nu}^{ij} &= \int d^2 t\, \mathcal N_{ t} \left( 0,  \tilde C_{\mu\nu}^{ij}\right) \sigma (t_1) \sigma(t_2)\,, \nonumber \\
\tilde C_{\mu\nu}^{ij} &= \begin{pmatrix}
C_{\mu\mu}^{ii} & C_{\mu\nu}^{ij} \\
C_{\mu\nu}^{ij} & C_{\nu\nu}^{jj}\,.
\end{pmatrix} \,.
\label{CNN_kernel_definition}
\end{align}
It is worth noticing that such a local kernel has been already found in the seminal work on the Gaussian process limit of infinite-width CNNs \cite{novak2019bayesian}. However, one finds that in the limit of infinitely many channels $N_c \gg P$, the CNN evolves according to the following averaged kernel:
\begin{equation}
\bar K_{\mu \nu} = \frac{1}{\lambda_1\lfloor N_0/S\rfloor}\sum_{i=1}^{\lfloor N_0/S\rfloor} K_{\mu \nu}^{ii}\,.
\label{aGPCNN}
\end{equation}
In our effective action framework we recover this result by solving the saddle-point equations for the matrix $\bar Q$ in the limit of $\alpha_c \to 0$, where one finds $\bar Q_{ij} = \delta_{ij}$ and only the diagonal components of the local kernel contribute to the prediction. 

Note that the derivation of this result is based on a Gaussian equivalence that is informally justified using a general class of central limit theorems (such as the Breuer-Major theorem in \cite{ariosto2022statistical}). This suggests that the dimension of the input $N_0$ should scale to infinity at the same rate as the size of the dataset $P$. If this is the case, as it was in \cite{ariosto2022statistical}, the stride $S$ should be chosen to scale extensively in $N_0$, so that the final effective action will depend on a finite number of order parameters.

\section{A simple mechanism for feature learning in finite-width CNNs: Local Kernel renormalization} 

\begin{table}
 \begin{tabular}{|c | c | c | c|} 
 \hline
 Architecture & Kernel type & IW Kernel & Renormalized Kernel \\ [0.5ex] 
 \hline\hline
 FC & $K_{\mu \nu}$ & $K_{\mu \nu}$ & $\bar Q K_{\mu \nu}$ \\
 \hline
 LCN & $K_{\mu \nu}^{i i}$ & $ \sum_{i}  K^{i i}_{\mu \nu}$ & $\sum_{i} \bar Q_{i} K^{i i}_{\mu \nu}$ \\
 \hline
 CNN & $K_{\mu \nu}^{i j}$ & $ \sum_{i}  K^{i i}_{\mu \nu}$ & $ \sum_{i j} \bar Q_{i j} K^{i j}_{\mu \nu}$ \\ [1ex] 
 \hline
 \end{tabular}
\caption{\textbf{Summary of the analytical results for one hidden layer networks}. We identify three different types of kernels in the Bayesian effective action for FC, LC and convolutional networks: a global kernel in the FC case, a local diagonal kernel in LCNs and a local (with additional non-diagonal components) in CNNs. The infinite-width kernel that describes LCNs and CNNs is the same, as shown in \cite{novak2019bayesian}. In the proportional limit, the renormalization of the FC kernel is scalar, i.e. the NNGP kernel is only globally rescaled by a data-dependent number $\bar Q$. In LCNs, we can adjust a vector of components $\bar Q_i$, which determines how much the self-correlations of the $i$-th patch of the diagonal local kernel do contribute to the final prediction. In CNNs, due to locality and weight-sharing, we can fine-tune a full matrix $\bar Q_{ij}$, which controls how much the local correlations of patches $i$ and $j$ will enter the effective kernel learned by the network after training. This is what we call \emph{local kernel renormalization}.}
\label{Table1}
\end{table}

Let us now analyze the saddle-point equations deriving from Eq. \eqref{CNNaction}. We can gain analytical insight in the zero temperature limit, where we find the following matrix equations that determine $Q$ and $\bar Q$:
\begin{align}
&\bar Q_{ij} = (\mathbb{1} + {Q})^{-1}_{ij}\\
&Q_{ij} = \frac{\alpha_c}{P} \textrm{Tr} \Big(K^{ij} \left[K_{\textrm{CNN}}^{(\mathrm{R})}(\bar Q)\right]^{-1}\Big) + \nonumber \\ 
&-\frac{\alpha_c}{P} {y}^T \Big( \left[K_{\textrm{CNN}}^{(\mathrm{R})}(\bar Q)\right]^{-1}  K^{ij} \left[K_{\textrm{CNN}}^{(\mathrm{R})}(\bar Q)\right]^{-1}\Big) {y}\,
\label{CNN_SP}
\end{align}
where the shortcut $K^{ij}$ has been introduced to indicate the partial trace over indices $(\mu, \nu)$ in the local kernel defined in Eq. \eqref{KR_CNN}. More explicitly, given a matrix $O$ with elements $O_{\mu \nu}$, the partial trace is defined as $\textrm{Tr} (K^{ij} O) := \sum_{\mu\nu=1}^P K^{ij}_{\mu\nu} O_{\mu\nu}$. The exact solution of these matrix equations cannot be obtained in closed form. However, it is possible to compute it perturbatively around the infinite-width limit $\alpha_c \to 0$. Parametrizing the solution of the first equation as $\bar Q_{ij} = \delta_{ij} + \alpha_c  \delta \bar Q_{ij} + o(\alpha_c)$, we find the simple result:
\begin{equation}
\delta \bar Q_{ij}= \frac{1}{P} \tilde y^T K^{ij} \tilde y -  \frac{1}{P}\textrm{Tr} \left( K^{ij} \bar K^{-1} \right)\,,
\label{Qp}
\end{equation}  
where $\tilde y^\mu = \sum_{\mu=1}^P \bar K^{-1}_{\mu\nu} y^\mu$ and $\bar K$ is the infinite-width CNN averaged kernel defined in Eq. \eqref{aGPCNN}. We now provide a physical interpretation of this finding.

Our claim is that the matrix $\bar Q$ provides a compact description of feature learning in the CNN model under consideration: the feature matrix elements $\bar Q_{ij}$ that are optimized during training are in fact in one-to-one correspondence with the combinations of pairs of the $\lfloor N_0/S \rfloor$ patches of the local covariance matrix, which is defined from the trainset elements $x^{\mu}$. From the first term of the r.h.s. of Eq. \eqref{Qp} we can notice that whenever the local kernel of the CNN at the spatial locations $(i,j)$ will have a significant overlap with the effective label vector $\tilde y$, the element $\bar Q_{ij}$ of the feature matrix $\bar Q$ will differ from one. In other words, this is a measure of the relative importance of the pairwise spatial correlations in the input dataset wrt the labels. From Eq. \eqref{KR_CNN}, we can now interpret the matrix $\bar Q$ as a feature, data-dependent matrix that tells us how much a given component of the local kernel contribute to the renormalized kernel $K_{\textrm{CNN}}^{(\mathrm{R})}$.

The local kernel renormalization that takes place in this CNN model in the proportional limit is completely different from the trivial global one that occurs in the corresponding FC one hidden layer network. At this point one can wonder whether both the ingredients in CNNs design, local connectivity and weight sharing, are needed to observe the phenomenon of local kernel renormalization. The answer is affirmative: local kernel renormalization does not occur in a shallow network with local connectivity only, as we explicitly check in the appendix. In that case, only the diagonal components of the local kernel enter the effective action, and the network is not sensible to spatial correlations between different patches (see also Table \ref{Table1} for a summary of the analytical results).

\section{Empirical evidence for global and local kernel renormalization}

\begin{figure*}
    \centering
    \includegraphics[width = 1.\textwidth]{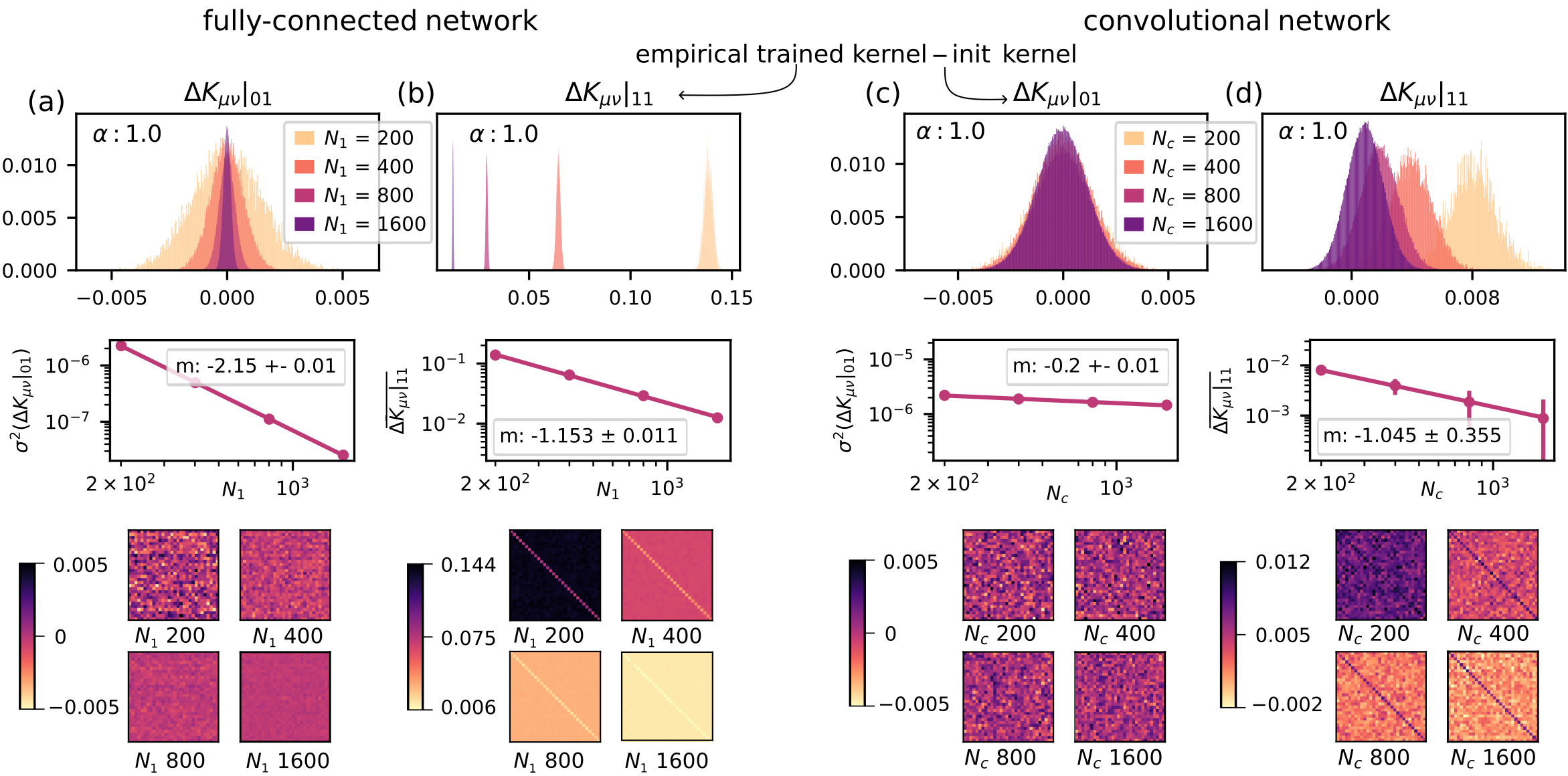}
    \caption{\textbf{Predicting the effect of global and local kernel renormalization on the internal representations of FCNs and CNNs. A finite-size scaling analysis.} 
 Here we show the results of numerical experiments performed in order to check the predictions of Eqs. \eqref{FCN_sim_matrix_main}, \eqref{CNN_sim_matrix_main} for the difference $\Delta K_{\mu\nu}$ between the similarity matrix after and before training. In particular, we train architectures with a single hidden FC/$1d$-convolutional layer with tanh activation (see details in Appendix \ref{num_exp}) on a learning task with random Gaussian inputs and zero/one labels given by a linear teacher. We perform a finite-size scaling analysis by keeping $\alpha_1 = 1$ and choosing increasing values of $P, N_1$. In panel (a) we observe that the variance of the empirical distribution of the matrix elements of the zero-one block $\Delta K_{\mu\nu}^{\textrm{FC}}|_{01}$ rapidly converges to zero as $N_1$ grows and $\alpha_1$ is kept fixed to one, in agreement with the prediction of Eq. \eqref{FCN_sim_matrix_main} (top). In particular, the variance goes to zero as $1/N_1^{2}$ (middle). The fact that the similarity matrices after and before training are more and more similar in the thermodynamic limit is also clearly visible in the explicit array plot of a subsample of the zero-one block $\Delta K_{\mu\nu}^{\textrm{FC}}|_{01}$ (bottom). In panel (b) we display the histogram of the distribution of the matrix elements of the one-one block $\Delta K_{\mu\nu}^{\textrm{FC}}|_{11}$ (top). As predicted by Eq. \eqref{FCN_sim_matrix_main}, the distribution shrinks while drifting towards zero at a rate $\sim 1/N_1$ (middle). Overall, (a) and (b) represent a clear indication that the change in the after-training internal representations in one hidden layer FCNs is a finite-size effect. In (c) and (d) we repeat the same experiment for an architecture with a $1d$-convolutional hidden layer. The striking difference wrt the FC case is that the variance of the empirical distribution of the matrix elements $\Delta K_{\mu\nu}^{\textrm{CNN}}$ does not shrink to zero as $N_c$ grows and $\alpha_c$ is kept fixed to one, as one can also appreciate by directly looking at the array plot of a subsample of the block matrices $\Delta K_{\mu\nu}^{\textrm{CNN}}|_{01}$ and $\Delta K_{\mu\nu}^{\textrm{CNN}}|_{11}$.
    }
    \label{fig:deltaK}
\end{figure*}

Extracting the renormalized kernel from the measure of a physical observable is not straightforward, and we have no way, at the moment, to directly access it in numerical experiments. One indirect experimental blueprint of the form of kernel renormalization at finite width in different architectures is given by the \textit{similarity matrix} of the internal representations before and after training. We define this observable for both FCNs and CNNs as: 
\begin{equation}
     O_{\mu\nu} \equiv \frac{1}{N_1} \sum_{i=1}^{N_1} \sigma\left(h^\mu_i\right) \sigma\left(h^\nu_i\right) \,,
\label{similarity_matrix}
\end{equation}   
 where $N_1$ denotes the number of neurons in the last layer (for CNNs we have $N_1 = N_c\lfloor N_0/S \rfloor$). We can track the effect of training by taking the difference between the similarity matrix at initialization, which is by definition the NNGP kernel (up to the Gaussian prior $\lambda_1$), and the same quantity after training: $\Delta K_{\mu \nu} \equiv \langle O_{\mu \nu} \rangle - K_{\mu \nu } $, where the average is done over the Gibbs ensemble of the weights. We analytically compute this observable for FCNs and CNNs, as shown in appendix \ref{6:Predicting observable: the similarity matrix}. The final result in the two cases respectively reads (at zero temperature):
\begin{align}
&\Delta K^{\textrm{FCN}}_{\mu\nu} = -\frac{1}{N_1} \left[ K_{\mu\nu} - \frac{\lambda_1}{\bar Q}y^\mu y^\nu \right];
\label{FCN_sim_matrix_main}  \\
&\Delta K^{\textrm{CNN}}_{\mu\nu} =  - \frac{1}{\lambda_1N_1}\sum_{ij,\lambda\rho = 1}^{\lfloor N_0/S\rfloor, P}\bar{Q}_{ij}P^{ij}_{\mu\lambda\nu\rho}\nonumber \\
& \, \, \, \times\left[\left(K^{(\mathrm{R})}_{\textrm{CNN}}\right)^{-1}_{\lambda\rho}  - \sum_{\epsilon\omega=1}^{P}\left(K^{(\mathrm{R})}_{\textrm{CNN}}\right)^{-1}_{\lambda\epsilon}  \left(K^{(\mathrm{R})}_{\textrm{CNN}}\right)^{-1}_{\rho\omega}y^\epsilon y^\omega\right] ,  
\label{CNN_sim_matrix_main}
\end{align}
where $P^{ij}_{\mu\lambda\nu\rho} \equiv \frac{1}{2} \sum_{k}\Big[K^{ik}_{\mu\lambda}K^{kj}_{\nu\rho} + K^{ki}_{\mu\lambda}K^{jk}_{\nu\rho}\Big]$. Let us now highlight a few interesting physical implications that directly follow from these formulas and can be checked with numerical experiments:
\begin{enumerate}[(i)]
    \item the difference between the trained and untrained similarity matrix $\Delta K^{\textrm{FCN}}_{\mu\nu}$ converges to zero in the thermodynamic limit $P, N_1\rightarrow \infty$ at fixed $\alpha_1=P/N_1$, as long as the terms in the square brackets are of order $ o(N_1)$. This means that in the proportional regime under consideration the effect of training on the internal representations is just a finite-size correction $O(1/N_1)$ and the trained similarity matrix stays very close to its value at initialization. In particular, we expect that the empirical distribution of the elements of the matrix $\Delta K^{\textrm{FCN}}_{\mu\nu}$ will be increasingly peaked around zero in finite-size scaling experiments where we consider increasing values of $P$ and $N_1$ keeping their ratio $\alpha_1=P/N_1$ fixed;
    \item by choosing a learning task with labels $y^\mu = 0,1$ and input patterns $x^\mu$ such that the matrix elements of the kernel $K_{\mu\nu}$ are on average centered in zero, the distribution of the matrix elements of the block $\Delta K^{\textrm{FCN}}_{\mu\nu}|_{11}$ (corresponding to the subset of training patterns with label one) should drift towards zero at a rate $1/N_1$, if one performs experiments at constant $\alpha_1$, while increasing $P$ and $N_1$. Interestingly, the remaining blocks of the matrix $\Delta K^{\textrm{FCN}}_{\mu\nu}$ should not be affected by this drift;
    \item the difference between the trained and untrained convolutional similarity matrix $\Delta K^{\textrm{CNN}}_{\mu\nu}$ displays a different behavior, due to local kernel renormalization. For instance, even if we restrict our analysis to the learning setting outlined in bullet (ii), the elements of the block $\Delta K^{\textrm{CNN}}_{\mu\nu}|_{01}$ should not identically converge to zero in the thermodynamic proportional limit. A straightforward scaling analysis in fact shows that the contribution due to the term in the square bracket does not vanish in the proportional limit.
\end{enumerate}
In Fig. \ref{fig:deltaK} we present the results of a finite-size scaling analysis to check the aforementioned predictions, where we trained one hidden layer FCNs and CNNs ($1d$ convolutions) on a synthetic dataset composed of random Gaussian patterns whose labels are given by a linear teacher function, i.e. $y = \frac{1}{2}\left[1+\textrm{sign}(t \cdot x)\right]$, where $t $ is an $N_0$-dimensional vector with unitary entries. Overall, the fact that the change in the internal representations of FCNs is a finite-size effect is confirmed by the experiments. On the other hand, the same numerical simulations with CNNs clearly identify a genuine change in the internal representations that occurs also in the thermodynamic proportional limit. We note that such a clean signature of the differences between FCNs and CNNs derive also from our choice of working close to zero temperature (where the architectures can precisely fit the training data). Equations at finite $T$ are more involved, and measuring the indirect effects of global and local kernel renormalization in this case is more difficult. Interestingly, even though our theory should be valid only in the Bayesian setting, it turns out to be also predictive for a learning algorithm that does not explicitly sample from the Gibbs posterior distribution: the aforementioned experiments were carried out using gradient descent and the so-called ADAM optimizer \cite{adam} (see Appendix \ref{num_exp} for more details).

Another empirical evidence for the local renormalization in CNNs can be found analyzing the bias/variance decomposition of the generalization error over a new example $(\mathbf x^0, y^0)$:
\begin{equation}
\begin{split}
\braket{\epsilon_\text{g}(\mathbf{x}^0,y^0)} &= (y^0-\Gamma)^2+\sigma^2
\end{split}
\label{methods:eq:gen_err0deep}
\end{equation} 
with:
\begin{align}
&\Gamma =\sum_{\mu\nu=1}^{P} \kappa^{\mathrm{(R)}}_\mu  \left(\frac{\mathbb{1}}{\beta}+K^{\mathrm{(R)}} ( \{\bar Q\}) \right)^{-1}_{\mu\nu}\; y_\nu, \label{gamma}\\
& \sigma^2 = \kappa^{(\mathrm{R})}_{0} -\sum_{\mu\nu=1}^{P} \kappa^{(\mathrm{R})}_{\mu} \left(\frac{\mathbb{1}}{\beta} + K^{(\mathrm{R})}( \{\bar Q\})\right)^{-1}_{\mu\nu}\kappa^{(\mathrm{R})}_{\nu}\label{sigma2}
\end{align}
These expressions are valid for FCNs, and can be easily extended to CNNs with the substitution: 
\begin{equation}
    K^{(\mathrm{R})} \to K^{(\mathrm{R})}_{\mathrm{CNN}} \quad     \kappa^{(\mathrm{R})} \to \kappa^{(\mathrm{R})}_{\mathrm{CNN}}\,,
\end{equation}
where the renormalized kernel matrices
$K^{(\mathrm{R})}$ and  $K^{(\mathrm{R})}_{\mathrm{CNN}}$ are given respectively in equations \eqref{K_LQ} and \eqref{KR_CNN}. We recall the definition of the additional kernel integrals that depend on the new example $x^0$, that is $\kappa_\mu^{(\mathrm{R})}$, where the index $\mu = 0, \ldots ,P $ covers both the dataset and the new test element $x^0$:
\begin{align}
&\kappa_\mu = \int \diff t \mathcal{N}_t \left(0, \tilde C_\mu \right) \sigma(t_1) \sigma (t_2)\,, \label{kmu}
%\\&\kappa_0 = \int \frac{dt}{\sqrt{2\pi C_{00}}} e^{-\frac{t^2}{2 C_{00}}}  \sigma(t)^2\,, \label{k0}
\end{align}
where:
\begin{align}
&\begin{gathered}
\tilde C_\mu = \begin{pmatrix} C_{\mu\mu} & C_{\mu0}\\ C_{\mu0} & C_{00}\end{pmatrix}\,, \, \, C_{\mu 0} =  \frac{x^\mu \cdot x^0}{\lambda_0 N_0} \,,
% \, \,  C_{0 0} =  \frac{\Vert x^0 \Vert}{\lambda_0 N_0} \,.
 \end{gathered}
\end{align}

In the CNN case, these quantities are computed starting from the local covariance matrix $ C^{ij}_{\mu \nu}$ defined in Eq. \eqref{local_covariance_matrix}. These terms undergo the same type of renormalization as the correspondent kernel matrices (see table \ref{Table1}), that is: 
\begin{align}
&{\kappa}^{(\mathrm{R})}_{\mu} = \frac{\bar Q}{\lambda_1} \kappa_{\mu}\,, \\
&\left[{\kappa}_{\textrm{CNN}}^{(\mathrm{R})} \right]_{\mu} = \frac{1}{\lambda_1\lfloor N_0/S \rfloor}\sum_{ij =1} ^{\lfloor N_0/S \rfloor} \bar Q_{ij} \kappa_{\mu}^{ij}\,,
\label{kappa_R_CNN}
\end{align}

The $\beta \to \infty$ limit of Eq. \eqref{gamma} yields different behaviors for the bias in the two cases:
\begin{align}
&\hphantom{^2}\Gamma^{\mathrm{FC}} =\sum_{\mu\nu=1}^P \kappa_\mu K ^{-1}_{\mu\nu}\; y_\nu \, , \label{FC_bias} \\
&\hphantom{^2}\Gamma^{\mathrm{CNN}} =\sum_{\mu\nu=1}^P \left[\kappa^{\mathrm{(R)}}_{\mathrm{CNN}}\right]_\mu \left( K^{\mathrm{(R)}}_{\mathrm{CNN}} ( \{\bar Q\}) \right)^{-1}_{\mu\nu}\; y_\nu,
\label{CONV_bias}
\end{align}
From these expressions one can make two observations: (i) as already discussed in \cite{ariosto2022statistical}, $\Gamma^{\textrm{FC}}$ at finite width is independent of $\bar Q$, and therefore the bias is identical to the infinite-width case for any $N_1$. (ii) In CNNs, on the other hand, the dependence of the bias from the renormalized Kernel does not disappear, and neither does the dependence on the width $N_1$.
%\st{ This feature allows to overcome the FCN limitation, allowing the test loss to be a non-monotonic function of $N_1$. }

In the large Gaussian prior regime $\lambda_1 \gg 1 $, the previous observations can be made into straightforward predictions that can be tested with numerical experiments. It is easy to check that the variance is vanishing in this regime, since it is always proportional to a term $1/\lambda_1$ both in FCNs and CNNs. Moreover, when $\lambda_1 \gg 1$, the analytical criterion found in \cite{ariosto2022statistical} to predict whether a finite-width FCN will outperform its infinite-width counterpart, is never satisfied. This is a special case of the general considerations already presented in section \ref{sec:fc_cannot_outperform}.
To summarize, in this regime we expect that (i) the finite FC network can not outperform its infinite-width counterpart. (ii) On the contrary, we expect CNNs to possibly beat their infinite-width performance, since the bias can be reduced through the optimization of the feature matrix $Q^{ij}$. 

\begin{figure}
    \centering
    \includegraphics[width = 0.49\textwidth]{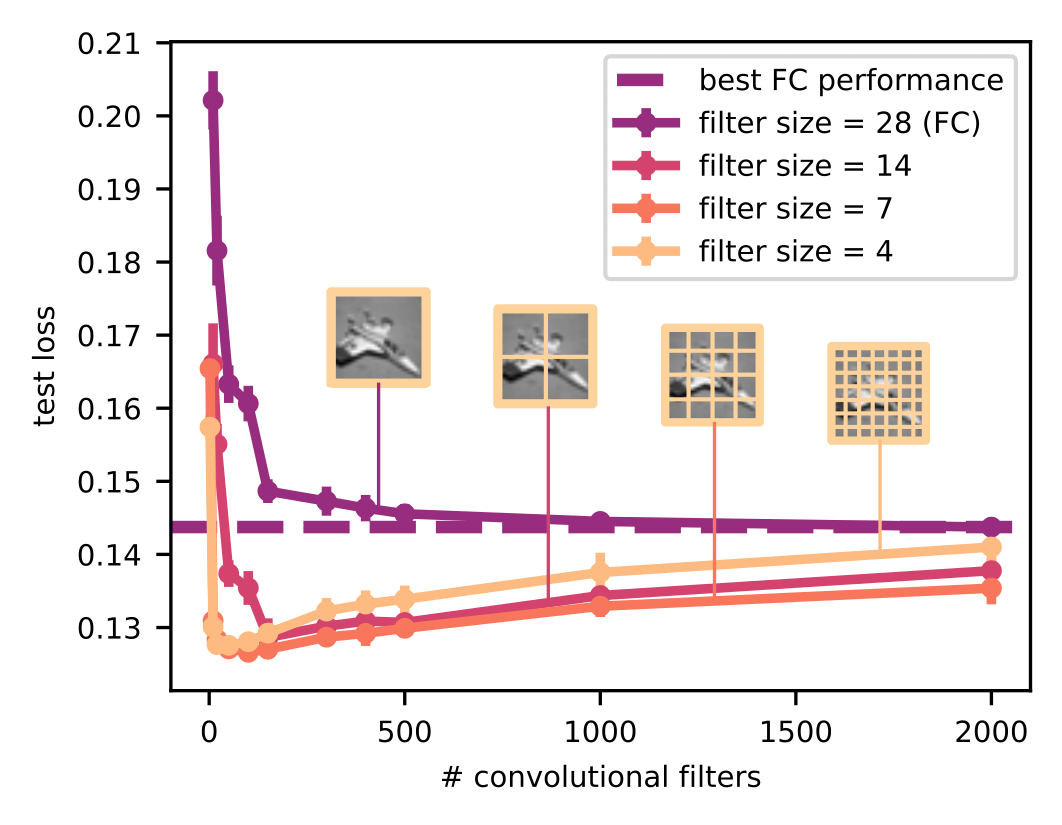}
        \caption{\textbf{Test loss vs number of $2d$ convolutional filters for one hidden layer CNNs with different filter sizes $M$.} All the networks are trained on the same binary regression task with $P=500$ examples from two classes of the CIFAR10 dataset, "cars" and "planes", respectively labeled $0$ and $1$. Error bars represent one standard deviation.  The images are gray-scaled and coarse-grained to $N_0 = 28 \times 28$. We choose the filter size $M$ to be an integer divisor of $28$, and the filters to be non overlapping (taking stride $S = M$). A sketch of how the images are parsed by filters is reported for each network. With this choice, the FC network is retrieved setting $M = 28$. The nets are trained with a zero temperature Langevin dynamics (see appendix REF), and with large Gaussian priors $\lambda_1$. In this bias-dominated regime, we observe that: (i) in FCNs, the best possible performance is the infinite-width one. Here the variance is monotonically decreasing with the width $N_c$, while the bias is a constant. Therefore, the test loss approaches its minimum monotonically from above, as predicted by our theory. (ii) finite width CNNs can outperform their infinite-width counterpart. The local kernel renormalization mechanism allows the CNN to optimize both bias and variance at finite width, making it even convenient to have a small number of filters (the minimum is reached for $N_c \sim 50$). Eventually, the CNNs reach their asymptotic performance from below. }
    \label{fig:conv2d}
\end{figure}

To test these predictions, we performed numerical simulations with $2d$ CNNs trained on CIFAR10 examples, close to the zero-temperature limit and in the bias-dominated setting where we let the Gaussian prior $\lambda_1$ be large. %Note that the large $\lambda_1$ regime implies, as discussed in \cite{ariosto2022statistical} and generalized in REFSECTION, that the IW performance is the best possible for a FCN in the proportional regime. 
%On the contrary, we expect CNNS to possibly beat their IW performance, since the bias can be reduced through the optimization of the feature matrix $Q^{ij}$.
As shown in Fig. \ref{fig:conv2d}, the test loss of the FCN is a monotonically decreasing function of the hidden layer size $N_1$ and it does reach its asymptotic (best) performance already for $\alpha \sim 1$. Compatibly to local kernel renormalization, CNNs with finite number of channels best their infinite-width counterpart, already for large filter sizes ($M = 14, 7$) and ultimately approach the infinite-width limit performance from below. 
%It is worth stressing that we do not expect that local kernel renormalization will lead to improved generalization performance for generic learning tasks, but only in those cases where the training patterns exhibit local spatial correlations, as it occurs in computer vision problems.

%that allows both bias and variance to be optimized at finite width through the optimization of the feature matrix $Q^{ij}$.  

\section{Discussion} 

In this work we have derived an effective action in the proportional limit for CNNs and LCNs with one hidden layer, and we have extensively compared these results with the FC case found in \cite{ariosto2022statistical}.

We have highlighted a striking difference in the kernel renormalization mechanism that the three different architectures undergo at finite width, summarized in Table \ref{Table1}, and we have suggested a mechanism for feature learning (local kernel renormalization) that can take place in CNNs, but not in architectures with LC or FC layers alone.

We have made a preliminary study on the effect of global and local kernel renormalization at the level of the internal representations of trained networks, which is in agreement with our predictions. We have also investigated the behavior of the generalization loss of one hidden layer networks in a controlled Bayesian setting where the performance is dominated by the bias, which confirms the observation that, in this regime, (i) finite-width one hidden layer FC networks cannot outperform their infinite-width counterpart; (ii) CNNs with finite number of channels can improve the infinite-width performance, which is in qualitative agreement with our local kernel renormalization hypothesis for the bias in Eq. \eqref{CONV_bias}.

We conclude by pointing out a number of possible follow-ups of this work: (i) it would be interesting to extend our framework to $L$ convolutional layers, following the same strategy proposed for FC layers in \cite{ariosto2022statistical}; (ii) our current approach cannot cover the mean-field regime, where the feature learning mechanism for FCNs and CNNs may quantitatively change. We think that the formalism presented in \cite{seroussi2023natcomm} might be more suited to investigate finite-width effects in this case; (iii) we are writing a detailed open-source numerical routine to extensively check the theoretical predictions of our effective theory for FCNs, LCNs and CNNs, which will also serve to point out the limitations of the present framework.

%We are now in the position to conjecture a general parametrization for feature learning in deep CNNs with $L$ layers. For each convolutional layer $\ell$ there will be GP local kernels and a different feature matrix: 
%\begin{equation}
%(K_{\ell})_{\mu\nu}^{i_\ell j_\ell} \quad Q^{(\ell)}_{i_{\ell} j_{\ell}} \quad i_{\ell},  j_{\ell} = 1, \dots,N_\ell\,.
%\end{equation}
%Local GP kernels will satisfy simple recurrence relations exacly equivalent to those found for FC networks in the infinite-width limit. The only source of complication will be to determine the coarse-graining operator associated to stride or pooling that allow the change in the dimension of the feature map from $N_\ell$ to $N_{\ell+1}$ at each layer. Feature learning will correspond to non-trivial and data-dependent solutions for each feature matrix $Q^{(\ell)}$ and it will determine the form of the renormalized final kernel $K_L^{(R)}$ describing the finite-width deep network at the end of the bayesian training.

\emph{Acknowledgements}.-- R.B. and P.R. are supported by $\#$NEXTGENERATIONEU (NGEU) and funded by the Ministry of University and Research (MUR), National Recovery and Resilience Plan (NRRP), project MNESYS (PE0000006) ``A Multiscale integrated approach to the study of the nervous system in health and disease” (DN. 1553 11.10.2022). The authors thank F. Cagnetta for giving feedback on a preliminary version of this manuscript and for pointing out interesting related works, G. Naveh, Z. Ringel and I. Seroussi for discussions and clarifications on their results in Refs. \cite{seroussi2023natcomm, NEURIPS2021_b24d2101} and P. Baglioni, M. Gherardi and M. Pastore for discussions.

\bibliography{biblio}

\begin{thebibliography}{55}%
\makeatletter
\providecommand \@ifxundefined [1]{%
 \@ifx{#1\undefined}
}%
\providecommand \@ifnum [1]{%
 \ifnum #1\expandafter \@firstoftwo
 \else \expandafter \@secondoftwo
 \fi
}%
\providecommand \@ifx [1]{%
 \ifx #1\expandafter \@firstoftwo
 \else \expandafter \@secondoftwo
 \fi
}%
\providecommand \natexlab [1]{#1}%
\providecommand \enquote  [1]{``#1''}%
\providecommand \bibnamefont  [1]{#1}%
\providecommand \bibfnamefont [1]{#1}%
\providecommand \citenamefont [1]{#1}%
\providecommand \href@noop [0]{\@secondoftwo}%
\providecommand \href [0]{\begingroup \@sanitize@url \@href}%
\providecommand \@href[1]{\@@startlink{#1}\@@href}%
\providecommand \@@href[1]{\endgroup#1\@@endlink}%
\providecommand \@sanitize@url [0]{\catcode `\\12\catcode `\$12\catcode
  `\&12\catcode `\#12\catcode `\^12\catcode `\_12\catcode `\%12\relax}%
\providecommand \@@startlink[1]{}%
\providecommand \@@endlink[0]{}%
\providecommand \url  [0]{\begingroup\@sanitize@url \@url }%
\providecommand \@url [1]{\endgroup\@href {#1}{\urlprefix }}%
\providecommand \urlprefix  [0]{URL }%
\providecommand \Eprint [0]{\href }%
\providecommand \doibase [0]{https://doi.org/}%
\providecommand \selectlanguage [0]{\@gobble}%
\providecommand \bibinfo  [0]{\@secondoftwo}%
\providecommand \bibfield  [0]{\@secondoftwo}%
\providecommand \translation [1]{[#1]}%
\providecommand \BibitemOpen [0]{}%
\providecommand \bibitemStop [0]{}%
\providecommand \bibitemNoStop [0]{.\EOS\space}%
\providecommand \EOS [0]{\spacefactor3000\relax}%
\providecommand \BibitemShut  [1]{\csname bibitem#1\endcsname}%
\let\auto@bib@innerbib\@empty
%</preamble>
\bibitem [{\citenamefont {Goodfellow}\ \emph {et~al.}(2016)\citenamefont
  {Goodfellow}, \citenamefont {Bengio},\ and\ \citenamefont
  {Courville}}]{GoodfellowBook}%
  \BibitemOpen
  \bibfield  {author} {\bibinfo {author} {\bibfnamefont {I.}~\bibnamefont
  {Goodfellow}}, \bibinfo {author} {\bibfnamefont {Y.}~\bibnamefont {Bengio}},\
  and\ \bibinfo {author} {\bibfnamefont {A.}~\bibnamefont {Courville}},\ }\href
  {http://www.deeplearningbook.org} {\emph {\bibinfo {title} {Deep Learning}}}\
  (\bibinfo  {publisher} {MIT Press},\ \bibinfo {year} {2016})\BibitemShut
  {NoStop}%
\bibitem [{\citenamefont {Bengio}\ \emph {et~al.}(2013)\citenamefont {Bengio},
  \citenamefont {Courville},\ and\ \citenamefont {Vincent}}]{6472238}%
  \BibitemOpen
  \bibfield  {author} {\bibinfo {author} {\bibfnamefont {Y.}~\bibnamefont
  {Bengio}}, \bibinfo {author} {\bibfnamefont {A.}~\bibnamefont {Courville}},\
  and\ \bibinfo {author} {\bibfnamefont {P.}~\bibnamefont {Vincent}},\
  }\bibfield  {title} {\bibinfo {title} {Representation learning: A review and
  new perspectives},\ }\href {https://doi.org/10.1109/TPAMI.2013.50} {\bibfield
   {journal} {\bibinfo  {journal} {IEEE Transactions on Pattern Analysis and
  Machine Intelligence}\ }\textbf {\bibinfo {volume} {35}},\ \bibinfo {pages}
  {1798} (\bibinfo {year} {2013})}\BibitemShut {NoStop}%
\bibitem [{\citenamefont {Yu}\ \emph {et~al.}(2013)\citenamefont {Yu},
  \citenamefont {Seltzer}, \citenamefont {Li}, \citenamefont {Huang},\ and\
  \citenamefont {Seide}}]{yu2013feature}%
  \BibitemOpen
  \bibfield  {author} {\bibinfo {author} {\bibfnamefont {D.}~\bibnamefont
  {Yu}}, \bibinfo {author} {\bibfnamefont {M.~L.}\ \bibnamefont {Seltzer}},
  \bibinfo {author} {\bibfnamefont {J.}~\bibnamefont {Li}}, \bibinfo {author}
  {\bibfnamefont {J.-T.}\ \bibnamefont {Huang}},\ and\ \bibinfo {author}
  {\bibfnamefont {F.}~\bibnamefont {Seide}},\ }\bibfield  {title} {\bibinfo
  {title} {Feature learning in deep neural networks-studies on speech
  recognition tasks},\ }\href@noop {} {\bibfield  {journal} {\bibinfo
  {journal} {arXiv preprint arXiv:1301.3605}\ } (\bibinfo {year}
  {2013})}\BibitemShut {NoStop}%
\bibitem [{\citenamefont {Neal}(1996)}]{Neal}%
  \BibitemOpen
  \bibfield  {author} {\bibinfo {author} {\bibfnamefont {R.~M.}\ \bibnamefont
  {Neal}},\ }\bibinfo {title} {Priors for infinite networks},\ in\ \href
  {https://doi.org/10.1007/978-1-4612-0745-0_2} {\emph {\bibinfo {booktitle}
  {Bayesian Learning for Neural Networks}}}\ (\bibinfo  {publisher} {Springer
  New York},\ \bibinfo {address} {New York, NY},\ \bibinfo {year} {1996})\ pp.\
  \bibinfo {pages} {29--53}\BibitemShut {NoStop}%
\bibitem [{\citenamefont {Williams}(1996)}]{NIPS1996_ae5e3ce4}%
  \BibitemOpen
  \bibfield  {author} {\bibinfo {author} {\bibfnamefont {C.}~\bibnamefont
  {Williams}},\ }\bibfield  {title} {\bibinfo {title} {Computing with infinite
  networks},\ }in\ \href
  {https://proceedings.neurips.cc/paper/1996/file/ae5e3ce40e0404a45ecacaaf05e5f735-Paper.pdf}
  {\emph {\bibinfo {booktitle} {Advances in Neural Information Processing
  Systems}}},\ Vol.~\bibinfo {volume} {9},\ \bibinfo {editor} {edited by\
  \bibinfo {editor} {\bibfnamefont {M.}~\bibnamefont {Mozer}}, \bibinfo
  {editor} {\bibfnamefont {M.}~\bibnamefont {Jordan}},\ and\ \bibinfo {editor}
  {\bibfnamefont {T.}~\bibnamefont {Petsche}}}\ (\bibinfo  {publisher} {MIT
  Press},\ \bibinfo {year} {1996})\BibitemShut {NoStop}%
\bibitem [{\citenamefont {de~G.~Matthews}\ \emph {et~al.}(2018)\citenamefont
  {de~G.~Matthews}, \citenamefont {Hron}, \citenamefont {Rowland},
  \citenamefont {Turner},\ and\ \citenamefont {Ghahramani}}]{g.2018gaussian}%
  \BibitemOpen
  \bibfield  {author} {\bibinfo {author} {\bibfnamefont {A.~G.}\ \bibnamefont
  {de~G.~Matthews}}, \bibinfo {author} {\bibfnamefont {J.}~\bibnamefont
  {Hron}}, \bibinfo {author} {\bibfnamefont {M.}~\bibnamefont {Rowland}},
  \bibinfo {author} {\bibfnamefont {R.~E.}\ \bibnamefont {Turner}},\ and\
  \bibinfo {author} {\bibfnamefont {Z.}~\bibnamefont {Ghahramani}},\ }\bibfield
   {title} {\bibinfo {title} {Gaussian process behaviour in wide deep neural
  networks},\ }in\ \href {https://openreview.net/forum?id=H1-nGgWC-} {\emph
  {\bibinfo {booktitle} {International Conference on Learning
  Representations}}}\ (\bibinfo {year} {2018})\BibitemShut {NoStop}%
\bibitem [{\citenamefont {Lee}\ \emph {et~al.}(2018)\citenamefont {Lee},
  \citenamefont {Sohl-dickstein}, \citenamefont {Pennington}, \citenamefont
  {Novak}, \citenamefont {Schoenholz},\ and\ \citenamefont
  {Bahri}}]{LeeGaussian}%
  \BibitemOpen
  \bibfield  {author} {\bibinfo {author} {\bibfnamefont {J.}~\bibnamefont
  {Lee}}, \bibinfo {author} {\bibfnamefont {J.}~\bibnamefont {Sohl-dickstein}},
  \bibinfo {author} {\bibfnamefont {J.}~\bibnamefont {Pennington}}, \bibinfo
  {author} {\bibfnamefont {R.}~\bibnamefont {Novak}}, \bibinfo {author}
  {\bibfnamefont {S.}~\bibnamefont {Schoenholz}},\ and\ \bibinfo {author}
  {\bibfnamefont {Y.}~\bibnamefont {Bahri}},\ }\bibfield  {title} {\bibinfo
  {title} {Deep neural networks as gaussian processes},\ }in\ \href
  {https://openreview.net/forum?id=B1EA-M-0Z} {\emph {\bibinfo {booktitle}
  {International Conference on Learning Representations}}}\ (\bibinfo {year}
  {2018})\BibitemShut {NoStop}%
\bibitem [{\citenamefont {Garriga-Alonso}\ \emph {et~al.}(2019)\citenamefont
  {Garriga-Alonso}, \citenamefont {Rasmussen},\ and\ \citenamefont
  {Aitchison}}]{garriga-alonso2018deep}%
  \BibitemOpen
  \bibfield  {author} {\bibinfo {author} {\bibfnamefont {A.}~\bibnamefont
  {Garriga-Alonso}}, \bibinfo {author} {\bibfnamefont {C.~E.}\ \bibnamefont
  {Rasmussen}},\ and\ \bibinfo {author} {\bibfnamefont {L.}~\bibnamefont
  {Aitchison}},\ }\bibfield  {title} {\bibinfo {title} {Deep convolutional
  networks as shallow gaussian processes},\ }in\ \href
  {https://openreview.net/forum?id=Bklfsi0cKm} {\emph {\bibinfo {booktitle}
  {International Conference on Learning Representations}}}\ (\bibinfo {year}
  {2019})\BibitemShut {NoStop}%
\bibitem [{\citenamefont {Novak}\ \emph {et~al.}(2019)\citenamefont {Novak},
  \citenamefont {Xiao}, \citenamefont {Bahri}, \citenamefont {Lee},
  \citenamefont {Yang}, \citenamefont {Abolafia}, \citenamefont {Pennington},\
  and\ \citenamefont {Sohl-dickstein}}]{novak2019bayesian}%
  \BibitemOpen
  \bibfield  {author} {\bibinfo {author} {\bibfnamefont {R.}~\bibnamefont
  {Novak}}, \bibinfo {author} {\bibfnamefont {L.}~\bibnamefont {Xiao}},
  \bibinfo {author} {\bibfnamefont {Y.}~\bibnamefont {Bahri}}, \bibinfo
  {author} {\bibfnamefont {J.}~\bibnamefont {Lee}}, \bibinfo {author}
  {\bibfnamefont {G.}~\bibnamefont {Yang}}, \bibinfo {author} {\bibfnamefont
  {D.~A.}\ \bibnamefont {Abolafia}}, \bibinfo {author} {\bibfnamefont
  {J.}~\bibnamefont {Pennington}},\ and\ \bibinfo {author} {\bibfnamefont
  {J.}~\bibnamefont {Sohl-dickstein}},\ }\bibfield  {title} {\bibinfo {title}
  {Bayesian deep convolutional networks with many channels are gaussian
  processes},\ }in\ \href {https://openreview.net/forum?id=B1g30j0qF7} {\emph
  {\bibinfo {booktitle} {International Conference on Learning
  Representations}}}\ (\bibinfo {year} {2019})\BibitemShut {NoStop}%
\bibitem [{\citenamefont {Jacot}\ \emph {et~al.}(2018)\citenamefont {Jacot},
  \citenamefont {Gabriel},\ and\ \citenamefont {Hongler}}]{JacotNTK}%
  \BibitemOpen
  \bibfield  {author} {\bibinfo {author} {\bibfnamefont {A.}~\bibnamefont
  {Jacot}}, \bibinfo {author} {\bibfnamefont {F.}~\bibnamefont {Gabriel}},\
  and\ \bibinfo {author} {\bibfnamefont {C.}~\bibnamefont {Hongler}},\
  }\bibfield  {title} {\bibinfo {title} {Neural tangent kernel: Convergence and
  generalization in neural networks},\ }in\ \href
  {https://proceedings.neurips.cc/paper/2018/file/5a4be1fa34e62bb8a6ec6b91d2462f5a-Paper.pdf}
  {\emph {\bibinfo {booktitle} {Advances in Neural Information Processing
  Systems}}},\ Vol.~\bibinfo {volume} {31},\ \bibinfo {editor} {edited by\
  \bibinfo {editor} {\bibfnamefont {S.}~\bibnamefont {Bengio}}, \bibinfo
  {editor} {\bibfnamefont {H.}~\bibnamefont {Wallach}}, \bibinfo {editor}
  {\bibfnamefont {H.}~\bibnamefont {Larochelle}}, \bibinfo {editor}
  {\bibfnamefont {K.}~\bibnamefont {Grauman}}, \bibinfo {editor} {\bibfnamefont
  {N.}~\bibnamefont {Cesa-Bianchi}},\ and\ \bibinfo {editor} {\bibfnamefont
  {R.}~\bibnamefont {Garnett}}}\ (\bibinfo  {publisher} {Curran Associates,
  Inc.},\ \bibinfo {year} {2018})\BibitemShut {NoStop}%
\bibitem [{\citenamefont {Chizat}\ \emph {et~al.}(2019)\citenamefont {Chizat},
  \citenamefont {Oyallon},\ and\ \citenamefont {Bach}}]{ChizatLazy}%
  \BibitemOpen
  \bibfield  {author} {\bibinfo {author} {\bibfnamefont {L.}~\bibnamefont
  {Chizat}}, \bibinfo {author} {\bibfnamefont {E.}~\bibnamefont {Oyallon}},\
  and\ \bibinfo {author} {\bibfnamefont {F.}~\bibnamefont {Bach}},\ }\bibfield
  {title} {\bibinfo {title} {On lazy training in differentiable programming},\
  }in\ \href
  {https://proceedings.neurips.cc/paper/2019/file/ae614c557843b1df326cb29c57225459-Paper.pdf}
  {\emph {\bibinfo {booktitle} {Advances in Neural Information Processing
  Systems}}},\ Vol.~\bibinfo {volume} {32},\ \bibinfo {editor} {edited by\
  \bibinfo {editor} {\bibfnamefont {H.}~\bibnamefont {Wallach}}, \bibinfo
  {editor} {\bibfnamefont {H.}~\bibnamefont {Larochelle}}, \bibinfo {editor}
  {\bibfnamefont {A.}~\bibnamefont {Beygelzimer}}, \bibinfo {editor}
  {\bibfnamefont {F.}~\bibnamefont {d\textquotesingle Alch\'{e}-Buc}}, \bibinfo
  {editor} {\bibfnamefont {E.}~\bibnamefont {Fox}},\ and\ \bibinfo {editor}
  {\bibfnamefont {R.}~\bibnamefont {Garnett}}}\ (\bibinfo  {publisher} {Curran
  Associates, Inc.},\ \bibinfo {year} {2019})\BibitemShut {NoStop}%
\bibitem [{\citenamefont {Lee}\ \emph {et~al.}(2019)\citenamefont {Lee},
  \citenamefont {Xiao}, \citenamefont {Schoenholz}, \citenamefont {Bahri},
  \citenamefont {Novak}, \citenamefont {Sohl-Dickstein},\ and\ \citenamefont
  {Pennington}}]{lee2019wide}%
  \BibitemOpen
  \bibfield  {author} {\bibinfo {author} {\bibfnamefont {J.}~\bibnamefont
  {Lee}}, \bibinfo {author} {\bibfnamefont {L.}~\bibnamefont {Xiao}}, \bibinfo
  {author} {\bibfnamefont {S.}~\bibnamefont {Schoenholz}}, \bibinfo {author}
  {\bibfnamefont {Y.}~\bibnamefont {Bahri}}, \bibinfo {author} {\bibfnamefont
  {R.}~\bibnamefont {Novak}}, \bibinfo {author} {\bibfnamefont
  {J.}~\bibnamefont {Sohl-Dickstein}},\ and\ \bibinfo {author} {\bibfnamefont
  {J.}~\bibnamefont {Pennington}},\ }\bibfield  {title} {\bibinfo {title} {Wide
  neural networks of any depth evolve as linear models under gradient
  descent},\ }in\ \href
  {https://proceedings.neurips.cc/paper/2019/file/0d1a9651497a38d8b1c3871c84528bd4-Paper.pdf}
  {\emph {\bibinfo {booktitle} {Advances in Neural Information Processing
  Systems}}},\ Vol.~\bibinfo {volume} {32},\ \bibinfo {editor} {edited by\
  \bibinfo {editor} {\bibfnamefont {H.}~\bibnamefont {Wallach}}, \bibinfo
  {editor} {\bibfnamefont {H.}~\bibnamefont {Larochelle}}, \bibinfo {editor}
  {\bibfnamefont {A.}~\bibnamefont {Beygelzimer}}, \bibinfo {editor}
  {\bibfnamefont {F.}~\bibnamefont {d\textquotesingle Alch\'{e}-Buc}}, \bibinfo
  {editor} {\bibfnamefont {E.}~\bibnamefont {Fox}},\ and\ \bibinfo {editor}
  {\bibfnamefont {R.}~\bibnamefont {Garnett}}}\ (\bibinfo  {publisher} {Curran
  Associates, Inc.},\ \bibinfo {year} {2019})\BibitemShut {NoStop}%
\bibitem [{\citenamefont {Cortes}\ and\ \citenamefont
  {Vapnik}(1995)}]{cortes1995support}%
  \BibitemOpen
  \bibfield  {author} {\bibinfo {author} {\bibfnamefont {C.}~\bibnamefont
  {Cortes}}\ and\ \bibinfo {author} {\bibfnamefont {V.}~\bibnamefont
  {Vapnik}},\ }\bibfield  {title} {\bibinfo {title} {Support-vector networks},\
  }\href {https://doi.org/10.1007/BF00994018} {\bibfield  {journal} {\bibinfo
  {journal} {Machine Learning}\ }\textbf {\bibinfo {volume} {20}},\ \bibinfo
  {pages} {273} (\bibinfo {year} {1995})}\BibitemShut {NoStop}%
\bibitem [{\citenamefont {Dietrich}\ \emph {et~al.}(1999)\citenamefont
  {Dietrich}, \citenamefont {Opper},\ and\ \citenamefont
  {Sompolinsky}}]{PhysRevLett.82.2975}%
  \BibitemOpen
  \bibfield  {author} {\bibinfo {author} {\bibfnamefont {R.}~\bibnamefont
  {Dietrich}}, \bibinfo {author} {\bibfnamefont {M.}~\bibnamefont {Opper}},\
  and\ \bibinfo {author} {\bibfnamefont {H.}~\bibnamefont {Sompolinsky}},\
  }\bibfield  {title} {\bibinfo {title} {Statistical mechanics of support
  vector networks},\ }\href {https://doi.org/10.1103/PhysRevLett.82.2975}
  {\bibfield  {journal} {\bibinfo  {journal} {Phys. Rev. Lett.}\ }\textbf
  {\bibinfo {volume} {82}},\ \bibinfo {pages} {2975} (\bibinfo {year}
  {1999})}\BibitemShut {NoStop}%
\bibitem [{\citenamefont {Bordelon}\ \emph {et~al.}(2020)\citenamefont
  {Bordelon}, \citenamefont {Canatar},\ and\ \citenamefont
  {Pehlevan}}]{pmlr-v119-bordelon20a}%
  \BibitemOpen
  \bibfield  {author} {\bibinfo {author} {\bibfnamefont {B.}~\bibnamefont
  {Bordelon}}, \bibinfo {author} {\bibfnamefont {A.}~\bibnamefont {Canatar}},\
  and\ \bibinfo {author} {\bibfnamefont {C.}~\bibnamefont {Pehlevan}},\
  }\bibfield  {title} {\bibinfo {title} {Spectrum dependent learning curves in
  kernel regression and wide neural networks},\ }in\ \href
  {https://proceedings.mlr.press/v119/bordelon20a.html} {\emph {\bibinfo
  {booktitle} {Proceedings of the 37th International Conference on Machine
  Learning}}},\ \bibinfo {series} {Proceedings of Machine Learning Research},
  Vol.\ \bibinfo {volume} {119},\ \bibinfo {editor} {edited by\ \bibinfo
  {editor} {\bibfnamefont {H.~D.}\ \bibnamefont {III}}\ and\ \bibinfo {editor}
  {\bibfnamefont {A.}~\bibnamefont {Singh}}}\ (\bibinfo  {publisher} {PMLR},\
  \bibinfo {year} {2020})\ pp.\ \bibinfo {pages} {1024--1034}\BibitemShut
  {NoStop}%
\bibitem [{\citenamefont {Canatar}\ \emph {et~al.}(2021)\citenamefont
  {Canatar}, \citenamefont {Bordelon},\ and\ \citenamefont
  {Pehlevan}}]{canatar2021}%
  \BibitemOpen
  \bibfield  {author} {\bibinfo {author} {\bibfnamefont {A.}~\bibnamefont
  {Canatar}}, \bibinfo {author} {\bibfnamefont {B.}~\bibnamefont {Bordelon}},\
  and\ \bibinfo {author} {\bibfnamefont {C.}~\bibnamefont {Pehlevan}},\
  }\bibfield  {title} {\bibinfo {title} {Spectral bias and task-model alignment
  explain generalization in kernel regression and infinitely wide neural
  networks},\ }\href {https://doi.org/10.1038/s41467-021-23103-1} {\bibfield
  {journal} {\bibinfo  {journal} {Nature communications}\ }\textbf {\bibinfo
  {volume} {12}},\ \bibinfo {pages} {1} (\bibinfo {year} {2021})}\BibitemShut
  {NoStop}%
\bibitem [{\citenamefont {Vyas}\ \emph {et~al.}(2022)\citenamefont {Vyas},
  \citenamefont {Bansal},\ and\ \citenamefont {Preetum}}]{Vyas2022}%
  \BibitemOpen
  \bibfield  {author} {\bibinfo {author} {\bibfnamefont {N.}~\bibnamefont
  {Vyas}}, \bibinfo {author} {\bibfnamefont {Y.}~\bibnamefont {Bansal}},\ and\
  \bibinfo {author} {\bibfnamefont {N.}~\bibnamefont {Preetum}},\ }\bibfield
  {title} {\bibinfo {title} {Limitations of the ntk for understanding
  generalization in deep learning},\ }\href@noop {} {\bibfield  {journal}
  {\bibinfo  {journal} {arXiv preprint arXiv:2206.10012}\ } (\bibinfo {year}
  {2022})}\BibitemShut {NoStop}%
\bibitem [{\citenamefont {Lee}\ \emph {et~al.}(2020)\citenamefont {Lee},
  \citenamefont {Schoenholz}, \citenamefont {Pennington}, \citenamefont
  {Adlam}, \citenamefont {Xiao}, \citenamefont {Novak},\ and\ \citenamefont
  {Sohl-Dickstein}}]{NEURIPS2020_ad086f59}%
  \BibitemOpen
  \bibfield  {author} {\bibinfo {author} {\bibfnamefont {J.}~\bibnamefont
  {Lee}}, \bibinfo {author} {\bibfnamefont {S.}~\bibnamefont {Schoenholz}},
  \bibinfo {author} {\bibfnamefont {J.}~\bibnamefont {Pennington}}, \bibinfo
  {author} {\bibfnamefont {B.}~\bibnamefont {Adlam}}, \bibinfo {author}
  {\bibfnamefont {L.}~\bibnamefont {Xiao}}, \bibinfo {author} {\bibfnamefont
  {R.}~\bibnamefont {Novak}},\ and\ \bibinfo {author} {\bibfnamefont
  {J.}~\bibnamefont {Sohl-Dickstein}},\ }\bibfield  {title} {\bibinfo {title}
  {Finite versus infinite neural networks: an empirical study},\ }in\ \href
  {https://proceedings.neurips.cc/paper/2020/file/ad086f59924fffe0773f8d0ca22ea712-Paper.pdf}
  {\emph {\bibinfo {booktitle} {Advances in Neural Information Processing
  Systems}}},\ Vol.~\bibinfo {volume} {33},\ \bibinfo {editor} {edited by\
  \bibinfo {editor} {\bibfnamefont {H.}~\bibnamefont {Larochelle}}, \bibinfo
  {editor} {\bibfnamefont {M.}~\bibnamefont {Ranzato}}, \bibinfo {editor}
  {\bibfnamefont {R.}~\bibnamefont {Hadsell}}, \bibinfo {editor} {\bibfnamefont
  {M.}~\bibnamefont {Balcan}},\ and\ \bibinfo {editor} {\bibfnamefont
  {H.}~\bibnamefont {Lin}}}\ (\bibinfo  {publisher} {Curran Associates, Inc.},\
  \bibinfo {year} {2020})\ pp.\ \bibinfo {pages} {15156--15172}\BibitemShut
  {NoStop}%
\bibitem [{\citenamefont {Atanasov}\ \emph {et~al.}(2022)\citenamefont
  {Atanasov}, \citenamefont {Bordelon}, \citenamefont {Sainathan},\ and\
  \citenamefont {Pehlevan}}]{atanasov2022onset}%
  \BibitemOpen
  \bibfield  {author} {\bibinfo {author} {\bibfnamefont {A.}~\bibnamefont
  {Atanasov}}, \bibinfo {author} {\bibfnamefont {B.}~\bibnamefont {Bordelon}},
  \bibinfo {author} {\bibfnamefont {S.}~\bibnamefont {Sainathan}},\ and\
  \bibinfo {author} {\bibfnamefont {C.}~\bibnamefont {Pehlevan}},\ }\bibfield
  {title} {\bibinfo {title} {The onset of variance-limited behavior for
  networks in the lazy and rich regimes},\ }\href@noop {} {\bibfield  {journal}
  {\bibinfo  {journal} {arXiv preprint arXiv:2212.12147}\ } (\bibinfo {year}
  {2022})}\BibitemShut {NoStop}%
\bibitem [{\citenamefont {Mei}\ \emph {et~al.}(2018)\citenamefont {Mei},
  \citenamefont {Montanari},\ and\ \citenamefont
  {Nguyen}}]{doi:10.1073/pnas.1806579115}%
  \BibitemOpen
  \bibfield  {author} {\bibinfo {author} {\bibfnamefont {S.}~\bibnamefont
  {Mei}}, \bibinfo {author} {\bibfnamefont {A.}~\bibnamefont {Montanari}},\
  and\ \bibinfo {author} {\bibfnamefont {P.-M.}\ \bibnamefont {Nguyen}},\
  }\bibfield  {title} {\bibinfo {title} {A mean field view of the landscape of
  two-layer neural networks},\ }\href {https://doi.org/10.1073/pnas.1806579115}
  {\bibfield  {journal} {\bibinfo  {journal} {Proceedings of the National
  Academy of Sciences}\ }\textbf {\bibinfo {volume} {115}},\ \bibinfo {pages}
  {E7665} (\bibinfo {year} {2018})},\ \Eprint
  {https://arxiv.org/abs/https://www.pnas.org/doi/pdf/10.1073/pnas.1806579115}
  {https://www.pnas.org/doi/pdf/10.1073/pnas.1806579115} \BibitemShut {NoStop}%
\bibitem [{\citenamefont {Geiger}\ \emph {et~al.}(2021)\citenamefont {Geiger},
  \citenamefont {Petrini},\ and\ \citenamefont {Wyart}}]{GEIGER20211}%
  \BibitemOpen
  \bibfield  {author} {\bibinfo {author} {\bibfnamefont {M.}~\bibnamefont
  {Geiger}}, \bibinfo {author} {\bibfnamefont {L.}~\bibnamefont {Petrini}},\
  and\ \bibinfo {author} {\bibfnamefont {M.}~\bibnamefont {Wyart}},\ }\bibfield
   {title} {\bibinfo {title} {Landscape and training regimes in deep
  learning},\ }\href
  {https://doi.org/https://doi.org/10.1016/j.physrep.2021.04.001} {\bibfield
  {journal} {\bibinfo  {journal} {Physics Reports}\ }\textbf {\bibinfo {volume}
  {924}},\ \bibinfo {pages} {1} (\bibinfo {year} {2021})},\ \bibinfo {note}
  {landscape and training regimes in deep learning}\BibitemShut {NoStop}%
\bibitem [{\citenamefont {Geiger}\ \emph {et~al.}(2020)\citenamefont {Geiger},
  \citenamefont {Spigler}, \citenamefont {Jacot},\ and\ \citenamefont
  {Wyart}}]{Geiger_2020}%
  \BibitemOpen
  \bibfield  {author} {\bibinfo {author} {\bibfnamefont {M.}~\bibnamefont
  {Geiger}}, \bibinfo {author} {\bibfnamefont {S.}~\bibnamefont {Spigler}},
  \bibinfo {author} {\bibfnamefont {A.}~\bibnamefont {Jacot}},\ and\ \bibinfo
  {author} {\bibfnamefont {M.}~\bibnamefont {Wyart}},\ }\bibfield  {title}
  {\bibinfo {title} {Disentangling feature and lazy training in deep neural
  networks},\ }\href {https://doi.org/10.1088/1742-5468/abc4de} {\bibfield
  {journal} {\bibinfo  {journal} {Journal of Statistical Mechanics: Theory and
  Experiment}\ }\textbf {\bibinfo {volume} {2020}},\ \bibinfo {pages} {113301}
  (\bibinfo {year} {2020})}\BibitemShut {NoStop}%
\bibitem [{\citenamefont {Seroussi}\ \emph {et~al.}(2023)\citenamefont
  {Seroussi}, \citenamefont {Naveh},\ and\ \citenamefont
  {Ringel}}]{seroussi2023natcomm}%
  \BibitemOpen
  \bibfield  {author} {\bibinfo {author} {\bibfnamefont {I.}~\bibnamefont
  {Seroussi}}, \bibinfo {author} {\bibfnamefont {G.}~\bibnamefont {Naveh}},\
  and\ \bibinfo {author} {\bibfnamefont {Z.}~\bibnamefont {Ringel}},\
  }\bibfield  {title} {\bibinfo {title} {Separation of scales and a
  thermodynamic description of feature learning in some cnns},\ }\href
  {https://doi.org/10.1038/s41467-023-36361-y} {\bibfield  {journal} {\bibinfo
  {journal} {Nature Communications}\ }\textbf {\bibinfo {volume} {14}},\
  \bibinfo {pages} {908} (\bibinfo {year} {2023})}\BibitemShut {NoStop}%
\bibitem [{\citenamefont {Naveh}\ and\ \citenamefont
  {Ringel}(2021)}]{NEURIPS2021_b24d2101}%
  \BibitemOpen
  \bibfield  {author} {\bibinfo {author} {\bibfnamefont {G.}~\bibnamefont
  {Naveh}}\ and\ \bibinfo {author} {\bibfnamefont {Z.}~\bibnamefont {Ringel}},\
  }\bibfield  {title} {\bibinfo {title} {A self consistent theory of gaussian
  processes captures feature learning effects in finite cnns},\ }in\ \href
  {https://proceedings.neurips.cc/paper/2021/file/b24d21019de5e59da180f1661904f49a-Paper.pdf}
  {\emph {\bibinfo {booktitle} {Advances in Neural Information Processing
  Systems}}},\ Vol.~\bibinfo {volume} {34},\ \bibinfo {editor} {edited by\
  \bibinfo {editor} {\bibfnamefont {M.}~\bibnamefont {Ranzato}}, \bibinfo
  {editor} {\bibfnamefont {A.}~\bibnamefont {Beygelzimer}}, \bibinfo {editor}
  {\bibfnamefont {Y.}~\bibnamefont {Dauphin}}, \bibinfo {editor} {\bibfnamefont
  {P.}~\bibnamefont {Liang}},\ and\ \bibinfo {editor} {\bibfnamefont {J.~W.}\
  \bibnamefont {Vaughan}}}\ (\bibinfo  {publisher} {Curran Associates, Inc.},\
  \bibinfo {year} {2021})\ pp.\ \bibinfo {pages} {21352--21364}\BibitemShut
  {NoStop}%
\bibitem [{\citenamefont {Zavatone-Veth}\ \emph {et~al.}(2021)\citenamefont
  {Zavatone-Veth}, \citenamefont {Canatar}, \citenamefont {Ruben},\ and\
  \citenamefont {Pehlevan}}]{NEURIPS2021_cf9dc5e4}%
  \BibitemOpen
  \bibfield  {author} {\bibinfo {author} {\bibfnamefont {J.}~\bibnamefont
  {Zavatone-Veth}}, \bibinfo {author} {\bibfnamefont {A.}~\bibnamefont
  {Canatar}}, \bibinfo {author} {\bibfnamefont {B.}~\bibnamefont {Ruben}},\
  and\ \bibinfo {author} {\bibfnamefont {C.}~\bibnamefont {Pehlevan}},\
  }\bibfield  {title} {\bibinfo {title} {Asymptotics of representation learning
  in finite bayesian neural networks},\ }in\ \href
  {https://proceedings.neurips.cc/paper/2021/file/cf9dc5e4e194fc21f397b4cac9cc3ae9-Paper.pdf}
  {\emph {\bibinfo {booktitle} {Advances in Neural Information Processing
  Systems}}},\ Vol.~\bibinfo {volume} {34},\ \bibinfo {editor} {edited by\
  \bibinfo {editor} {\bibfnamefont {M.}~\bibnamefont {Ranzato}}, \bibinfo
  {editor} {\bibfnamefont {A.}~\bibnamefont {Beygelzimer}}, \bibinfo {editor}
  {\bibfnamefont {Y.}~\bibnamefont {Dauphin}}, \bibinfo {editor} {\bibfnamefont
  {P.}~\bibnamefont {Liang}},\ and\ \bibinfo {editor} {\bibfnamefont {J.~W.}\
  \bibnamefont {Vaughan}}}\ (\bibinfo  {publisher} {Curran Associates, Inc.},\
  \bibinfo {year} {2021})\ pp.\ \bibinfo {pages} {24765--24777}\BibitemShut
  {NoStop}%
\bibitem [{\citenamefont {Zavatone-Veth}\ \emph {et~al.}(2022)\citenamefont
  {Zavatone-Veth}, \citenamefont {Tong},\ and\ \citenamefont
  {Pehlevan}}]{PhysRevE.105.064118}%
  \BibitemOpen
  \bibfield  {author} {\bibinfo {author} {\bibfnamefont {J.~A.}\ \bibnamefont
  {Zavatone-Veth}}, \bibinfo {author} {\bibfnamefont {W.~L.}\ \bibnamefont
  {Tong}},\ and\ \bibinfo {author} {\bibfnamefont {C.}~\bibnamefont
  {Pehlevan}},\ }\bibfield  {title} {\bibinfo {title} {Contrasting random and
  learned features in deep bayesian linear regression},\ }\href
  {https://doi.org/10.1103/PhysRevE.105.064118} {\bibfield  {journal} {\bibinfo
   {journal} {Phys. Rev. E}\ }\textbf {\bibinfo {volume} {105}},\ \bibinfo
  {pages} {064118} (\bibinfo {year} {2022})}\BibitemShut {NoStop}%
\bibitem [{\citenamefont {Zavatone-Veth}\ and\ \citenamefont
  {Pehlevan}(2021)}]{zavatone-veth2021exact}%
  \BibitemOpen
  \bibfield  {author} {\bibinfo {author} {\bibfnamefont {J.~A.}\ \bibnamefont
  {Zavatone-Veth}}\ and\ \bibinfo {author} {\bibfnamefont {C.}~\bibnamefont
  {Pehlevan}},\ }\bibfield  {title} {\bibinfo {title} {Exact marginal prior
  distributions of finite bayesian neural networks},\ }in\ \href
  {https://openreview.net/forum?id=MxE7xFzv0N8} {\emph {\bibinfo {booktitle}
  {Advances in Neural Information Processing Systems}}},\ \bibinfo {editor}
  {edited by\ \bibinfo {editor} {\bibfnamefont {A.}~\bibnamefont
  {Beygelzimer}}, \bibinfo {editor} {\bibfnamefont {Y.}~\bibnamefont
  {Dauphin}}, \bibinfo {editor} {\bibfnamefont {P.}~\bibnamefont {Liang}},\
  and\ \bibinfo {editor} {\bibfnamefont {J.~W.}\ \bibnamefont {Vaughan}}}\
  (\bibinfo {year} {2021})\BibitemShut {NoStop}%
\bibitem [{\citenamefont {Roberts}\ \emph {et~al.}(2022)\citenamefont
  {Roberts}, \citenamefont {Yaida},\ and\ \citenamefont {Hanin}}]{PDLT-2022}%
  \BibitemOpen
  \bibfield  {author} {\bibinfo {author} {\bibfnamefont {D.~A.}\ \bibnamefont
  {Roberts}}, \bibinfo {author} {\bibfnamefont {S.}~\bibnamefont {Yaida}},\
  and\ \bibinfo {author} {\bibfnamefont {B.}~\bibnamefont {Hanin}},\
  }\href@noop {} {\emph {\bibinfo {title} {The Principles of Deep Learning
  Theory}}}\ (\bibinfo  {publisher} {Cambridge University Press},\ \bibinfo
  {year} {2022})\ \bibinfo {note} {\url{https://deeplearningtheory.com}},\
  \Eprint {https://arxiv.org/abs/2106.10165} {arXiv:2106.10165 [cs.LG]}
  \BibitemShut {NoStop}%
\bibitem [{\citenamefont {Hanin}(2023)}]{hanin2023random}%
  \BibitemOpen
  \bibfield  {author} {\bibinfo {author} {\bibfnamefont {B.}~\bibnamefont
  {Hanin}},\ }\href@noop {} {\bibinfo {title} {Random fully connected neural
  networks as perturbatively solvable hierarchies}} (\bibinfo {year} {2023}),\
  \Eprint {https://arxiv.org/abs/2204.01058} {arXiv:2204.01058 [math.PR]}
  \BibitemShut {NoStop}%
\bibitem [{\citenamefont {Antognini}(2019)}]{antognini2019finite}%
  \BibitemOpen
  \bibfield  {author} {\bibinfo {author} {\bibfnamefont {J.~M.}\ \bibnamefont
  {Antognini}},\ }\href@noop {} {\bibinfo {title} {Finite size corrections for
  neural network gaussian processes}} (\bibinfo {year} {2019}),\ \Eprint
  {https://arxiv.org/abs/1908.10030} {arXiv:1908.10030 [cs.LG]} \BibitemShut
  {NoStop}%
\bibitem [{\citenamefont {Yaida}(2020)}]{yaida2020nonGauss}%
  \BibitemOpen
  \bibfield  {author} {\bibinfo {author} {\bibfnamefont {S.}~\bibnamefont
  {Yaida}},\ }\bibfield  {title} {\bibinfo {title} {Non-{G}aussian processes
  and neural networks at finite widths},\ }in\ \href
  {https://proceedings.mlr.press/v107/yaida20a.html} {\emph {\bibinfo
  {booktitle} {Proceedings of The First Mathematical and Scientific Machine
  Learning Conference}}},\ \bibinfo {series} {Proceedings of Machine Learning
  Research}, Vol.\ \bibinfo {volume} {107},\ \bibinfo {editor} {edited by\
  \bibinfo {editor} {\bibfnamefont {J.}~\bibnamefont {Lu}}\ and\ \bibinfo
  {editor} {\bibfnamefont {R.}~\bibnamefont {Ward}}}\ (\bibinfo  {publisher}
  {PMLR},\ \bibinfo {year} {2020})\ pp.\ \bibinfo {pages}
  {165--192}\BibitemShut {NoStop}%
\bibitem [{\citenamefont {Aitchison}(2020)}]{aitchison2020bigger}%
  \BibitemOpen
  \bibfield  {author} {\bibinfo {author} {\bibfnamefont {L.}~\bibnamefont
  {Aitchison}},\ }\bibfield  {title} {\bibinfo {title} {Why bigger is not
  always better: on finite and infinite neural networks},\ }in\ \href
  {https://proceedings.mlr.press/v119/aitchison20a.html} {\emph {\bibinfo
  {booktitle} {Proceedings of the 37th International Conference on Machine
  Learning}}},\ \bibinfo {series} {Proceedings of Machine Learning Research},
  Vol.\ \bibinfo {volume} {119},\ \bibinfo {editor} {edited by\ \bibinfo
  {editor} {\bibfnamefont {H.~D.}\ \bibnamefont {III}}\ and\ \bibinfo {editor}
  {\bibfnamefont {A.}~\bibnamefont {Singh}}}\ (\bibinfo  {publisher} {PMLR},\
  \bibinfo {year} {2020})\ pp.\ \bibinfo {pages} {156--164}\BibitemShut
  {NoStop}%
\bibitem [{\citenamefont {Yang}\ \emph {et~al.}(2023)\citenamefont {Yang},
  \citenamefont {Robeyns}, \citenamefont {Milsom}, \citenamefont {Schoots},\
  and\ \citenamefont {Aitchison}}]{yang2023theory}%
  \BibitemOpen
  \bibfield  {author} {\bibinfo {author} {\bibfnamefont {A.~X.}\ \bibnamefont
  {Yang}}, \bibinfo {author} {\bibfnamefont {M.}~\bibnamefont {Robeyns}},
  \bibinfo {author} {\bibfnamefont {E.}~\bibnamefont {Milsom}}, \bibinfo
  {author} {\bibfnamefont {N.}~\bibnamefont {Schoots}},\ and\ \bibinfo {author}
  {\bibfnamefont {L.}~\bibnamefont {Aitchison}},\ }\href@noop {} {\bibinfo
  {title} {A theory of representation learning in deep neural networks gives a
  deep generalisation of kernel methods}} (\bibinfo {year} {2023}),\ \Eprint
  {https://arxiv.org/abs/2108.13097} {arXiv:2108.13097 [stat.ML]} \BibitemShut
  {NoStop}%
\bibitem [{\citenamefont {Favaro}\ \emph {et~al.}(2023)\citenamefont {Favaro},
  \citenamefont {Hanin}, \citenamefont {Marinucci}, \citenamefont {Nourdin},\
  and\ \citenamefont {Peccati}}]{favaro2023quantitative}%
  \BibitemOpen
  \bibfield  {author} {\bibinfo {author} {\bibfnamefont {S.}~\bibnamefont
  {Favaro}}, \bibinfo {author} {\bibfnamefont {B.}~\bibnamefont {Hanin}},
  \bibinfo {author} {\bibfnamefont {D.}~\bibnamefont {Marinucci}}, \bibinfo
  {author} {\bibfnamefont {I.}~\bibnamefont {Nourdin}},\ and\ \bibinfo {author}
  {\bibfnamefont {G.}~\bibnamefont {Peccati}},\ }\bibfield  {title} {\bibinfo
  {title} {Quantitative clts in deep neural networks},\ }\href@noop {}
  {\bibfield  {journal} {\bibinfo  {journal} {arXiv preprint arXiv:2307.06092}\
  } (\bibinfo {year} {2023})}\BibitemShut {NoStop}%
\bibitem [{\citenamefont {Cagnetta}\ \emph {et~al.}(2023)\citenamefont
  {Cagnetta}, \citenamefont {Favero},\ and\ \citenamefont
  {Wyart}}]{cagnetta2023what}%
  \BibitemOpen
  \bibfield  {author} {\bibinfo {author} {\bibfnamefont {F.}~\bibnamefont
  {Cagnetta}}, \bibinfo {author} {\bibfnamefont {A.}~\bibnamefont {Favero}},\
  and\ \bibinfo {author} {\bibfnamefont {M.}~\bibnamefont {Wyart}},\ }\href
  {https://openreview.net/forum?id=m3QhpKNXU6-} {\bibinfo {title} {What can be
  learnt with wide convolutional neural networks?}} (\bibinfo {year}
  {2023})\BibitemShut {NoStop}%
\bibitem [{\citenamefont {Favero}\ \emph {et~al.}(2021)\citenamefont {Favero},
  \citenamefont {Cagnetta},\ and\ \citenamefont {Wyart}}]{favero2021locality}%
  \BibitemOpen
  \bibfield  {author} {\bibinfo {author} {\bibfnamefont {A.}~\bibnamefont
  {Favero}}, \bibinfo {author} {\bibfnamefont {F.}~\bibnamefont {Cagnetta}},\
  and\ \bibinfo {author} {\bibfnamefont {M.}~\bibnamefont {Wyart}},\ }\bibfield
   {title} {\bibinfo {title} {Locality defeats the curse of dimensionality in
  convolutional teacher-student scenarios},\ }in\ \href
  {https://openreview.net/forum?id=sBBnfOFtPc} {\emph {\bibinfo {booktitle}
  {Advances in Neural Information Processing Systems}}},\ \bibinfo {editor}
  {edited by\ \bibinfo {editor} {\bibfnamefont {A.}~\bibnamefont
  {Beygelzimer}}, \bibinfo {editor} {\bibfnamefont {Y.}~\bibnamefont
  {Dauphin}}, \bibinfo {editor} {\bibfnamefont {P.}~\bibnamefont {Liang}},\
  and\ \bibinfo {editor} {\bibfnamefont {J.~W.}\ \bibnamefont {Vaughan}}}\
  (\bibinfo {year} {2021})\BibitemShut {NoStop}%
\bibitem [{\citenamefont {Petrini}\ \emph {et~al.}(2022)\citenamefont
  {Petrini}, \citenamefont {Cagnetta}, \citenamefont {Vanden-Eijnden},\ and\
  \citenamefont {Wyart}}]{petrini2022learning}%
  \BibitemOpen
  \bibfield  {author} {\bibinfo {author} {\bibfnamefont {L.}~\bibnamefont
  {Petrini}}, \bibinfo {author} {\bibfnamefont {F.}~\bibnamefont {Cagnetta}},
  \bibinfo {author} {\bibfnamefont {E.}~\bibnamefont {Vanden-Eijnden}},\ and\
  \bibinfo {author} {\bibfnamefont {M.}~\bibnamefont {Wyart}},\ }\bibfield
  {title} {\bibinfo {title} {Learning sparse features can lead to overfitting
  in neural networks},\ }in\ \href
  {https://openreview.net/forum?id=dZEZu7zxJBF} {\emph {\bibinfo {booktitle}
  {Advances in Neural Information Processing Systems}}},\ \bibinfo {editor}
  {edited by\ \bibinfo {editor} {\bibfnamefont {A.~H.}\ \bibnamefont {Oh}},
  \bibinfo {editor} {\bibfnamefont {A.}~\bibnamefont {Agarwal}}, \bibinfo
  {editor} {\bibfnamefont {D.}~\bibnamefont {Belgrave}},\ and\ \bibinfo
  {editor} {\bibfnamefont {K.}~\bibnamefont {Cho}}}\ (\bibinfo {year}
  {2022})\BibitemShut {NoStop}%
\bibitem [{\citenamefont {Ariosto}\ \emph {et~al.}(2022)\citenamefont
  {Ariosto}, \citenamefont {Pacelli}, \citenamefont {Pastore}, \citenamefont
  {Ginelli}, \citenamefont {Gherardi},\ and\ \citenamefont
  {Rotondo}}]{ariosto2022statistical}%
  \BibitemOpen
  \bibfield  {author} {\bibinfo {author} {\bibfnamefont {S.}~\bibnamefont
  {Ariosto}}, \bibinfo {author} {\bibfnamefont {R.}~\bibnamefont {Pacelli}},
  \bibinfo {author} {\bibfnamefont {M.}~\bibnamefont {Pastore}}, \bibinfo
  {author} {\bibfnamefont {F.}~\bibnamefont {Ginelli}}, \bibinfo {author}
  {\bibfnamefont {M.}~\bibnamefont {Gherardi}},\ and\ \bibinfo {author}
  {\bibfnamefont {P.}~\bibnamefont {Rotondo}},\ }\bibfield  {title} {\bibinfo
  {title} {Statistical mechanics of deep learning beyond the infinite-width
  limit},\ }\href@noop {} {\bibfield  {journal} {\bibinfo  {journal} {arXiv
  preprint arXiv:2209.04882}\ } (\bibinfo {year} {2022})}\BibitemShut {NoStop}%
\bibitem [{\citenamefont {Li}\ and\ \citenamefont
  {Sompolinsky}(2021)}]{SompolinskyLinear}%
  \BibitemOpen
  \bibfield  {author} {\bibinfo {author} {\bibfnamefont {Q.}~\bibnamefont
  {Li}}\ and\ \bibinfo {author} {\bibfnamefont {H.}~\bibnamefont
  {Sompolinsky}},\ }\bibfield  {title} {\bibinfo {title} {Statistical mechanics
  of deep linear neural networks: The backpropagating kernel renormalization},\
  }\href {https://doi.org/10.1103/PhysRevX.11.031059} {\bibfield  {journal}
  {\bibinfo  {journal} {Phys. Rev. X}\ }\textbf {\bibinfo {volume} {11}},\
  \bibinfo {pages} {031059} (\bibinfo {year} {2021})}\BibitemShut {NoStop}%
\bibitem [{\citenamefont {Hanin}\ and\ \citenamefont
  {Zlokapa}(2023)}]{doi:10.1073/pnas.2301345120}%
  \BibitemOpen
  \bibfield  {author} {\bibinfo {author} {\bibfnamefont {B.}~\bibnamefont
  {Hanin}}\ and\ \bibinfo {author} {\bibfnamefont {A.}~\bibnamefont
  {Zlokapa}},\ }\bibfield  {title} {\bibinfo {title} {Bayesian interpolation
  with deep linear networks},\ }\href {https://doi.org/10.1073/pnas.2301345120}
  {\bibfield  {journal} {\bibinfo  {journal} {Proceedings of the National
  Academy of Sciences}\ }\textbf {\bibinfo {volume} {120}},\ \bibinfo {pages}
  {e2301345120} (\bibinfo {year} {2023})},\ \Eprint
  {https://arxiv.org/abs/https://www.pnas.org/doi/pdf/10.1073/pnas.2301345120}
  {https://www.pnas.org/doi/pdf/10.1073/pnas.2301345120} \BibitemShut {NoStop}%
\bibitem [{\citenamefont {Li}\ and\ \citenamefont
  {Sompolinsky}(2022)}]{li2022globally}%
  \BibitemOpen
  \bibfield  {author} {\bibinfo {author} {\bibfnamefont {Q.}~\bibnamefont
  {Li}}\ and\ \bibinfo {author} {\bibfnamefont {H.}~\bibnamefont
  {Sompolinsky}},\ }\bibfield  {title} {\bibinfo {title} {Globally gated deep
  linear networks},\ }\href@noop {} {\bibfield  {journal} {\bibinfo  {journal}
  {arXiv preprint arXiv:2210.17449}\ } (\bibinfo {year} {2022})}\BibitemShut
  {NoStop}%
\bibitem [{\citenamefont {Mei}\ and\ \citenamefont
  {Montanari}(2019)}]{mei2019}%
  \BibitemOpen
  \bibfield  {author} {\bibinfo {author} {\bibfnamefont {S.}~\bibnamefont
  {Mei}}\ and\ \bibinfo {author} {\bibfnamefont {A.}~\bibnamefont
  {Montanari}},\ }\bibfield  {title} {\bibinfo {title} {The generalization
  error of random features regression: Precise asymptotics and the double
  descent curve},\ }\href {https://doi.org/10.1002/cpa.22008} {\bibfield
  {journal} {\bibinfo  {journal} {Communications on Pure and Applied
  Mathematics}\ } (\bibinfo {year} {2019})}\BibitemShut {NoStop}%
\bibitem [{\citenamefont {Goldt}\ \emph {et~al.}(2020)\citenamefont {Goldt},
  \citenamefont {Loureiro}, \citenamefont {Reeves}, \citenamefont {Krzakala},
  \citenamefont {M{\'e}zard},\ and\ \citenamefont
  {Zdeborov{\'a}}}]{goldt2020gaussian}%
  \BibitemOpen
  \bibfield  {author} {\bibinfo {author} {\bibfnamefont {S.}~\bibnamefont
  {Goldt}}, \bibinfo {author} {\bibfnamefont {B.}~\bibnamefont {Loureiro}},
  \bibinfo {author} {\bibfnamefont {G.}~\bibnamefont {Reeves}}, \bibinfo
  {author} {\bibfnamefont {F.}~\bibnamefont {Krzakala}}, \bibinfo {author}
  {\bibfnamefont {M.}~\bibnamefont {M{\'e}zard}},\ and\ \bibinfo {author}
  {\bibfnamefont {L.}~\bibnamefont {Zdeborov{\'a}}},\ }\bibfield  {title}
  {\bibinfo {title} {The gaussian equivalence of generative models for learning
  with shallow neural networks},\ }\href@noop {} {\bibfield  {journal}
  {\bibinfo  {journal} {arXiv preprint arXiv:2006.14709}\ } (\bibinfo {year}
  {2020})}\BibitemShut {NoStop}%
\bibitem [{\citenamefont {Gerace}\ \emph {et~al.}(2021)\citenamefont {Gerace},
  \citenamefont {Loureiro}, \citenamefont {Krzakala}, \citenamefont {Mézard},\
  and\ \citenamefont {Zdeborová}}]{Gerace_2021}%
  \BibitemOpen
  \bibfield  {author} {\bibinfo {author} {\bibfnamefont {F.}~\bibnamefont
  {Gerace}}, \bibinfo {author} {\bibfnamefont {B.}~\bibnamefont {Loureiro}},
  \bibinfo {author} {\bibfnamefont {F.}~\bibnamefont {Krzakala}}, \bibinfo
  {author} {\bibfnamefont {M.}~\bibnamefont {Mézard}},\ and\ \bibinfo {author}
  {\bibfnamefont {L.}~\bibnamefont {Zdeborová}},\ }\bibfield  {title}
  {\bibinfo {title} {Generalisation error in learning with random features and
  the hidden manifold model},\ }\href
  {https://doi.org/10.1088/1742-5468/ac3ae6} {\bibfield  {journal} {\bibinfo
  {journal} {Journal of Statistical Mechanics: Theory and Experiment}\ }\textbf
  {\bibinfo {volume} {2021}},\ \bibinfo {pages} {124013} (\bibinfo {year}
  {2021})}\BibitemShut {NoStop}%
\bibitem [{\citenamefont {Loureiro}\ \emph {et~al.}(2021)\citenamefont
  {Loureiro}, \citenamefont {Gerbelot}, \citenamefont {Cui}, \citenamefont
  {Goldt}, \citenamefont {Krzakala}, \citenamefont {Mezard},\ and\
  \citenamefont {Zdeborov{\'a}}}]{loureiro2021learning}%
  \BibitemOpen
  \bibfield  {author} {\bibinfo {author} {\bibfnamefont {B.}~\bibnamefont
  {Loureiro}}, \bibinfo {author} {\bibfnamefont {C.}~\bibnamefont {Gerbelot}},
  \bibinfo {author} {\bibfnamefont {H.}~\bibnamefont {Cui}}, \bibinfo {author}
  {\bibfnamefont {S.}~\bibnamefont {Goldt}}, \bibinfo {author} {\bibfnamefont
  {F.}~\bibnamefont {Krzakala}}, \bibinfo {author} {\bibfnamefont
  {M.}~\bibnamefont {Mezard}},\ and\ \bibinfo {author} {\bibfnamefont
  {L.}~\bibnamefont {Zdeborov{\'a}}},\ }\bibfield  {title} {\bibinfo {title}
  {Learning curves of generic features maps for realistic datasets with a
  teacher-student model},\ }\href@noop {} {\bibfield  {journal} {\bibinfo
  {journal} {Advances in Neural Information Processing Systems}\ }\textbf
  {\bibinfo {volume} {34}} (\bibinfo {year} {2021})}\BibitemShut {NoStop}%
\bibitem [{\citenamefont {Breuer}\ and\ \citenamefont {Major}(1983)}]{BM}%
  \BibitemOpen
  \bibfield  {author} {\bibinfo {author} {\bibfnamefont {P.}~\bibnamefont
  {Breuer}}\ and\ \bibinfo {author} {\bibfnamefont {P.}~\bibnamefont {Major}},\
  }\bibfield  {title} {\bibinfo {title} {Central limit theorems for non-linear
  functionals of gaussian fields},\ }\href
  {https://doi.org/10.1016/0047-259X(83)90019-2} {\bibfield  {journal}
  {\bibinfo  {journal} {Journal of Multivariate Analysis}\ }\textbf {\bibinfo
  {volume} {13}},\ \bibinfo {pages} {425} (\bibinfo {year} {1983})}\BibitemShut
  {NoStop}%
\bibitem [{\citenamefont {Shah}\ \emph {et~al.}(2014)\citenamefont {Shah},
  \citenamefont {Wilson},\ and\ \citenamefont {Ghahramani}}]{Shah2014}%
  \BibitemOpen
  \bibfield  {author} {\bibinfo {author} {\bibfnamefont {A.}~\bibnamefont
  {Shah}}, \bibinfo {author} {\bibfnamefont {A.}~\bibnamefont {Wilson}},\ and\
  \bibinfo {author} {\bibfnamefont {Z.}~\bibnamefont {Ghahramani}},\ }\bibfield
   {title} {\bibinfo {title} {{Student-t Processes as Alternatives to Gaussian
  Processes}},\ }in\ \href {https://proceedings.mlr.press/v33/shah14.html}
  {\emph {\bibinfo {booktitle} {Proceedings of the Seventeenth International
  Conference on Artificial Intelligence and Statistics}}},\ \bibinfo {series}
  {Proceedings of Machine Learning Research}, Vol.~\bibinfo {volume} {33},\
  \bibinfo {editor} {edited by\ \bibinfo {editor} {\bibfnamefont
  {S.}~\bibnamefont {Kaski}}\ and\ \bibinfo {editor} {\bibfnamefont
  {J.}~\bibnamefont {Corander}}}\ (\bibinfo  {publisher} {PMLR},\ \bibinfo
  {address} {Reykjavik, Iceland},\ \bibinfo {year} {2014})\ pp.\ \bibinfo
  {pages} {877--885}\BibitemShut {NoStop}%
\bibitem [{\citenamefont {Cui}\ \emph {et~al.}(2023)\citenamefont {Cui},
  \citenamefont {Krzakala},\ and\ \citenamefont
  {Zdeborov{\'a}}}]{cui2023optimal}%
  \BibitemOpen
  \bibfield  {author} {\bibinfo {author} {\bibfnamefont {H.}~\bibnamefont
  {Cui}}, \bibinfo {author} {\bibfnamefont {F.}~\bibnamefont {Krzakala}},\ and\
  \bibinfo {author} {\bibfnamefont {L.}~\bibnamefont {Zdeborov{\'a}}},\
  }\bibfield  {title} {\bibinfo {title} {Optimal learning of deep random
  networks of extensive-width},\ }\href@noop {} {\bibfield  {journal} {\bibinfo
   {journal} {arXiv preprint arXiv:2302.00375}\ } (\bibinfo {year}
  {2023})}\BibitemShut {NoStop}%
\bibitem [{\citenamefont {Schr{\"o}der}\ \emph {et~al.}(2023)\citenamefont
  {Schr{\"o}der}, \citenamefont {Cui}, \citenamefont {Dmitriev},\ and\
  \citenamefont {Loureiro}}]{schroder2023deterministic}%
  \BibitemOpen
  \bibfield  {author} {\bibinfo {author} {\bibfnamefont {D.}~\bibnamefont
  {Schr{\"o}der}}, \bibinfo {author} {\bibfnamefont {H.}~\bibnamefont {Cui}},
  \bibinfo {author} {\bibfnamefont {D.}~\bibnamefont {Dmitriev}},\ and\
  \bibinfo {author} {\bibfnamefont {B.}~\bibnamefont {Loureiro}},\ }\bibfield
  {title} {\bibinfo {title} {Deterministic equivalent and error universality of
  deep random features learning},\ }\href@noop {} {\bibfield  {journal}
  {\bibinfo  {journal} {arXiv preprint arXiv:2302.00401}\ } (\bibinfo {year}
  {2023})}\BibitemShut {NoStop}%
\bibitem [{\citenamefont {Camilli}\ \emph {et~al.}(2023)\citenamefont
  {Camilli}, \citenamefont {Tieplova},\ and\ \citenamefont
  {Barbier}}]{camilli2023fundamental}%
  \BibitemOpen
  \bibfield  {author} {\bibinfo {author} {\bibfnamefont {F.}~\bibnamefont
  {Camilli}}, \bibinfo {author} {\bibfnamefont {D.}~\bibnamefont {Tieplova}},\
  and\ \bibinfo {author} {\bibfnamefont {J.}~\bibnamefont {Barbier}},\
  }\bibfield  {title} {\bibinfo {title} {Fundamental limits of overparametrized
  shallow neural networks for supervised learning},\ }\href@noop {} {\bibfield
  {journal} {\bibinfo  {journal} {arXiv preprint arXiv:2307.05635}\ } (\bibinfo
  {year} {2023})}\BibitemShut {NoStop}%
\bibitem [{\citenamefont {Guerra}\ and\ \citenamefont
  {Toninelli}(2002)}]{guerra2002thermodynamic}%
  \BibitemOpen
  \bibfield  {author} {\bibinfo {author} {\bibfnamefont {F.}~\bibnamefont
  {Guerra}}\ and\ \bibinfo {author} {\bibfnamefont {F.~L.}\ \bibnamefont
  {Toninelli}},\ }\bibfield  {title} {\bibinfo {title} {The thermodynamic limit
  in mean field spin glass models},\ }\href
  {https://doi.org/10.1007/s00220-002-0699-y} {\bibfield  {journal} {\bibinfo
  {journal} {Communications in Mathematical Physics}\ }\textbf {\bibinfo
  {volume} {230}},\ \bibinfo {pages} {71} (\bibinfo {year} {2002})}\BibitemShut
  {NoStop}%
\bibitem [{\citenamefont {Agliari}\ \emph {et~al.}(2020)\citenamefont
  {Agliari}, \citenamefont {Alemanno}, \citenamefont {Barra},\ and\
  \citenamefont {Fachechi}}]{AGLIARI2020254}%
  \BibitemOpen
  \bibfield  {author} {\bibinfo {author} {\bibfnamefont {E.}~\bibnamefont
  {Agliari}}, \bibinfo {author} {\bibfnamefont {F.}~\bibnamefont {Alemanno}},
  \bibinfo {author} {\bibfnamefont {A.}~\bibnamefont {Barra}},\ and\ \bibinfo
  {author} {\bibfnamefont {A.}~\bibnamefont {Fachechi}},\ }\bibfield  {title}
  {\bibinfo {title} {Generalized guerra’s interpolation schemes for dense
  associative neural networks},\ }\href
  {https://doi.org/https://doi.org/10.1016/j.neunet.2020.05.009} {\bibfield
  {journal} {\bibinfo  {journal} {Neural Networks}\ }\textbf {\bibinfo {volume}
  {128}},\ \bibinfo {pages} {254} (\bibinfo {year} {2020})}\BibitemShut
  {NoStop}%
\bibitem [{\citenamefont {Ingrosso}\ and\ \citenamefont
  {Goldt}(2022)}]{doi:10.1073/pnas.2201854119}%
  \BibitemOpen
  \bibfield  {author} {\bibinfo {author} {\bibfnamefont {A.}~\bibnamefont
  {Ingrosso}}\ and\ \bibinfo {author} {\bibfnamefont {S.}~\bibnamefont
  {Goldt}},\ }\bibfield  {title} {\bibinfo {title} {Data-driven emergence of
  convolutional structure in neural networks},\ }\href
  {https://doi.org/10.1073/pnas.2201854119} {\bibfield  {journal} {\bibinfo
  {journal} {Proceedings of the National Academy of Sciences}\ }\textbf
  {\bibinfo {volume} {119}},\ \bibinfo {pages} {e2201854119} (\bibinfo {year}
  {2022})},\ \Eprint
  {https://arxiv.org/abs/https://www.pnas.org/doi/pdf/10.1073/pnas.2201854119}
  {https://www.pnas.org/doi/pdf/10.1073/pnas.2201854119} \BibitemShut {NoStop}%
\bibitem [{\citenamefont {Kingma}\ and\ \citenamefont {Ba}(2014)}]{adam}%
  \BibitemOpen
  \bibfield  {author} {\bibinfo {author} {\bibfnamefont {D.~P.}\ \bibnamefont
  {Kingma}}\ and\ \bibinfo {author} {\bibfnamefont {J.}~\bibnamefont {Ba}},\
  }\bibfield  {title} {\bibinfo {title} {Adam: A method for stochastic
  optimization},\ }\href@noop {} {\bibfield  {journal} {\bibinfo  {journal}
  {arXiv preprint arXiv:1412.6980}\ } (\bibinfo {year} {2014})}\BibitemShut
  {NoStop}%
\bibitem [{\citenamefont {Abadi}\ \emph {et~al.}(2015)\citenamefont {Abadi},
  \citenamefont {Agarwal}, \citenamefont {Barham}, \citenamefont {Brevdo},
  \citenamefont {Chen}, \citenamefont {Citro}, \citenamefont {Corrado},
  \citenamefont {Davis}, \citenamefont {Dean}, \citenamefont {Devin},
  \citenamefont {Ghemawat}, \citenamefont {Goodfellow}, \citenamefont {Harp},
  \citenamefont {Irving}, \citenamefont {Isard}, \citenamefont {Jia},
  \citenamefont {Jozefowicz}, \citenamefont {Kaiser}, \citenamefont {Kudlur},
  \citenamefont {Levenberg}, \citenamefont {Man\'{e}}, \citenamefont {Monga},
  \citenamefont {Moore}, \citenamefont {Murray}, \citenamefont {Olah},
  \citenamefont {Schuster}, \citenamefont {Shlens}, \citenamefont {Steiner},
  \citenamefont {Sutskever}, \citenamefont {Talwar}, \citenamefont {Tucker},
  \citenamefont {Vanhoucke}, \citenamefont {Vasudevan}, \citenamefont
  {Vi\'{e}gas}, \citenamefont {Vinyals}, \citenamefont {Warden}, \citenamefont
  {Wattenberg}, \citenamefont {Wicke}, \citenamefont {Yu},\ and\ \citenamefont
  {Zheng}}]{tensorflow2015-whitepaper}%
  \BibitemOpen
  \bibfield  {author} {\bibinfo {author} {\bibfnamefont {M.}~\bibnamefont
  {Abadi}}, \bibinfo {author} {\bibfnamefont {A.}~\bibnamefont {Agarwal}},
  \bibinfo {author} {\bibfnamefont {P.}~\bibnamefont {Barham}}, \bibinfo
  {author} {\bibfnamefont {E.}~\bibnamefont {Brevdo}}, \bibinfo {author}
  {\bibfnamefont {Z.}~\bibnamefont {Chen}}, \bibinfo {author} {\bibfnamefont
  {C.}~\bibnamefont {Citro}}, \bibinfo {author} {\bibfnamefont {G.~S.}\
  \bibnamefont {Corrado}}, \bibinfo {author} {\bibfnamefont {A.}~\bibnamefont
  {Davis}}, \bibinfo {author} {\bibfnamefont {J.}~\bibnamefont {Dean}},
  \bibinfo {author} {\bibfnamefont {M.}~\bibnamefont {Devin}}, \bibinfo
  {author} {\bibfnamefont {S.}~\bibnamefont {Ghemawat}}, \bibinfo {author}
  {\bibfnamefont {I.}~\bibnamefont {Goodfellow}}, \bibinfo {author}
  {\bibfnamefont {A.}~\bibnamefont {Harp}}, \bibinfo {author} {\bibfnamefont
  {G.}~\bibnamefont {Irving}}, \bibinfo {author} {\bibfnamefont
  {M.}~\bibnamefont {Isard}}, \bibinfo {author} {\bibfnamefont
  {Y.}~\bibnamefont {Jia}}, \bibinfo {author} {\bibfnamefont {R.}~\bibnamefont
  {Jozefowicz}}, \bibinfo {author} {\bibfnamefont {L.}~\bibnamefont {Kaiser}},
  \bibinfo {author} {\bibfnamefont {M.}~\bibnamefont {Kudlur}}, \bibinfo
  {author} {\bibfnamefont {J.}~\bibnamefont {Levenberg}}, \bibinfo {author}
  {\bibfnamefont {D.}~\bibnamefont {Man\'{e}}}, \bibinfo {author}
  {\bibfnamefont {R.}~\bibnamefont {Monga}}, \bibinfo {author} {\bibfnamefont
  {S.}~\bibnamefont {Moore}}, \bibinfo {author} {\bibfnamefont
  {D.}~\bibnamefont {Murray}}, \bibinfo {author} {\bibfnamefont
  {C.}~\bibnamefont {Olah}}, \bibinfo {author} {\bibfnamefont {M.}~\bibnamefont
  {Schuster}}, \bibinfo {author} {\bibfnamefont {J.}~\bibnamefont {Shlens}},
  \bibinfo {author} {\bibfnamefont {B.}~\bibnamefont {Steiner}}, \bibinfo
  {author} {\bibfnamefont {I.}~\bibnamefont {Sutskever}}, \bibinfo {author}
  {\bibfnamefont {K.}~\bibnamefont {Talwar}}, \bibinfo {author} {\bibfnamefont
  {P.}~\bibnamefont {Tucker}}, \bibinfo {author} {\bibfnamefont
  {V.}~\bibnamefont {Vanhoucke}}, \bibinfo {author} {\bibfnamefont
  {V.}~\bibnamefont {Vasudevan}}, \bibinfo {author} {\bibfnamefont
  {F.}~\bibnamefont {Vi\'{e}gas}}, \bibinfo {author} {\bibfnamefont
  {O.}~\bibnamefont {Vinyals}}, \bibinfo {author} {\bibfnamefont
  {P.}~\bibnamefont {Warden}}, \bibinfo {author} {\bibfnamefont
  {M.}~\bibnamefont {Wattenberg}}, \bibinfo {author} {\bibfnamefont
  {M.}~\bibnamefont {Wicke}}, \bibinfo {author} {\bibfnamefont
  {Y.}~\bibnamefont {Yu}},\ and\ \bibinfo {author} {\bibfnamefont
  {X.}~\bibnamefont {Zheng}},\ }\href {https://www.tensorflow.org/} {\bibinfo
  {title} {{TensorFlow}: Large-scale machine learning on heterogeneous
  systems}} (\bibinfo {year} {2015}),\ \bibinfo {note} {software available from
  tensorflow.org}\BibitemShut {NoStop}%
\end{thebibliography}%

\onecolumngrid
\appendix
\section{Derivation of the effective action for a shallow Locally Connected Network}

Here we present the explicit computation of the effective action for one hidden layer LCNs \cite{novak2019bayesian}.
%As for the CNN case in the main text 
For simplicity, we consider one-dimensional local interaction, but the calculation can be generalized to a $d$-dimensional setting. For LCNs, the pre-activations of the hidden layer are given by:
\begin{equation}
   h_i^a (x)= \sum_{m=-\lfloor M/2\rfloor}^{\lfloor M/2 \rfloor}\frac{W_{i m}^a x_{S i + m}}{\sqrt{M}}\,,
\end{equation}
where $M$ indicates the size of the filter and $S$ denotes the value of the stride. The index $a= 1 \ldots N_c$ runs over the filters. From the pre-activations, we can define the output of the LCN as:
\begin{equation}
f_{\textrm{LCN}}(x) =  \frac{1}{\sqrt{N_c \lfloor N_0/S \rfloor}}\sum_{i = 1}^{\lfloor N_0/S \rfloor} \sum_{a = 1}^{N_c} v_i^a \sigma\left( h_i^a\right).
\label{LCNoutput}
\end{equation}
 We recall the notation for the total number of weights in the last layer $N_1 = N_c \lfloor N_0/S \rfloor$. Note that in the special case $S=N_0$ we recover the FC architecture.

Our objective is to construct the data-dependent partition function for this learning problem in the proportional limit: %mirroring the procedure detailed in \cite{ariosto2022statistical},
\begin{align}
Z = &\int \prod_{i,a} dv_i^a \prod_{i,m,a} dW_{im}^a \, \exp \Big\{-\frac{\lambda_1}{2}\|{v}\|^2 -\frac{\lambda_0}{2}\|{W}\|^2 -\frac{\beta}{2}\sum_{\mu=1}^P \left[y^\mu - f_{\textrm{LCN}}(x^\mu)\right]^2 \Big \} \,.
\label{LCN_1_partition}  
\end{align}
Here, we employ the same notation of \cite{ariosto2022statistical}, where $\lambda_0$ and $\lambda_1$ are respectively the Gaussian priors of the hidden and last layer.
We introduce two sets of Dirac deltas, corresponding to the pre-activations of the hidden layer and the output of the network. We denote these new degrees of freedom as $h_{i a}^\mu$ and $s^\mu$ respectively: 
\begin{equation}
    \prod_{ i, a, \mu}\delta\Big(h_{i a}^\mu-\frac{1}{\sqrt{M}}\sum_{m=-\lfloor \frac{M}{2}\rfloor}^{\lfloor \frac{M}{2}\rfloor}W_{i m}^a x_{S i+m}^\mu\Big)\,, \qquad \prod_{\mu}\delta\Big(s^{\mu}-\frac{1}{\sqrt{N_1}} \sum_{i, a = 1}^{\lfloor\frac{N_0}{S}\rfloor,N_c} v_i^a \sigma(h_{i a}^\mu)\Big) \,.
\end{equation}
%\begin{align}
%Z = &\int \prod_{i,a,\mu} dh_{i a}^\mu \prod_{\mu} ds^\mu \, e^{ -\frac{\beta}{2}\sum_{\mu=1}^P (y^\mu - s^\mu)^2}\nonumber\\
%& \times \Bigg\langle \prod_{\mu}\delta\Big(s^{\mu}-\frac{1}{\sqrt{N_1}} \sum_{i, a = 1}^{\lfloor\frac{N_0}{S}\rfloor,N_c} v_i^a \sigma(h_{i a}^\mu)\Big) \Bigg\rangle_{{v}} \nonumber\\
%& \times \Bigg\langle \prod_{ i, a, \mu}\delta\Big(h_{i a}^\mu-\frac{1}{\sqrt{M}}\sum_{m=-\lfloor \frac{M}{2}\rfloor}^{\lfloor \frac{M}{2}\rfloor}W_{i m}^a x_{S i+m}^\mu\Big) \Bigg\rangle_{{W}} \,,
%\label{LCN_2_partition}  
%\end{align}
%
Expressing the deltas directly in their standard Fourier representation, we have:
\begin{align}
Z = &\int \prod_{i,a,\mu} dh_{i a}^\mu d\bar{h}_{i a}^\mu \prod_{\mu} ds^\mu d\bar{s}^\mu  e^{ -\frac{\beta}{2}\sum_{\mu} (y^\mu - s^\mu)^2  + i\sum_{\mu} s^\mu \bar{s}^\mu + i\sum_{\mu, i} h_{i a}^\mu \bar{h}_{i a}^\mu         }  \nonumber\\
& \times \Bigg\langle e^{-i\sum_{\mu} \bar{s}^{\mu}\Big(\frac{1}{\sqrt{N_1}} \sum_{i} v_i^a \sigma(h_{i a}^\mu)\Big)}  \Bigg\rangle_{{v}} \Bigg\langle e^{-i\sum_{\mu, i, a}\bar{h}_{i a}^\mu\Big(\frac{1}{\sqrt{M}}\sum_{m}W_{i m}^a x_{S i+m}^\mu\Big)} \Bigg\rangle_{{W}} \,.
\label{LCN_3_partition}  
\end{align}
where the average is over the distribution of the weights.
Performing these integrals yield a simple expression for the quantities in brackets. Respectively: 
\begin{align}
& \prod_{i,a}  \exp\Bigg\{ -\frac{ \left[ \sum_{\mu=1}^P \bar{s}^\mu\sigma(h_{i a}^\mu) \right] ^2}{2\lambda_1 N_1} \Bigg\} \, , \qquad  \prod_{i,a}  \exp\Bigg\{ -\frac{ \sum_{\mu,\nu = 1}^P \bar{h}_{i a}^\mu C^{ii}_{\mu\nu}  \bar{h}_{i a}^\nu }{2} \Bigg\} \,,
\label{LCN_4_partition}  
\end{align}
where the matrix $C^{ii}_{\mu\nu}$ is the diagonal part of the local covariance matrix defined in Eq. \eqref{local_covariance_matrix}, i.e. 
\begin{equation}
 C^{ii}_{\mu\nu} = \frac{1}{\lambda_0 M}\sum_{m=-\lfloor M/2 \rfloor}^{\lfloor M/2 \rfloor} x^\mu_{S i+ m}x^\nu_{S i+ m}.
\end{equation}
This quantity provides information on the self-correlation of the local patch in position $S i$ for any pair of training patterns $\mu, \nu$ in the dataset. These steps allow to factorize the integrals on the $h$-$\bar{h}$ variables over the patch indices $i$ and channel one $a$, leading to the following equation for the partition function 
\begin{align}
Z &=\int  \prod_{\mu} ds^\mu d\bar{s}^\mu  e^{ -\frac{\beta}{2}\sum_{\mu=1}^P (y^\mu - s^\mu)^2  + i\sum_{\mu=1}^P s^\mu \bar{s}^\mu}  \nonumber\\
 & \, \, \, \times \Bigg\{  \int \prod_i d^Ph_i \,d^P\bar{h}_i \, e^{-\frac{1}{2} \sum_{\mu\nu i} \bar{h}_{i}^\mu C^{ii}_{\mu\nu} \bar{h}_{i}^\nu + i \sum_{\mu i} h_i^\mu \bar{h}_i^\mu  -\frac{1}{2\lambda_1 N_1} \sum_i [\sum_{\mu=1}^P \bar{s}^\mu\sigma(h_{i}^\mu) ] ^2 }\Bigg\}^{N_c}.
\label{LCN_5_partition}  
\end{align}
After the integration over the $\bar{h}$-variables, the quantity in curly brackets becomes:

\begin{equation}
    \prod_i \int d^Ph_i \, P_1^i(\{ h^\mu\}) e^{-\frac{1}{2\lambda_1 N_1} [\sum_{\mu=1}^P \bar{s}^\mu\sigma(h_{i}^\mu) ] ^2  }\,,
    \label{curly_brackets}
\end{equation}
where the probability distribution $P_1^{(i)}$ is given by:
\begin{equation}
P_1^{(i)}(\{h^\mu\}) \equiv \frac{e^{-\frac{1}{2}{h}_i^\top(C^{i i})^{-1}{h}_i}}{\sqrt{(2\pi)^P\det C^{ii}}},
\label{P_1^beta}
\end{equation}
To proceed in the calculation, we introduce a new set of variables $q_i$ through Dirac's $\delta$ identities. In this way Eq. \eqref{curly_brackets} becomes:

\begin{align}
   & \qquad  \prod_i \int d q_i e^{-\frac{1}{2}q_i^2} P(q_i) \, , \qquad P(q_i) = \int d^Ph_i P_1^i(\{h^\mu\}) \delta\Big(q_i - \frac{1}{\sqrt{\lambda_1 N_1}}\sum_{\mu=1}^P\bar{s}^\mu\sigma(h^{\mu}_i)\Big) \nonumber\\
\end{align}
Similarly to \cite{ariosto2022statistical}, we perform a Gaussian approximation on the limiting distribution of $P(q_i)$, heuristically justified by the Breuer-Major theorem \cite{BM}:
\begin{align}
P(q_i) \rightarrow & \mathcal{N}_{q_i}(0,Q_{i}),
\label{BM_th_application}
\end{align}
where the variances are:
\begin{align}
Q_i(\bar{s},C^i) =& \frac{1}{\lambda_1 N_1}\sum_{\mu\nu=1}^P\bar{s}^\mu\Big[\int d^Ph P_1^i(\{h^\mu\}) \sigma(h^\mu) \sigma(h^\nu)\Big] \bar{s}^\nu 
\equiv \frac{1}{\lambda_1 N_1}\sum_{\mu\nu=1}^P\bar{s}^\mu K_{\mu\nu}^i\bar{s}^\nu \, ,
\label{LCN_normalizatoin_vectore}
\end{align}
here $K_{\mu\nu}^i \equiv K_{\mu\nu}^{ii}$ is a compact notation for the diagonal part (in the spatial indices) of the kernel matrix defined in \eqref{CNN_kernel_definition}. After performing the (Gaussian) integrals in the $q_i$ variables, we have: 
\begin{equation}
\Big\{ \prod_i (1+Q_i(\bar{s},C^i))^{-\frac{1}{2}} \Big\}^{N_c} = e^{-\frac{N_c}{2}\sum_i\log{(1+Q_i(\bar{s},C^i))}}.
\label{LCN_sbar_distribution}
\end{equation}
We insert a new set of deltas to handle the $\bar{s}$ dependency in $Q_i(\bar{s},C^i))$:
\begin{equation}
1 = \int dQ_i\,\delta\Big( Q_i - \frac{1}{\lambda_1 N_1}\sum_{\mu\nu=1}^P\bar{s}^\mu K_{\mu\nu}^i\bar{s}^\nu \Big).
\label{LCN_last_delta}
\end{equation}
In this way we are able to perform the Gaussian integration on the $\bar{s}$ variable, after a rescaling  ${\bar{Q}} \rightarrow -i\frac{N_c}{2}{\bar Q}$, obtaining: 
\begin{align}
Z = &\int \prod_i dQ_i d\bar{Q}_i \prod_\mu ds^\mu \, \exp (-\frac{\beta}{2}\sum_{\mu=1}^P(y^\mu-s^\mu)^2) \nonumber \\
&\times \exp \left( -\frac{1}{2}{s}^\top \left[{K}_{\textrm{LCN}}^{(\mathrm{R})}\right]^{-1} {s} -\frac{1}{2}\log \det \left[{K}_{\textrm{LCN}}^{(\mathrm{R})}\right]  - \frac{N_c}{2} \sum_i \left(  Q_i\bar{Q}_i -\log(1+Q_i) \right)  \right) \,,
\label{LCN_7_partition} 
\end{align}
%
%\begin{align}
%Z = &\int \prod_i dQ_i d\bar{Q}_i \prod_\mu ds^\mu d\bar{s}^\mu\, \exp (-\frac{\beta}{2}\sum_{\mu=1}^P(y^\mu-s^\mu)^2 )\nonumber \\
%&\times \exp (i\sum_i Q_i\bar{Q}_i -\frac{N_c}{2}\sum_i\log{(1+Q_i)}    ) \nonumber\\
%&\times \exp (i\sum_{\mu=1}^P s^\mu\bar{s}^\mu- i\sum_i \bar{Q}_i\frac{1}{\lambda_1 N_1}\sum_{\mu\nu=1}^P\bar{s}^\mu K_{\mu\nu}^i\bar{s}^\nu) \,.
%\label{LCN_6_partition} 
%\end{align}
%

where we have identified the renormalized kernel for the LCN:
\begin{align}
\left[K_{\textrm{LCN}}^{(\mathrm{R})} (\bar Q)\right]_{\mu\nu}  \equiv \frac{1}{\lfloor N_0/S\rfloor}\sum_{i=1}^{\lfloor N_0/S\rfloor} \bar Q_{i } K_{\mu\nu}^{i }  \,.
\end{align}
which contains only the diagonal elements of the local kernel matrix defined in Eq. \eqref{KR_CNN}.
%
%The next step is to integrate over $s^\mu$, after a rescaling  ${\bar{Q}} \rightarrow -i\frac{N_c}{2}{\bar Q}$, we have:
%

%
%It is easy to see that the term which depends explicitly on the variable ${s}$ is again Gaussian:
%\begin{equation}
%e^{-\frac{1}{2} {s}^\top \Big[\beta \mathbb{1} +  \left[K_{\textrm{LCN}}^{(\mathrm{R})}\right]^{-1}  \Big] {s} + \beta {y}^\top{s}   },
%\label{s_gaussian}
%\end{equation}
%where we omitted ${y}^\top{y}$, because is a constant term, which does not affect the partition function.\\
Finally, we perform this last Gaussian integral and we are left with a partition function which depends only on ${Q}$ and ${\bar{Q}}$:
\begin{equation}
Z = \int \prod_i dQ_i d\bar{Q}_i e^{-\frac{N_c}{2} S_{\textrm{LCN}}(Q,\bar(Q)},
\label{LCN_8_partition} 
\end{equation}
where the LCN effective action $S_{\textrm{LCN}}$ is given by:
\begin{align}
S_{\textrm{LCN}} \equiv & - \sum_i(  Q_i\bar{Q}_i -\log(1+Q_i)     )  + \frac{\alpha_c}{P} \textrm{Tr} \log \beta \Big(\frac{\mathbb{1}}{\beta} +  {K}_{\textrm{LCN}}^{(\mathrm{R})}\Big) + \frac{\alpha_c}{P} {y}^\top  \Big(\frac{\mathbb{1}}{\beta} +  {K}_{\textrm{LCN}}^{(\mathrm{R})} \Big)^{-1}   {y}.
\label{LCNaction}
\end{align}
We recall that $\Big(\frac{\mathbb{1}}{\beta} +  {K}_{\textrm{LCN}}^{(\mathrm{R})}\Big)$ is a $P\times P$ matrix. Note that this action shares the same functional form as the one for FCN found in \cite{ariosto2022statistical}.
%and ${y}^\top  M    {y} \equiv \sum_{\mu\nu} M_{\mu\nu}y_\mu y_\nu$.
%
\section{Derivation of the effective action for a shallow CNN}
In this section we address the case of shallow CNN architectures, where there is combination of locality and weight sharing. Using the same notation as the previous section, the output of a CNN reads: 
%and so the weight matrices will have only two indices, the channel one (previously indicated with $a$) and the convolution index $A$. As mentioned in the main text, the CNN function reads
\begin{equation}
f_{\textrm{CNN}} (x^\mu) = \frac{1}{\sqrt{N_1}}\sum_{i = 1}^{\lfloor N_0/S \rfloor} \sum_{a = 1}^{N_c} v_i^a \sigma\left(\sum_{m=-\lfloor M/2\rfloor}^{\lfloor M/2\rfloor}\frac{W_{m}^a x_{S i +m}^\mu}{\sqrt{M}} \right),
\label{cnn_function}
\end{equation}
Similarly to the LCN case, we need two sets of deltas, one for the preactivations and one for the outputs, that we directly insert through their standard Fourier representation:
%
%\begin{align}
%Z = &\int \prod_{i,a,\mu} dh_{i a}^\mu \prod_{\mu} ds^\mu \, e^{ -\frac{\beta}{2}\sum_{\mu=1}^P (y^\mu - s^\mu)^2}  \nonumber\\
%& \times \Bigg\langle \prod_{\mu}\delta\Big(s^{\mu}-\frac{1}{\sqrt{N_1}} \sum_{i, a = 1}^{\lfloor\frac{N_0}{S} \rfloor,N_c} v_i^a \sigma(h_{i a}^\mu)\Big) \Bigg\rangle_{{v}} \nonumber\\
%& \times \Bigg\langle \prod_{ i, a, \mu}\delta\Big(h_{i a}^\mu-\frac{1}{\sqrt{M}}\sum_{m=-\lfloor \frac{M}{2} \rfloor}^{\lfloor \frac{M}{2} \rfloor} W_{m}^a x_{S i +m}^\mu\Big) \Bigg\rangle_{{W}} \,,
%\label{CNN_1_partition}  
%\end{align}
%
%where we have already made the delta substitution $f_{\textrm{CNN}} (x^\mu) \rightarrow s^\mu$. We then employ the Fourier representation of the deltas to transform the above equation. This gives us:
%
\begin{align}
Z = &\int \prod_{i,a,\mu} dh_{i a}^\mu d\bar{h}_{i a}^\mu \prod_{\mu} ds^\mu d\bar{s}^\mu  e^{ -\frac{\beta}{2}\sum_{\mu=1}^P (y^\mu - s^\mu)^2  + i\sum_{\mu=1}^P s^\mu \bar{s}^\mu + i\sum_{\mu i a} h_{i a}^\mu \bar{h}_{i a}^\mu         }  \nonumber\\
& \times \Bigg\langle e^{-i\sum_{\mu=1}^P \bar{s}^{\mu}\Big(\frac{1}{\sqrt{N_1}} \sum_{i a} v_i^a \sigma(h_{i a}^\mu)\Big)}  \Bigg\rangle_{{v}}  \Bigg\langle e^{-i\sum_{ i, a, \mu}\bar{h}_{i a}^\mu\Big(\frac{1}{\sqrt{M}}\sum_{m=-\lfloor \frac{M}{2} \rfloor}^{\lfloor \frac{M}{2} \rfloor}W_{m}^a x_{S i +m}^\mu\Big)} \Bigg\rangle_{{W}} \,.
\label{CNN_2_partition}  
\end{align}
Similarly to the LCN case, we get simple contributions for the averaged quantities:
\begin{align}
& \prod_{a}  \exp\Bigg\{ -\frac{1}{2\lambda_1 N_1} \sum_i \left[ \sum_{\mu=1}^P \bar{s}^\mu\sigma(h_{i a}^\mu) \right] ^2\Bigg\} \,, \qquad \prod_{a}  \exp\Bigg\{ -\frac{1}{2}\sum_{i,j} \sum_{\mu,\nu} \bar{h}_{i a}^\mu C_{\mu\nu}^{ij} \bar{h}_{j a}^\nu  \Bigg\} \,,
\label{CNN_3_partition}  
\end{align}
where we have used the definition \eqref{local_covariance_matrix} of the local covariance matrix: 
\begin{equation}
C_{\mu\nu}^{ij} = \frac{1}{\lambda_0 M} \sum_{m = -\lfloor M/2\rfloor}^{\lfloor M/2\rfloor} x^\mu_{S i+m} x^\nu_{S j+m},
\label{local_covariance_matrix_appendix}
\end{equation}
Factorizing over the channel index, the partition function reads:
\begin{align}
Z = &\int  \prod_{\mu} ds^\mu d\bar{s}^\mu  e^{ -\frac{\beta}{2}\sum_{\mu=1}^P (y^\mu - s^\mu)^2  + i\sum_{\mu=1}^P s^\mu \bar{s}^\mu} \nonumber\\
& \times \Bigg\{  \int \prod_i d^Ph_i \,d^P\bar{h}_i \, e^{-\frac{1}{2} \sum_{ij}\sum_{\mu,\nu} \bar{h}_{i a}^\mu C_{\mu\nu}^{ij} \bar{h}_{j}^\nu }  \, e^{ i \sum_{\mu i} h_i^\mu \bar{h}_i^\mu  -\frac{1}{2\lambda_1 N_1}\sum_i [\sum_{\mu=1}^P \bar{s}^\mu\sigma(h_{i}^\mu) ] ^2 }\Bigg\}^{N_c} .
\label{CNN_4_partition}  
\end{align}
The Gaussian integral on the ${\bar{h}}$ variables can be performed. Inside the curly brackets we have:
\begin{equation}
\int \prod_{\mu i} dh_i^\mu \, \mathcal{N}_{h}(0,{C}) e^{-\frac{1}{2\lambda_1 N_1}\sum_i  [\sum_{\mu=1}^P \bar{s}^\mu\sigma(h_{i}^\mu) ] ^2} \, ,
\label{h_bar_integration}
\end{equation}
where the notation $\mathcal{N}_{h}(0,{C})$ indicates a multivariate normalized Gaussian on the $h$ variables, with zero mean and covariance matrix $C_{\mu\nu}^{ij}$  . 
In order to deal with the non linearity term, we define
%repeat the previous strategy, by inserting
a collection of $\lfloor N_0/S \rfloor$ deltas:
\begin{equation}
1 = \int \prod_i dq_i \,\delta\Big(q_i - \frac{1}{\sqrt{\lambda_1 N_1}}\sum_{\mu=1}^P\bar{s}^\mu\sigma(h^{\mu}_i)\Big).
\label{q_beta_deltas}
\end{equation}
Here again, we assume that the joint distribution of the variables $q_i$ can be approximated by a multivariate Normal distribution with $0$ mean and covariance matrix $Q$ with elements:
\begin{equation}
Q_{ij}(\bar{{s}},{C}) \equiv\langle q_i q_j\rangle_{{h}} = \frac{1}{\lambda_1 N_1}\sum_{\mu, \nu}\bar{s}^\mu \langle \sigma(h_i^\mu) \sigma(h_j^\nu) \rangle_{{h}} \bar{s}^\nu \equiv \frac{1}{\lambda_1 N_1}\sum_{\mu, \nu}\bar{s}^\mu K^{ij}_{\mu\nu} \bar{s}^\nu,
\label{covariance_q_beta}
\end{equation}
%where we kept explicit the kernel matrix $K_{\mu\nu}^{ij} \equiv \langle \sigma(h_i^\mu) \sigma(h_j^\nu) \rangle_{{h}}$. 
Note that this would be heuristically justified by a multivariate version of the Breuer-Major theorem. 
%However reasonable, we are currently not able to provide a reference to support this approximation.  
%Note that, since the data covariance ${C}$ is a 4-index matrix which mixes different spatial patches, for the CNN case the covariance ${Q}$ is a matrix, having two indices. 
The quantity in curly brackets is again a multidimensional Gaussian integral on the $q_i$ variables:
\begin{align}
\Big\{\dots \Big\}^{N_c} = & \Bigg\{\int \prod_i dq_j \, \frac{e^{-\frac{1}{2}\sum_{ij}q_i ( \delta_{ij} +  
Q_{ij}^{-1} )q_j  } }{\sqrt{\det2\pi{Q}}} \Bigg\}^{N_c} = \Big\{ \det (\mathbb{1}+{Q})\Big\}^{-\frac{N_c}{2}} = e^{-\frac{N_c}{2}\textrm{Tr} \log  (\mathbb{1}+{Q})} 
%\nonumber \\ = e^{-\frac{N_c}{2}\log \det (\mathbb{1}+{Q})} 
\label{q_gaussian_integral}  
\end{align}
%
%Following this, we can see that the only remaining integrals are ${s}$ and ${\bar{s}}$. However, t
To deal with the implicit dependence of ${Q}$ on ${\bar{s}}$, we introduce one last family of deltas:
%
%\begin{align}
%Z = &\int \prod_{ij} dQ_{ij} d\bar{Q}_{ij} \prod_\mu ds^\mu d\bar{s}^\mu\, \exp (-\frac{\beta}{2}\sum_{\mu=1}^P(y^\mu-s^\mu)^2 )\nonumber \\
%\times& \exp (i\sum_{ij} Q_{ij}\bar{Q}_{ij} -\frac{N_c}{2}\textrm{Tr} \log  (\mathbb{1}+{Q})   ) \nonumber\\
%\times& \exp (i\sum_{\mu=1}^P s^\mu\bar{s}^\mu- i\sum_{ij} \bar{Q}_{ij}\frac{1}{\lambda_1 N_1}\sum_{\mu\nu=1}^P\bar{s}^\mu K_{\mu\nu}^{ij}\bar{s}^\nu)
%\label{CNN_5_partition} 
%\end{align}
%
\begin{equation}
1 = \int dQ_{ij}\,\delta\Big( Q_{ij} - \frac{1}{\lambda_1 N_1}\sum_{\mu\nu=1}^P\bar{s}^\mu K_{\mu\nu}^{ij}\bar{s}^\nu \Big) \, ,
\label{LCN_last_delta_2}
\end{equation}
that allow to perform the integration over $\bar s^\mu$
after the transformation $\bar{Q}_{ij} \rightarrow -\frac{iN_c}{2}\bar{Q}_{ij}$:
\begin{align}
Z = &\int \prod_{ij} dQ_{ij} d\bar{Q}_{ij} \prod_\mu ds^\mu \, \exp \left( \frac{N_c}{2}\sum_{ij}Q_{ij}\bar{Q}_{ij} + \beta\sum_{\mu=1}^P s^\mu y^\mu -\frac{\beta}{2}\sum_{\mu=1}^P y^\mu y^\mu \right) \nonumber \\
\times&\exp\left(-\frac{N_c}{2}\,\textrm{Tr} \log  (\mathbb{1}+{Q}) -\frac{1}{2} \log \det \left[{K}_{\textrm{CNN}}^{(\mathrm{R})}\right] -\frac{1}{2}\sum_{\mu\nu=1}^Ps^\mu\Big(\beta\delta_{\mu\nu} + \left[{K}_{\textrm{CNN}}^{(\mathrm{R})}\right]_{\mu\nu}^{-1}\Big)s^\nu \right)  \, ,
\label{CNN_6_partition} 
\end{align}
where $\left[{K}_{\textrm{CNN}}^{(\mathrm{R})}\right]_{\mu\nu} $ is the renormalized CNN kernel defined in \eqref{KR_CNN}.
%Finally, we are left with one last Gaussian integral on $s^\mu$.
After performing the last Gaussian integral on $s^\mu$, we can finally write:
\begin{equation}
Z = \int \prod_{ij} dQ_{ij} d\bar{Q}_{ij} e^{-\frac{N_c}{2} S_{\textrm{CNN}}({Q},{\bar{Q}})},
\label{CNN_7_partition} 
\end{equation}
where the CNN effective action $S_{\textrm{CNN}}$ reads:
\begin{align}
S_{\textrm{CNN}}({Q},{\bar{Q}}) \equiv & - \sum_{ij}Q_{ij}\bar{Q}_{ij} +\textrm{Tr} \log  (\mathbb{1}+{Q}) +  \frac{\alpha_c}{P} \textrm{Tr} \log \beta \Big(\frac{\mathbb{1}}{\beta} +  {K}_{\textrm{CNN}}^{(\mathrm{R})}\Big) + \frac{\alpha_c}{P} {y}^\top  \Big(\frac{\mathbb{1}}{\beta} +  {K}_{\textrm{CNN}}^{(\mathrm{R})} \Big)^{-1}   {y},
\label{CNNaction_appendix}
\end{align}
where again $\Big(\frac{\mathbb{1}}{\beta} +  {K}_{\textrm{CNN}}^{(\mathrm{R})}\Big)$ is a $P\times P$ matrix, while $\Big(\mathbb{1}+{Q}\Big)$ has dimensions $\lfloor  \frac{N_0}{S} \rfloor \times \lfloor  \frac{N_0}{S} \rfloor$.

\subsection{Saddle point equations from the CNN effective action}
%\begin{equation}
%    S_{\textrm{CNN}}({Q},{\bar{Q}}) \underset{\beta \to \infty}{=} -  \sum_{ij}Q_{ij}\bar{Q}_{ij} +\textrm{Tr} \log  (\mathbb{1}+  {Q}) +  \frac{\alpha_c}{P} \textrm{Tr} \log \beta     \Big(\frac{\mathbb{1}}{\beta} +     {K}_{\textrm{CNN}}^{(\mathrm{R})}\Big) + \frac{\alpha_c}{P} {y}^\top     \Big( {K}_{\textrm{CNN}}^{(\mathrm{R})}     \Big)^{-1}   {y},
%    \label{CNNaction_0_temp}
%\end{equation}
In this section we explicitly derive the saddle point equations of the effective CNN action given in Eq. \eqref{CNNaction_appendix}, that is:
\begin{align}
    & \frac{\partial}{\partial{Q_{ij}}}S_{\textrm{CNN}}({Q},{\bar{Q}}) = 0 \label{SP_barQ}\\
    & \frac{\partial}{\partial{\bar Q_{ij}}}S_{\textrm{CNN}}({Q},{\bar{Q}}) = 0 \label{SP_Q}
\end{align}
Recalling that $\textrm{Tr} \log  (\mathbb{1}+{Q}) \equiv \log \det  (\mathbb{1}+{Q})$ and $\frac{\partial}{\partial{A_{ij}}}\det {A} = \textrm{adj}({A})_{ij}$, Eq. \eqref{SP_barQ} reads:
\begin{equation}
0 = - \bar{Q}_{ij} + (\textrm{adj}(\mathbb{1}+{Q}))_{ij}(\det (\mathbb{1}+{Q}))^{-1}.
\end{equation}
Using the identity ${A}^{-1} = \textrm{adj}({A})(\det{A})^{-1}$ we find
\begin{equation}
\bar{Q}_{ij} = (\mathbb{1}+{Q})^{-1}_{ij},
\label{barQ_saddle_point}
\end{equation}
which holds for each element of ${\bar{Q}}$ and so ${\bar{Q}} = (\mathbb{1}+{Q})^{-1}$.

To compute the derivative in \eqref{SP_barQ}, one should observe that $\left( \frac{\mathbb{1}}{\beta} +  {K}_{\textrm{CNN}}^{(\mathrm{R})}\right) $ can be thought as a $P \times P$ matrix field of the matrix variable $\bar Q$. We use of the following identity for a generic matrix  $A = A(x)$:
\begin{equation}
    \partial_{x_i}\det {A} = (\det {A})\cdot \textrm{Tr} ({A}^{-1}\partial_{x_i}{A}) \, ,
    \label{jacobi}
\end{equation}
and obtain:
\begin{align}
%\partial_{\bar Q_{ij}} S_{\textrm{CNN}}({Q},{\bar{Q}}) 
0 = - Q_{ij} + \frac{\alpha_c}{P}\textrm{Tr}\left[ \Big(\frac{\mathbb{1}}{\beta} + K^{(\mathrm{R})}\Big)^{-1}K^{ij}\right]
- \frac{\alpha_c}{P} \sum_{\mu\rho\lambda\nu}y_\mu \Big(\frac{\mathbb{1}}{\beta} + K^{(\mathrm{R})}\Big)^{-1}_{\mu\rho} K^{ij}_{\rho\lambda}\Big(\frac{\mathbb{1}}{\beta} + K^{(\mathrm{R})}\Big)^{-1}_{\lambda\nu} y_\nu,
\end{align}
which reduces to Eq. \eqref{CNN_SP} in the zero temperature limit.

\section{The similarity matrix of internal representations: differences between FCNs and CNNs \label{6:Predicting observable: the similarity matrix}}

The theoretical framework built so far allows us to predict statistical values of observables. In this section, we will show how to analytically compute the so-called similarity matrix, i.e. the covariance matrix of the hidden representations of the trainset. The more general definition of this observable can be found in the main text in Eq. \ref{similarity_matrix}. Specializing respectively to FCNs and CNNs, we have:
\begin{align}
    & O^{\textrm{FCN}}_{\mu\nu} =     \frac{1}{N_1} \sum_{i=1}^{N_1} \sigma \left( h_i^\mu \right) \sigma\left( h_i^\mu \right) , \qquad \qquad \qquad \qquad h_i^\mu \equiv \frac{{W_i}\cdot {x}^\mu}{\sqrt{N_0}} \label{FCN_kernel_avg}\\
    & O^{\textrm{CNN}}_{\mu\nu} =  \frac{1}{N_c}\sum_{a = 1}^{N_c} \frac{1}{\lfloor\frac{N_0}{S} \rfloor} \sum_{i = 1}^{\lfloor N_0/S \rfloor}\sigma(h^{a\mu}_i )\sigma(h^{a\nu}_i ) , \qquad \, \, \,  h_i^{a \mu} \equiv \frac{1}{\sqrt{M}}\sum_{m=-\lfloor M/2 \rfloor}^{\lfloor M/2 \rfloor}W^a_m x^\mu_{S i+ m}
\label{CNN_kernel_avg}
\end{align}
 The average FCN similarity matrix turns out to have a simple form, and it depends only on the NNGP kernel $K_{\mu\nu}$ and on a naive combination of the labels $y^\mu$:
\begin{equation}
\langle O^{\textrm{FCN}}_{\mu\nu} \rangle = \left(1-\frac{1}{N_1}\right) K_{\mu\nu} + \frac{\lambda_1}{N_1\bar Q}y^\mu y^\nu.
   \label{FCN_sim_matrix} 
\end{equation}
Note that our formalism allows to retrieve this observable for deep linear networks, consistently finding the same result of \cite{SompolinskyLinear}, that is the same expression with the replacement of the NNGP kernel with the data covariance matrix $C_{\mu\nu}$.\\
The CNN case turns out to be more complicated. Indeed, there is again a dependency on the NNGP kernel, but the convolutional architecture mixes different labels in a non-trivial way:
\begin{align}
\langle O^{\textrm{CNN}}_{\mu\nu} \rangle = & \sum_{i}K^{ii}_{\mu\nu} - \frac{1}{\lambda_1N_1}\sum_{ij\lambda\rho}\bar{Q}_{ij}P^{ij}_{\mu\lambda\nu\rho}\nonumber [(K^{(\mathrm{R})})^{-1}_{\lambda\rho}  -\sum_{\epsilon\omega} (K^{(\mathrm{R})})^{-1}_{\lambda\epsilon}  (K^{(\mathrm{R})})^{-1}_{\rho\omega}y^\epsilon y^\omega],
   \label{CNN_sim_matrix} 
\end{align}
where $P^{ij}_{\mu\lambda\nu\rho} \equiv \frac{1}{2} \sum_{k}\Big[K^{ik}_{\mu\lambda}K^{kj}_{\nu\rho} + K^{ki}_{\mu\lambda}K^{jk}_{\nu\rho}\Big]$. In this case, the labels are mixed through the application of the renormalized kernel, incorporating the effect of training.
In the next subsections, we will present a sketch of the computation of the average similarity matrices.

\subsection{FCN averaged similarity matrix}
As a standard practice in statistical mechanics, to compute the averaged similarity matrix, we need to define a modified partition function adding a source term to one of the identical $N_1$ neurons of the last layer: $\frac{1}{2N_1} \left( \sum_\mu J_\mu\sigma(h_1^\mu) \right)^2 $. Since the variables $h^\mu_i$ are statistically identical, the choice $i=1$ will not affect the result. The extended partition function reads: 
\begin{equation}
    Z_{J}^{(\textrm{FCN})} = \int P(\theta) \mathrm{d} \theta \, \exp \left( -\frac{\beta}{2}\sum_{\mu=1}^P \left[y^\mu - f_{\textrm{LCN}}(x^\mu)\right]^2 + \frac{1}{2N_1}\left( \sum_\mu J_\mu\sigma(h_1^\mu) \right)^2 \right) \, 
\end{equation}
where we have reabsorbed the Gaussian priors $\lambda_0, \lambda_1$ in the term $P(\theta) \equiv P(v, W)$. The expected value of the observable in Eq. \eqref{FCN_kernel_avg} is easily computed from the derivatives of the partition function with respect to the source $J_\mu$. It is easy to check that:
\begin{equation}
\langle \sigma(h_1^\mu) \sigma(h_1^\nu) \rangle = - \lambda_1N_1 \frac{1}{Z_{J}^{(\textrm{FCN})}}\frac{\partial^2Z_{J}^{(\textrm{FCN})}}{\partial J^\mu\partial J^\nu} \Big|_{{J}=0},
\label{FCN_sim_twopoint}
\end{equation}

%The computation of the similarity matrix follows the standard approach of Statistical Mechanics, where observable are computed solving $\langle O \rangle = Z^{-1}\int \mathcal{D}x e^{-\beta S(x)}$, where $S$ is the action. The subtle point is that we are not able to directly apply the Breuer-Major theorem when dealing with the integral on the variables $h$, due to the factor $\sigma(h^\mu)\sigma(h^\nu)$. Then, we apply the strategy of adding a specific source term to the partition function, computing derivatives of the new $Z$ respect to this source and then taking the result for vanishing source. The term we add is $\exp\big(\frac{1}{2N_1}(\sum_\mu J_\mu\sigma(h^\mu))^2\big)$. 
Let's take the FCN partition function at the stage where we have already integrated over $v$, $W$ and $\bar h$:
\begin{align}
Z_{J}^{(\textrm{FCN})} =&
\int  \prod_{\mu} ds^\mu d\bar{s}^\mu  e^{ -\frac{\beta}{2}\sum_\mu (y^\mu - s^\mu)^2 + i\sum_\mu s^\mu \bar{s}^\mu}  \Bigg\{  \int dP(h^\mu) \, e^{-\frac{1}{2\lambda_1 N_1} \left[ \sum_{\mu}\bar{s}^\mu \sigma(h^\mu)\right]^2} \Bigg\}^{N_1-1} \nonumber\\
&\times  \int dP(h^\mu_1) \, e^{-\frac{1}{2\lambda_1 N_1} [ \sum_{\mu}\bar{s}^\mu \sigma(h^\mu_1) ]^2} e^{-\frac{1}{2\lambda_1N_1} [ \sum_{\mu}J^\mu \sigma(h^\mu_1) ]^2},
\label{Z_j_FCN}  
\end{align}
%where we have explicitly separated one of the identical $N_1$ integrals on $h^\mu$ and add to it the source term. From now on, we will drop the suffix $i$, because the result will be independent on the particular choice of the neuron. Indeed, in the definition \eqref{FCN_kernel_avg}, the sum and the average operator commute, and in the following we will exploit this feature.\\
%It's easy to check that 
where
\begin{equation}
    dP(h^\mu) \equiv \prod_{\mu}\frac{dh^\mu}{\sqrt{\det{2\pi{C}}}}e^{-\frac{1}{2}\sum_{\mu\nu }h^{\mu} [C^{-1}]_{\mu\nu}h^{\nu}}.
\end{equation}
Here we have separated the $h_1$ integral, but we can immediately drop the dummy index $i=1$, that was reported for the sake of clarity. The integral in curly brackets is the same found in the calculation of the partition function in \cite{ariosto2022statistical}. Leveraging on the same Gaussian equivalence as in the previous sections, we can write: 
\begin{equation}
     \Bigg\{  \int dP(h^\mu) \, e^{-\frac{1}{2\lambda_1 N_1} [ \sum_{\mu}\bar{s}^\mu \sigma(h^\mu)]^2} \Bigg\}^{N_1-1} \to \left[Q(\bar{s})+1\right]^{-\frac{N_1-1}{2}} \, ,
\end{equation}
where $Q(\bar{s})\equiv \frac{1}{\lambda_1 N_1}{\bar{s}}^\top{K}{\bar{s}}$ and $K$ has components $K_{\mu\nu} = \int dP(\lbrace h \rbrace )\sigma(h^\mu)\sigma(h^\nu)$.
To further proceed in the calculation, we need to introduce two more delta function identities: 
\begin{equation}
    1 = \int d q_1 \delta \left(q_1 - \frac{1}{\sqrt{\lambda_1 N_1}} \sum_{\mu}\bar{s}^\mu \sigma(h^\mu) \right) \, , \quad 1 = \int d q_2 \delta \left(q_2 - \frac{1}{\sqrt{\lambda_1 N_1}} \sum_{\mu}J^\mu \sigma(h^\mu) \right)
\end{equation}
Identifying a collective variable ${q} \equiv (q_1, q_2)$, the last integral in \eqref{Z_j_FCN} can be rewritten: 
\begin{equation}
    \int d^2 {q} \, P({q}) \, e^{-\frac{1}{2} \Vert \ q \Vert^2 }\,, \qquad P({q} ) = \int dP(h^\mu)  \delta \left(q_1 - \frac{1}{\sqrt{\lambda_1 N_1}} \sum_{\mu}\bar{s}^\mu \sigma(h^\mu) \right) \delta \left(q_2 - \frac{1}{\sqrt{\lambda_1 N_1}} \sum_{\mu}J^\mu \sigma(h^\mu) \right) 
\end{equation}
%We want to compute this derivative under the integral sign, but right after integrating over $h^\mu$, which is done exploiting the B-M theorem. This time we have to apply it with two variables, let's say $q_1$ and $q_2$.
If the BM theorem's hypotheses hold for $q_1$ and $q_2$, we can use it again to justify a Gaussian equivalence for the distribution $P({q})$:
\begin{equation}
P({q}) \to \mathcal{N}_{q}(0,{\Sigma(\bar s)}) \equiv \frac{e^{-\frac{1}{2}{q}{\Sigma}^{-1}{q}^\top}}{\sqrt{(2\pi)^2\det({\Sigma(\bar s)})}},
\end{equation}
where ${\Sigma (\bar s)}$ is a $2\times2$ covariance matrix defined as:
\begin{equation}
 {\Sigma} = \frac{1}{\lambda_1N_1}
 \begin{pmatrix}
 {\bar{s}}^\top {K} {\bar{s}} & {\bar{s}}^\top {K} {J}\\
 {\bar{s}}^\top {K} {J} & {J}^\top {K} {J}
 \end{pmatrix}
 \, ,
 \label{sigma_FCN_matrix}
\end{equation}
where ${K}$ is the kernel matrix, with components $K_{\mu\nu} = \int dP(h^\rho)\sigma(h^\mu)\sigma(h^\nu)$.
In this way, Eq. \eqref{Z_j_FCN} becomes: 
\begin{equation}
    Z_{J}^{(\textrm{FCN})} = \int  \prod_{\mu} ds^\mu d\bar{s}^\mu  e^{ -\frac{\beta}{2}\sum_\mu (y^\mu - s^\mu)^2 + i\sum_\mu s^\mu \bar{s}^\mu}  \left[Q(\bar{s})+1\right]^{-\frac{N_1-1}{2}} \left[ \mathrm{det} \left( \mathbb{1} + \Sigma(\bar s) \right)\right]^{-\frac{1}{2}}
    \label{Z_j_stage2}
\end{equation}
%Thus, the integral on the last row of Eq. \eqref{Z_j_FCN} becomes equal to
%\begin{equation}
%\int \frac{dq_1dq_2}{\sqrt{(2\pi)^2\det({\Sigma})}} \, e^{-\frac{1}{2}{q}^\top(\mathbb{1}+{\Sigma})^{-1}{q}} = (\det(\mathbb{1} + {\Sigma}))^{-\frac{1}{2}}.
%\label{det_sigma}
%\end{equation}
Let us focus on the determinant in the equation above, which is the only term that depends on the source $J^\mu$. Firstly we point out that, when the source is vanishing, equation \eqref{Z_j_stage2} correctly reduces to the partition function computed in \cite{ariosto2022statistical}:
\begin{equation}
\det(\mathbb{1} + {\Sigma})\Big|_{{J}=0} =  Q(\bar{s}) +1,
\end{equation}
To avoid heavy notation, we have dropped the explicit dependence $\Sigma(\bar s)$, which will be understood implicitly in the following. 
The derivatives with respect to $J^\mu$ and $J^\nu$ can be computed using the Jacobi identity in Eq. \eqref{jacobi}:
\begin{align}
\partial_{J^\mu}\partial_{J^\nu} \left[ \det(\mathbb{1}+{\Sigma})\right]^{-\frac{1}{2}} =  \left[\det(\mathbb{1}+{\Sigma})\right]^{-\frac{1}{2}} \Bigg\{\frac{1}{4}\textrm{Tr}\Big((\mathbb{1}+{\Sigma})^{-1}\partial_{J^\nu}{\Sigma}\Big)\textrm{Tr}\Big((\mathbb{1}+{\Sigma})^{-1}\partial_{J^\mu}{\Sigma}\Big) + \nonumber \\
-\frac{1}{2}\textrm{Tr}\Big((\mathbb{1}+{\Sigma})^{-1}\partial_{J^\mu}{\Sigma}(\mathbb{1}+{\Sigma})^{-1}\partial_{J^\nu}{\Sigma}\Big) + \frac{1}{2}\textrm{Tr}\Big((\mathbb{1}+{\Sigma})^{-1}\partial_{J^\mu}\partial_{J^\nu}{\Sigma}\Big)
\Bigg\} \, ,
\label{second_derivative}
\end{align}
\begin{align}
&\frac{\partial{\Sigma}}{\partial{J^\mu}}\Big|_{{J}=0} = \frac{1}{\lambda_1 N_1}
 \begin{pmatrix}
 0 &  ({K} {\bar{s}})^\mu\\
 ({K} {\bar{s}})^\mu & 0
 \end{pmatrix} \, , \qquad \frac{\partial^2{\Sigma}}{\partial J^\mu\partial J^\nu}\Big|_{{J}=0} = \frac{2}{\lambda_1 N_1}
 \begin{pmatrix}
 0 &  0\\
 0 & K_{\mu\nu}
 \end{pmatrix}
 \, ,
\label{sigma_FCN_derivatives}
\end{align}
where $(K\bar s)^\mu\equiv \sum_\rho K_{\mu\rho}\bar s^\rho$.
%
% Now, it useful to note that
% \begin{equation}
%         (\mathbb{1}+{\Sigma})^{-1}\Big|_{{J}=0} = 
%  \begin{pmatrix}
%  (1+ Q(\bar{s}))^{-1} &  0 \\
%  0 & 1
%  \end{pmatrix}
%  \, .
% \end{equation}
One can see that the first term in curly brackets in Eq. \eqref{second_derivative} vanishes when ${J} = 0$:
\begin{equation}
     (\mathbb{1}+{\Sigma})^{-1}\partial_{J^\mu}{\Sigma}\Big|_{{J}=0} = \frac{1}{\lambda_1 N_1}
 \begin{pmatrix}
 0 & (1+ Q(\bar{s}))^{-1} ({K} {\bar{s}})^\mu\\
 ({K} {\bar{s}})^\mu & 0
 \end{pmatrix}
 \, .
 \label{first_term_vanishes}
\end{equation}
%has zero trace and so the first term in curl brackets in Eq. \eqref{second_derivative} vanishes. The second term is simply the trace of the product of two matrices like the one in Eq. \eqref{first_term_vanishes} and, when ${J} = 0$, it is equal to
The two non-zero contributions give:
\begin{align}
-&\frac{1}{2}\textrm{Tr}\Big((\mathbb{1}+{\Sigma})^{-1}\partial_{J^\mu}{\Sigma}(\mathbb{1}+{\Sigma})^{-1}\partial_{J^\nu}{\Sigma}\Big) = -\frac{1}{(\lambda_1 N_1)^2}\frac{({K} {\bar{s}})^\mu({K} {\bar{s}})^\nu}{1 + Q(\bar{s})} \, , \\
&  \frac{1}{2}\textrm{Tr}\Big((\mathbb{1}+{\Sigma})^{-1}\partial_{J^\mu}\partial_{J^\nu}{\Sigma}\Big)= K_{\mu\nu}/(\lambda_1 N_1)\,,
\end{align}
%Finally, the third term is simply $K_{\mu\nu}/(\lambda_1 N_1)$ and 
Collecting what we have computed so far, we have:
\begin{align}
- \lambda_1 N_1\partial_{J^\mu}\partial_{J^\nu}(\det(\mathbb{1}+{\Sigma}))^{-\frac{1}{2}}\Big|_{{J}=0} = (1 + Q(\bar{s}))^{-\frac{1}{2}} \Big\{ K_{\mu\nu} - \frac{1}{\lambda_1 N_1}\frac{({K} {\bar{s}})^\mu({K} {\bar{s}})^\nu}{1 + Q(\bar{s})}     \Big\}.
\end{align}

The next step is to put back this result in Eq.\eqref{FCN_sim_twopoint}, having already set $J=0$, and to insert the usual Dirac's delta for the $Q({\bar{s}})$. The integral on $s$ is Gaussian, and dropping constant terms we get
\begin{align}
 \langle \sigma(h^\mu) \sigma (h^\nu) \rangle =&
\frac{1}{Z}\int  dQ d\bar{Q} \, e^{-\frac{N_1}{2}\log(1+Q)+\frac{N_1}{2}Q\bar{Q}} \\
& \, \times \int \prod_{\mu} d\bar{s}^\mu e^{ -\frac{1}{2\beta}\sum_\mu (\bar{s}^\mu)^2 + i\sum_\mu y^\mu \bar{s}^\mu -\frac{\bar{Q}}{2\lambda_1}{\bar{s}}^\top{K} {\bar{s}}}  \Big[K_{\mu\nu} - \frac{1}{\lambda_1 N_1}\frac{({K}\cdot{\bar{s}})^\mu({K}\cdot{\bar{s}})^\nu}{1+Q } \Big].
\label{linear_kernel_3}  
\end{align}
The integral on $\bar{s}$ is again Gaussian and can be written as
\begin{align}
e^{-\frac{1}{2}{y}^\top[{\tilde{K}}^{(\mathrm{R})}]^{-1}{y}}\int d^P{\bar{s}} &\,\,  e^{-\frac{1}{2} \big({\bar{s}} +i[{\tilde{K}}^{(\mathrm{R})}]^{-1}\cdot{y} \big)^\top{\tilde{K}}^{(\mathrm{R})} \big({\bar{s}} +i[{\tilde{K}}^{(\mathrm{R})}]^{-1}\cdot{y} \big) } \Big[K_{\mu\nu} - \frac{1}{\lambda_1N_1}\frac{({K}\cdot{\bar{s}})^\mu({K}\cdot{\bar{s}})^\nu}{1+Q } \Big],
\end{align}
where we call ${\tilde{K}}^{(\mathrm{R})} \equiv \Big(\frac{\mathbb{1}}{\beta} +  {K}^{(\mathrm{R})} \Big)$, with ${K}^{(\mathrm{R})} \equiv \bar{Q}{K}/\lambda_1$ the \textit{renormalized} kernel matrix. Note that, in the zero-temperature limit, ${\tilde{K}}^{(\mathrm{R})}$ reduces to ${K}^{(\mathrm{R})}$.
The first term in the square brackets does not depend on $\bar{s}$ and so it will give us a term $K_{\mu\nu}Z$, proportional to the partition function. After the transformation $\bar{s}^\mu \rightarrow \bar{s}^\mu - i\big([{\tilde{K}}^{(\mathrm{R})}]^{-1}\cdot{y}\big)^{\mu}$, we have a zero-mean Gaussian integral, which reads:
\begin{align}
&-\frac{[\det({\tilde{K}}^{(\mathrm{R})})]^{-\frac{1}{2}} }{\lambda_1N_1(1+Q)}\int d^P{\bar{s}} \, \mathcal{N}_{\bar s}(0,({\tilde{K}}^{(\mathrm{R})})^{-1})\Big[({K}\cdot{\bar{s}})^\mu({K}\cdot{\bar{s}})^\nu - ({K}\cdot [{\tilde{K}}^{(\mathrm{R})}]^{-1} \cdot{y})^\mu({K}\cdot [{\tilde{K}}^{(\mathrm{R})}]^{-1}\cdot{y})^\nu         \Big] = \\
&-\frac{[\det({\tilde{K}}^{(\mathrm{R})})]^{-\frac{1}{2}} }{\lambda_1N_1(1+Q)}\sum_{\rho\lambda}K_{\mu\rho}K_{\nu\lambda}\Big[ ([{\tilde{K}}^{(\mathrm{R})}]^{-1})_{\rho\lambda}  -  ([{\tilde{K}}^{(\mathrm{R})}]^{-1}\cdot{y})^\rho ([{\tilde{K}}^{(\mathrm{R})}]^{-1}\cdot{y})^\lambda        \Big].
\end{align}
Then, we can write
\begin{align}
\langle \sigma(h^\mu) \sigma(h^\nu) \rangle =&K_{\mu\nu}+
\frac{1}{Z}\int  dQd\bar{Q} e^{ -\frac{N_1}{2}S_{\textrm{FCN}}(Q,\bar{Q})} \nonumber \\
&\times\frac{1}{\lambda_1N_1(1+Q)}\Big[-\sum_{\lambda\rho}K_{\mu\lambda}K_{\nu\rho}([{\tilde{K}}^{(\mathrm{R})}]^{-1})_{\lambda\rho}  +\sum_{\lambda\rho\epsilon\omega}K_{\mu\lambda}K_{\nu\rho}([{\tilde{K}}^{(\mathrm{R})}]^{-1})_{\lambda\epsilon}([{\tilde{K}}^{(\mathrm{R})}]^{-1})_{\rho\omega}\,y^\epsilon y^\omega       \Big]\,.
\label{linear_kernel_4}  
\end{align}
Finally, we can take the saddle point solution of the effective action, for which $(1+Q)^{-1}=\bar{Q}$ and in the zero-temperature limit  $\beta\rightarrow \infty$, we get
\begin{equation}
\langle \sigma(h^\mu) \sigma(h^\nu) \rangle = \Big(1-\frac{1}{N_1}\Big)K_{\mu\nu}+\frac{\lambda_1}{N_1\bar{Q}}y^\mu y^\nu \,.  
\end{equation}
Since this result is independent on the index $i$, the average of the observable in \eqref{FCN_kernel_avg} is trivially retrieved: 
\begin{equation}
   \langle O^\mathrm{FCN}_{\mu \nu} \rangle = \langle\frac{1}{N_1} \sum_{i=1}^{N_1} \sigma \left( h^\mu_i \right) \sigma\left(h_i^\nu \right)\rangle \equiv \frac{1}{N_1} \sum_{i=1}^{N_1} \langle \sigma \left( h^\mu_i \right) \sigma\left(h_i^\nu \right)\rangle =  \langle \sigma \left( h^\mu \right) \sigma\left(h^\nu \right)\rangle .
\end{equation}

\subsection{CNN's averaged similarity matrix}

In this section, we extend the previous calculation to CNNs. We want to compute the average $\langle O^{\mathrm{CNN}}_{\mu \nu}\rangle$, where the observable $O^{\mathrm{CNN}}_{\mu \nu}$ is defined in \eqref{CNN_kernel_avg}.
%\begin{equation}
%    O^{\textrm{CNN}}_{\mu\nu} \equiv \frac{1}{N_c}\sum_{a = 1}^{N_c} \frac{1}{\lfloor\frac{N_0}{S} \rfloor} \sum_{i = 1}^{\lfloor N_0/S \rfloor}\sigma(h^{a\mu}_i)\sigma(h^{a\nu}_i) .
%\end{equation}
%We recall that the hidden layer variables are defined as $h^{a\mu}_i \equiv (M)^{-1/2}\sum_{m=-\lfloor M/2 \rfloor}^{\lfloor M/2 \rfloor}W^a_m x^\mu_{S i +m}$. 
Similarly to the previous section, we add a source term to the CNN's partition function $\frac{1}{\lambda_1 N_1}\sum_i \left[ \sum_\mu J^\mu \sigma (h^{\mu}_i) \right]^2 $. Here $N_1 = N_c \lfloor N_0/S  \rfloor$ denotes the number of neurons in the last layer of the CNN, reabsorbing the index $a$. The modified partition function reads: 
%taken at the step where we have already integrated out the weights variables ${v}$, ${W}$ and $\bar h$:
%
\begin{align}
&Z_{J}^{(\textrm{CNN})} =
\int  \prod_{\mu} ds^\mu d\bar{s}^\mu  e^{ -\frac{\beta}{2}\sum_\mu (y^\mu - s^\mu)^2 + i\sum_\mu s^\mu \bar{s}^\mu} \Bigg\{  \int dP(h^\mu_i) \, e^{-\frac{1}{2\lambda_1N_1} \sum_i[ \sum_{\mu}\bar{s}^\mu \sigma(h^\mu_i)]^2} \Bigg\}^{N_c-1} \nonumber\\
& \, \, \, \times  \int dP(h^\mu_i) \, e^{-\frac{1}{2\lambda_1N_1} \sum_i[ \sum_{\mu}\bar{s}^\mu \sigma(h^\mu_i) ]^2} e^{-\frac{1}{2\lambda_1N_1}\sum_i [ \sum_{\mu}J^\mu \sigma(h^\mu_i) ]^2},
\label{Z_j_CNN}  
\end{align}
where:
\begin{equation}
    dP(h^\mu_i) \equiv \prod_{\mu i}\frac{dh^\mu_i}{\sqrt{\det{2\pi{C}}}}e^{-\frac{1}{2}\sum_{\mu\nu ij}h^{\mu}_i [C^{-1}]^{ij}_{\mu\nu}h^{\nu}_j}.
\end{equation}
Analogously to the FCN case, one can check that:

\begin{equation}
\Big\langle \frac{1}{\lfloor\frac{N_0}{S} \rfloor} \sum_{i = 1}^{\lfloor N_0/S \rfloor}\sigma(h^{a\mu}_i)\sigma(h^{a\nu}_i) \Big\rangle = - \lambda_1N_c \frac{1}{Z_{J}^{(\textrm{FCN})}}\frac{\partial^2Z_{J}^{(\textrm{FCN})}}{\partial J^\mu\partial J^\nu} \Big|_{{J}=0}.
\label{derivative_partition}
\end{equation}
The next step is to insert two families of Dirac's $\delta$'s:
%, both for ${\bar s}$ and ${J}$: 
\begin{align}
& \int \prod_i dq^i_{\bar s} \delta\big(q^i_{\bar s} - \frac{1}{\sqrt{\lambda_1N_1}}\sum_\mu\bar s^\mu\sigma(h^\mu_i)\big) \, , \quad \int \prod_i dq^i_{J} \delta\big(q^i_{J} - \frac{1}{\sqrt{\lambda_1N_1}}\sum_\mu\bar J^\mu\sigma(h^\mu_i)\big)  \,,
\end{align}
To make progress, we need again to make a Gaussian approximation for the joint distribution of these new variables $P(q_{\bar s}, q_J )$: 
%Leveraging on the B-M theorem, we heuristically justify this Gaussian equivalence and write:
\begin{equation}
    P({q_{\bar s}}, {q_J}) \to  (\det{2\pi{\Sigma}})^{-1/2}\exp{\Big(-\frac{1}{2}({q_{\bar s}}\, |\, {q_{J}})^\top {\Sigma}^{-1}({q_{\bar s}}\, |\,{q_{J}})\Big)}\, ,
\end{equation}
where we made use of the notation for the concatenation of two vectors:
\begin{equation}
    ({q_{\bar s}}\, |\, {q_{J}}) \equiv (q_{\bar s}^{(1)}, \cdots ,q_{\bar s}^{(\lfloor N_0/S\rfloor)},q_{J}^{(1)},\cdots,q_{J}^{(\lfloor N_0/S\rfloor)}).
\end{equation}
Furthermore, the covariance matrix of this joint distribution is a block matrix of the form
\begin{equation}
 {\Sigma} = \frac{1}{\lambda_1N_1}
 \begin{pmatrix}
 {\bar{s}}^\top {K^{ij}} {\bar{s}} &\rvline& {\bar{s}}^\top {K^{ij}} {J}\\ \hline
 {\bar{s}}^\top {K^{ij}} {J} &\rvline& {J}^\top {K^{ij}} {J}
 \end{pmatrix}
 \, ,
 \label{sigma_CNN_matrix}
\end{equation}
where ${\bar{s}}^\top {K^{ij}} {\bar{s}}$ represents the $\lfloor N_0/S \rfloor \times \lfloor N_0/S \rfloor$ matrix which remains after contracting $K^{ij}_{\mu\nu}$ with two vectors on the pattern indices $\mu$ and $\nu$. The integral on the ${q}$ variables is Gaussian and gives the term $\det(\mathbb{1} + {\Sigma})^{-1/2}$. Then, we have to compute the second derivative of this term, analogously as was done for the FCN case, recovering Eq. \eqref{second_derivative}, obviously with a different definition of the matrix ${\Sigma}$.
Then, for the CNN case, we have

\begin{align}
&(\mathbb{1}+{\Sigma})^{-1}\Big|_{{J}=0} = 
 \begin{pmatrix}
 (\mathbb{1}+ {Q}(\bar{s}))^{-1} &\rvline&  {0} \\ \hline
 {0} &\rvline& \mathbb{1}
 \end{pmatrix} \nonumber \\
&(\mathbb{1}+{\Sigma})^{-1}\partial_{J^\mu}{\Sigma}\Big|_{{J}=0} =  
 \frac{1}{\lambda_1N_1}\begin{pmatrix}
 {0} &\rvline&  \sum_{ k}(\mathbb{1}+ {Q}(\bar{s}))^{-1}_{i k} ({K^{ kj}} {\bar{s}})^\mu\\ \hline
 ({K^{ij}} {\bar{s}})^\mu &\rvline& {0}
 \end{pmatrix}
 \nonumber \\
 &(\mathbb{1}+{\Sigma})^{-1}\partial_{J^\mu}{\Sigma}(\mathbb{1}+{\Sigma})^{-1}\partial_{J^\nu}{\Sigma}\Big|_{{J}=0} = 
 \frac{1}{(\lambda_1N_1)^2} 
 \begin{pmatrix}
 \sum_{ k l}(\mathbb{1}+ {Q}(\bar{s}))^{-1}_{i k} ({K^{ k l}} {\bar{s}})^\mu  ({K^{ l j}} {\bar{s}})^\nu &\rvline& {0}\\ \hline
 {0} &\rvline& \sum_{ k l}({K^{i k}} {\bar{s}})^\mu(\mathbb{1}+ {Q}(\bar{s}))^{-1}_{ k l}   ({K^{ l j }} {\bar{s}})^\nu 
 \end{pmatrix} \nonumber \\
 &(\mathbb{1}+{\Sigma})^{-1}\partial_{J^\mu}\partial_{J^\nu}{\Sigma}\Big|_{{J}=0} = 
 \frac{2}{\lambda_1N_1} 
 \begin{pmatrix}
 {0} &\rvline&  {0} \\ \hline
 {0} &\rvline& K^{ij}_{\mu\nu}
 \end{pmatrix}
 \, ,
 \label{det_terms_CNN}
\end{align}

where we have defined $[Q(\bar{s})]_{ij} \equiv  \frac{1}{\lambda_1N_1}{\bar{s}}^\top {K^{ij}} {\bar{s}}$, and $({K^{ij}} {\bar{s}})^\mu \equiv \sum_{\nu} K^{ij}_{\mu\nu}\bar{s}^\nu$. In these equations, the second one has zero trace, while the trace of the third reads
\begin{align}
\textrm{Tr}\Big((\mathbb{1}+{\Sigma})^{-1}\partial_{J^\mu}{\Sigma}(\mathbb{1}+{\Sigma})^{-1}\partial_{J^\nu}{\Sigma}\Big|_{{J}=0}\Big) = \sum_{ij\lambda\rho} \frac{2}{(\lambda_1N_1)^2}(\mathbb{1}+{Q(\bar s}))^{-1}_{ij} \bar{s}^{\lambda} \bar{s}^{\rho} P^{ij}_{\mu \lambda \nu \rho} \, ,
\end{align}
where 
\begin{equation}
 P^{ij}_{\mu \lambda \nu \rho}  = \frac{1}{2} \sum_{ k} K^{i k}_{\mu\lambda}K^{ k j}_{\nu\rho} + K^{ ki}_{\mu\lambda}K^{ j k}_{\nu\rho}\,.
\end{equation}
%where we have divided and multiplied by $2$ in order to get rid of the $1/2$ coefficient, and from now on we will refer to the quantity in square bracket as $P^{ij}_{\mu\lambda\nu\rho}$.\\
Now we are able to solve Eq. \eqref{second_derivative} and we get
\begin{align}
-\lambda_1N_1\partial_{J^\mu}\partial_{J^\nu}(\det(\mathbb{1} + {\Sigma}))^{-\frac{1}{2}} =\sum_{i} K^{ii}_{\mu\nu} -\frac{1}{\lambda_1N_1}\sum_{ij\lambda\rho}(\mathbb{1} + {Q(\bar s)})^{-1}_{ij}\bar s^\lambda\bar s^\rho P^{ij}_{\mu\lambda\nu\rho},
\end{align}
where $\sum_{i}K^{ii}_{\mu\nu}$ is the trace of ${K}$ along the $ij$ indices and is a constant, then we can take it outside the integral. Thus, let's compute the second contribution, that we will call $P_{\mu\nu}$ for sake of notation. First, we insert the deltas for the ${Q}$ variables, in order to get rid of the explicit dependency on ${\bar s}$, and we make use again of the Fourier representation of these deltas, with the variables ${\bar Q}$. Since the new term depends only on ${\bar s}$, we can easily integrate on ${s}$, obtaining (we write only the $\bar s$ integral)
\begin{align}
&\int \prod_\mu d\bar s^\mu \, e^{-\frac{1}{2\beta}\sum_\mu\bar s^{\mu 2} + i\sum_\mu y^\mu \bar s^\mu - \frac{1}{2}\sum_{\mu\nu} K^{(\mathrm{R})}_{\mu\nu}\bar s^\mu\bar s^\nu} P_{\mu\nu} = \nonumber \\
&  e^{-\frac{1}{2}{y}^\top {\tilde{K}^{(\mathrm{R})}}{y}} \int \prod_\mu d\bar s^\mu \, e^{-\frac{1}{2}({\bar s} + i({\tilde{K}^{(\mathrm{R})}})^{-1}{y} )^\top {K^{(\mathrm{R})}}({\bar s} + i({\tilde{K}^{(\mathrm{R})}})^{-1}{y} )   } P_{\mu\nu} \, ,
\end{align}
where $K^{(\mathrm{R})}$ is the renormalized kernel defined in \eqref{KR_CNN} and $\tilde{K}^{(\mathrm{R})} \equiv \frac{\mathbb{1}}{\beta} + K^{(\mathrm{R})} $. This integral is easy to solve once we perform the transformation $\bar s^\mu \rightarrow \bar s^\mu - i\sum_\epsilon[\tilde{K}^{(\mathrm{R})})^{-1}]_{\mu\epsilon} y^\epsilon$, which implies that $P_{\mu\nu}$ becomes
\begin{equation}
\sum_{ij\lambda\rho}\frac{1}{\lambda_1N_1}(\mathbb{1} + {Q(\bar s)})^{-1}_{ij}P^{ij}_{\mu\lambda\nu\rho}[\bar s^\lambda\bar s^\rho  -\sum_{\epsilon\omega} (\tilde{K}^{(\mathrm{R})})^{-1}_{\lambda\epsilon} (\tilde{K}^{(\mathrm{R})})^{-1}_{\rho\omega} y^\epsilon y^\omega].
\end{equation}
With this change of variables, we can perform the Gaussian integral, and we are left only with the partition function having only the ${Q}, {\bar Q}$ terms. Finally, we take all the contribution at the saddle-point solution and at the zero-temperature regime, which means
\begin{align}
(\mathbb{1} + {Q})^{-1} \rightarrow {\bar Q}, \nonumber \\
\tilde{K}^{(\mathrm{R})} \rightarrow {K}^{(\mathrm{R})}.
\end{align}
Collecting what we have so far and noting that Eq. \eqref{derivative_partition} is independent wrt to the channel index, we get the final form of the similarity matrix for a CNN shallow architecture
\begin{align}
\langle O^{\textrm{CNN}}_{\mu\nu} \rangle  =\langle \frac{1}{N_c}\sum_{a = 1}^{N_c} \frac{1}{\lfloor\frac{N_0}{S} \rfloor} \sum_{i = 1}^{\lfloor N_0/S \rfloor}\sigma(h^{a\mu}_i)\sigma(h^{a\nu}_i)\rangle \equiv  \frac{1}{N_c}\sum_{a = 1}^{N_c}\langle \frac{1}{\lfloor\frac{N_0}{S} \rfloor} \sum_{i = 1}^{\lfloor N_0/S \rfloor}\sigma(h^{a\mu}_i)\sigma(h^{a\nu}_i)\rangle = \nonumber\\ \sum_{i}K^{ii}_{\mu\nu} - \frac{1}{\lambda_1N_1}\sum_{ij\lambda\rho}\bar{Q}_{ij}P^{ij}_{\mu\lambda\nu\rho}\nonumber [(K^{(\mathrm{R})})^{-1}_{\lambda\rho}  - \sum_{\epsilon\omega}(K^{(\mathrm{R})})^{-1}_{\lambda\epsilon}  (K^{(\mathrm{R})})^{-1}_{\rho\omega}y^\epsilon y^\omega].
\end{align}

\section{Numerical experiments  \label{num_exp}}

We perform numerical experiments both with shallow FCNs, and CNNs with 1d and 2d convolutions. The networks are trained on two different regression tasks. Respectively: (i) a synthetic random dataset, whose elements have entries sampled from a Gaussian distribution $\mathcal{N}(0,1)$, with labels given by a linear teacher function with unitary weights $t = \lbrace 1 \ldots 1 \rbrace $: 
\begin{equation}
    y^\mu = \frac{1}{2}\left(1+\mathrm{sign}\left( t  \cdot x^\mu \right) \right)
\end{equation}
(ii) a computer vision task: we use the 0 and 1 classes of the CIFAR10 datasets, respectively corresponding to the labels ``cars'' and ``planes''. The images are coarse grained to $N_0=28 \times 28 = D^2 $ pixels and converted to grayscale.

\subsection{Experiments with 1d convolutions}

\begin{figure*}[h]
    \centering
    \includegraphics[width = 1.\textwidth]{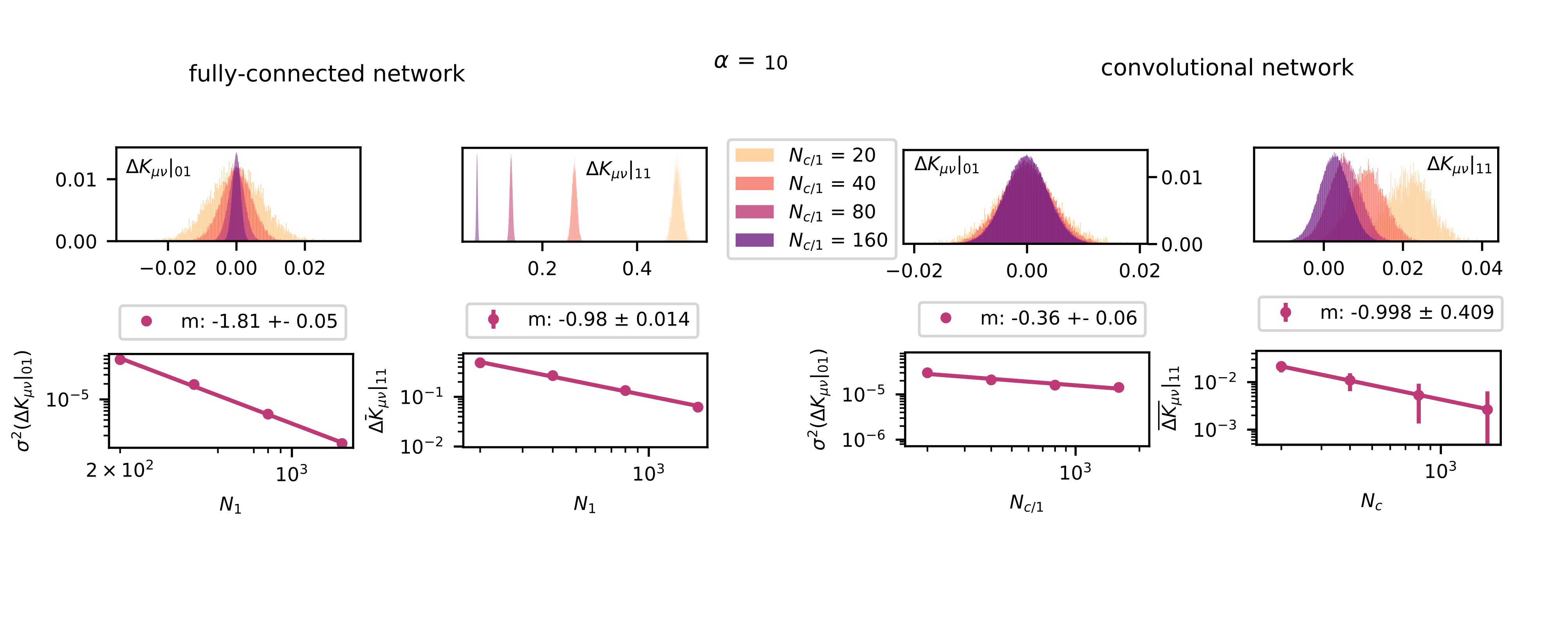}
    \caption{\textbf{Predicting the effect of global and local kernel renormalization on the internal representations of FCNs and CNNs. A finite-size scaling analysis.} Additional experiments at $\alpha_1 = 10$.}
    \label{fig:alpha_10}
\end{figure*}

\begin{figure*}[h]
    \centering
    \includegraphics[width = 1.\textwidth]{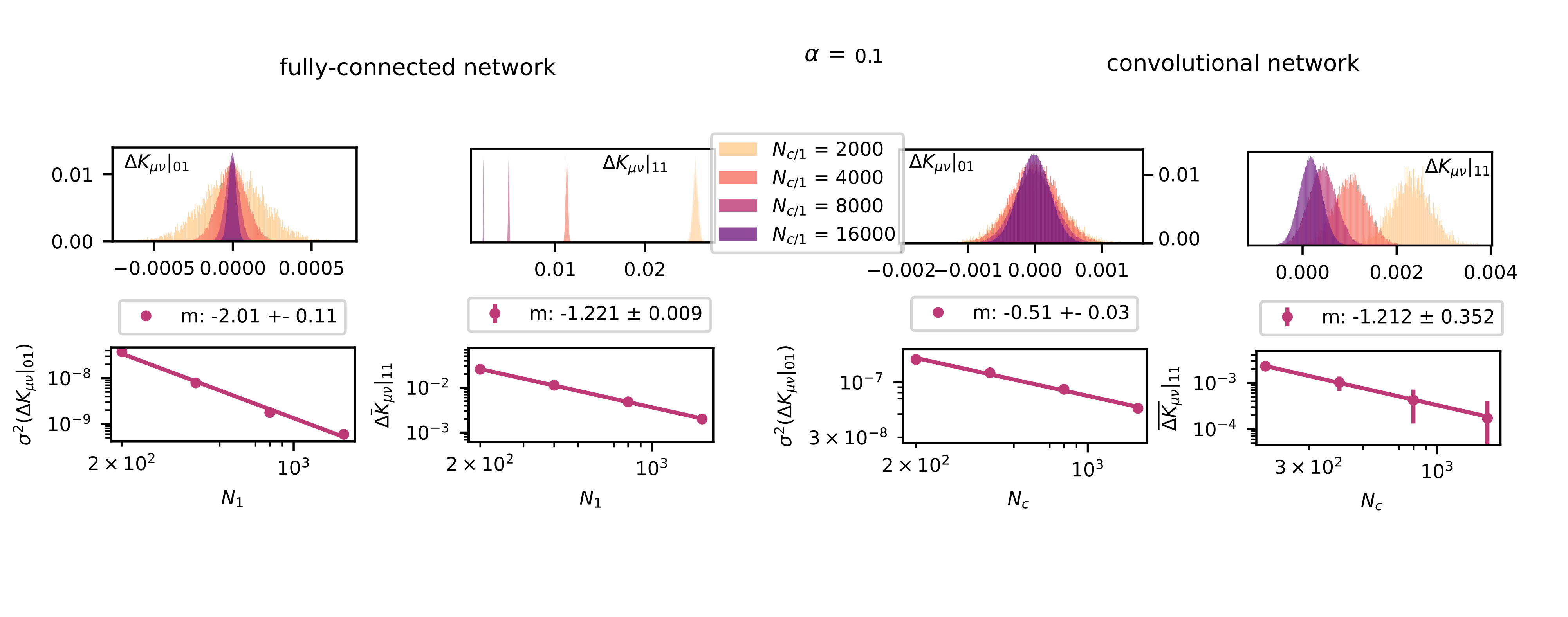}
    \caption{\textbf{Predicting the effect of global and local kernel renormalization on the internal representations of FCNs and CNNs. A finite-size scaling analysis.} Additional experiments at $\alpha_1 = 0.1$}
    \label{fig:alpha_01}
\end{figure*}

The finite-width analysis presented in Fig. \ref{fig:deltaK} was carried out training a one hidden layer network with 1d convolutional filters on the synthetic dataset described in the previous section. The network implements exactly the function reported in Eq. \eqref{1hl_CNN}, with activation function $\sigma(x) = \mathrm{tanh}(x)$. The FC architecture is retrieved setting both the filter size $M$ and the stride $S$ equal to the input dimension $N_0 = M = S$. This correctly implements the function given in Eq. \eqref{FCN_function}. Here the networks are trained using full-batch gradient descent, with ADAM optimizer, implemented with TensorFlow (TF) \cite{tensorflow2015-whitepaper}. We train using a large value of the last layer Gaussian prior $\lambda_1$, until the loss reaches a value of $O(10^{-10})$. We employed a scheduler for the learning rate, reducing its value from $10^{-3}$ to $10^{-7}$ with the so-called ReduceLROnPlateau scheduler of TF.
%, which decreases the learning-rate following the descending rate of the training loss. 
The experiments shown in Fig. \eqref{fig:deltaK} were performed fixing the value of $\alpha_1 = P/N_1 = 1$ with increasing values of $N_1$.
%, in order to study finite-size effects. 
The input size is set to $N_0 = 6400$ and we choose non-overlapping convolutional filters, taking the size of the mask and the stride equal to $M = S = 400$. We built a statistical sample of $O(10^{2})$ networks, trained independently, over which we average each result.
%then we averaged each result over this sample. 
%Then we repeated the same experiment, but changing the value of $\alpha$ in order to cover three orders of magnitude.
In Fig. \ref{fig:alpha_10} and \ref{fig:alpha_01} we show the result for $\alpha_1 = 10$ and $\alpha_1 = 0.1$, respectively. In order to test the consistency of our results, we carried out the same experiments for a smaller value of $N_0$, in particular we chose $N_0 = 1600$ and the same values of $P$ and $N_1$. 
%We found similar results, also in this regime where $N_0$ is comparable with $N_1$.
%The finite-width analysis presented in Fig. \ref{fig:deltaK} was carried out training a 1HL network with fully-connected or 1d convolutional filters on the synthetic dataset described in the previous section. The network implements exactly the function reported in \ref{FCN_function} and \ref{1hl_CNN}, respectively for FCN and CNN, with activation function $\sigma(x) = \mathrm{tanh}(x)$. Here the networks are trained using full-batch gradient descent, with ADAM as optimizer, implemented with TensorFlow \cite{tensorflow2015-whitepaper}. We made use of a dynamical learning-rate, reducing its value from $10^{-3}$ to $10^{-7}$, using the so called ReduceLROnPlateau schedule of TensorFlow, which decreases the learning-rate following the descending rate of the training loss. In every experiment we set a large value of the last layer Gaussian prior and we continued the training until reaching a training loss of $O(10^{-10})$. The experiments were carried out fixing the value of $\alpha_1 = P/N_1$ and taking increasing value of $N_1$, in order to study finite-size effects, in Fig. \ref{fig:deltaK} we show only the $\alpha_1 = 1$ results. The size of the input was $N_0 = 6400$ and we chose non-overlapping convolution filters, setting the size of the mask and of the stride equal to $M = S = 400$. We built a statistical sample of $O(10^{2})$ networks, trained independently, then we averaged each result over this sample.
%.%

\begin{table}[]
\begin{tabular}{|c|c|cc|cc|}
\hline
\multirow{2}{*}{$N_0$} & \multirow{2}{*}{$\alpha_{1/c} = P/N_{1/c}$} & \multicolumn{2}{c|}{Fully-connected network}                                               & \multicolumn{2}{c|}{Convolutional network}                                                 \\ \cline{3-6} 
                       &                                     & \multicolumn{1}{c|}{$\sigma^2(\Delta K_{\mu\nu}|_{01})$} & $\overline{\Delta K_{\mu\nu}}|_{11}$ & \multicolumn{1}{c|}{$\sigma^2(\Delta K_{\mu\nu}|_{01})$} & $\overline{\Delta K_{\mu\nu}}|_{11}$ \\ [1ex] \hline\hline
\multirow{3}{*}{6400}  & 0.1                                 & \multicolumn{1}{c|}{$-2.01 \pm 0.11$}                    & $1.221 \pm 0.009$               & \multicolumn{1}{c|}{$-0.51 \pm 0.03$}                    & $-1.212 \pm 0.352$              \\ \cline{2-6} 
                       & 1                                   & \multicolumn{1}{c|}{$-2.15 \pm 0.01$}                    & $1.153 \pm 0.011$               & \multicolumn{1}{c|}{$-0.2  \pm 0.01$}                    & $-1.045 \pm  0.355$             \\ \cline{2-6} 
                       & 10                                  & \multicolumn{1}{c|}{$-1.81 \pm 0.05$}                    & $ -0.980 \pm 0.014$              & \multicolumn{1}{c|}{$-0.36 \pm 0.06$}                    & $0.998 \pm 0.409$               \\ \hline\hline
\multirow{3}{*}{1600}  & 0.1                                 & \multicolumn{1}{c|}{$-1.34 \pm 0.23$}                    & $-1.278 \pm 0.018$              & \multicolumn{1}{c|}{$-0.014 \pm 0.046$}                  & $-0.9 \pm 2.9$                  \\ \cline{2-6} 
                       & 1                                   & \multicolumn{1}{c|}{$-1.43\pm 0.22$}                     & $-1.154 \pm 0.017$              & \multicolumn{1}{c|}{$0.008 \pm 0.051$}                   & $-0.9 \pm 2.8$                  \\ \cline{2-6} 
                       & 10                                  & \multicolumn{1}{c|}{$-1.35 \pm 0.13$}                    & $-1.003 \pm 0.021$              & \multicolumn{1}{c|}{$-0.087 \pm 0.029$}                  & $-0.9 \pm 2.8$                  \\ [1ex]  \hline
\end{tabular}
\label{tab: sim_matrix_summary}
\caption{\textbf{Additional finite-size scaling experiments on the similarity matrix.} In this table we summarize the results of the various fit for $\alpha_{1/c} = 0.1, 1 , 10$, with two choices of $N_0 = 6400,1600$. The values of $P$ used in all experiments are $P = 200,400,800,1600$. The column denoted by $\sigma^2(\Delta K_{\mu\nu}|_{01})$ shows the trend of the variance of the elements in the submatrix with labels $0,1$, while in the one denoted by $\Delta \bar{K}_{\mu\nu}|_{11}$ we put the fit of the mean value of the elements in the submatrix with labels $11$.}
\end{table}


\subsection{Experiments with 2d convolutions}

The results shown in Fig. \ref{fig:conv2d} are obtained training a one hidden layer architecture with Erf activation and $2d$ non-overlapping convolutional filters on the CIFAR10 binary task discussed above. This is achieved setting the stride $S$ to be equal to the filter mask size $M$. To avoid information loss, we choose $M$ to be an integer divisor of the linear input size $d = 28$. 
To ensure convergence of the posterior weights distribution to the Gibbs ensemble, we train our networks using a discretized Langevin dynamics, similarly to what is done in \cite{SompolinskyLinear,seroussi2023natcomm,ariosto2022statistical}. At each training step $t$ the parameters $\theta = \lbrace W, v\rbrace $ are updated according to: 
\begin{equation}
    \theta(t+1) =  \theta(t) - \eta \nabla_\theta \mathcal{L}(\theta(t)) +\sqrt{2T\eta}\epsilon(t)
\end{equation}
where $T=1/\beta$ is the temperature, $\eta$ is the learning rate, $\epsilon(t)$ is a white Gaussian noise vector with entries drawn from a standard normal distribution, and the loss is the one defined in equation \eqref{loss}. We employ $T = \eta =  2 \cdot 10^{-3}$ throughout all these experiments. This is sufficient to approximate the $T=0$ dynamics in the regime we are considering. This dynamics requires $ \sim 10^6$ steps to reach thermalization, in particular we run the experiment for $5 \cdot 10^6 $ epochs. When possible, we extract the generalization loss within a single run: after the train error has reached its minimum and the test loss is thermalized, we average test loss values every $10^3$ epochs. In the case of FC architecture in Fig. \ref{fig:conv2d}, we averaged over $n=3$ samples to reduce the error.
%To test our theory in the zero-mean activation function case, we used the $\text{Erf}$ function, for which the NNGP kernel can be computed analytically~\cite{NIPS1996_ae5e3ce4, PANG2019270}: 
%\begin{equation}
%K^{\textrm{Erf}}_{\mu\nu} (C) = \frac{2}{\pi} \arcsin\left(\frac{2 C_{\mu\nu}}{\sqrt{\left(1+ 2 C_{\mu\mu}\right)\left(1+ 2 C_{\nu\nu}\right)}}\right)\,.
%\end{equation}

\end{document}